\documentclass[journal]{IEEEtran}
\usepackage{amsmath,amsfonts}
\usepackage{algorithmic}
\usepackage{algorithm}
\usepackage{array}
\usepackage{booktabs}
\usepackage{multirow}
\usepackage{makecell}
\usepackage{tabularx}
\usepackage[caption=false,font=normalsize,labelfont=sf,textfont=sf]{subfig}
\usepackage{textcomp}
\usepackage{stfloats}
\usepackage{url}
\usepackage{verbatim}
\usepackage{graphicx}
\usepackage{cite}
\usepackage{xurl}
\usepackage{fontawesome}
\usepackage{xspace}
\usepackage[table]{xcolor}
\usepackage[hidelinks]{hyperref}
\definecolor{groupbg}{RGB}{229, 239, 255} 
\definecolor{bestcolor}{RGB}{234, 250, 241} 
\definecolor{secondcolor}{RGB}{255, 249, 227} 
\definecolor{htmlblue}{rgb}{0.21,0.49,0.74}

\newcommand{\projectpageurl}{https://buaa-colalab.github.io/vistr-bench-page}
\newcommand{\leaderboardurl}{https://buaa-colalab.github.io/vistr-bench-page/leaderboard.html}
\newcommand{\githuburl}{https://github.com/buaa-colalab/ViSTR-Bench}
\newcommand{\dataseturl}{https://huggingface.co/datasets/homothetic/ViSTR-Bench-Public}

\newcommand{\worldwideweb}{\raisebox{-1.5pt}{\includegraphics[height=1.05em]{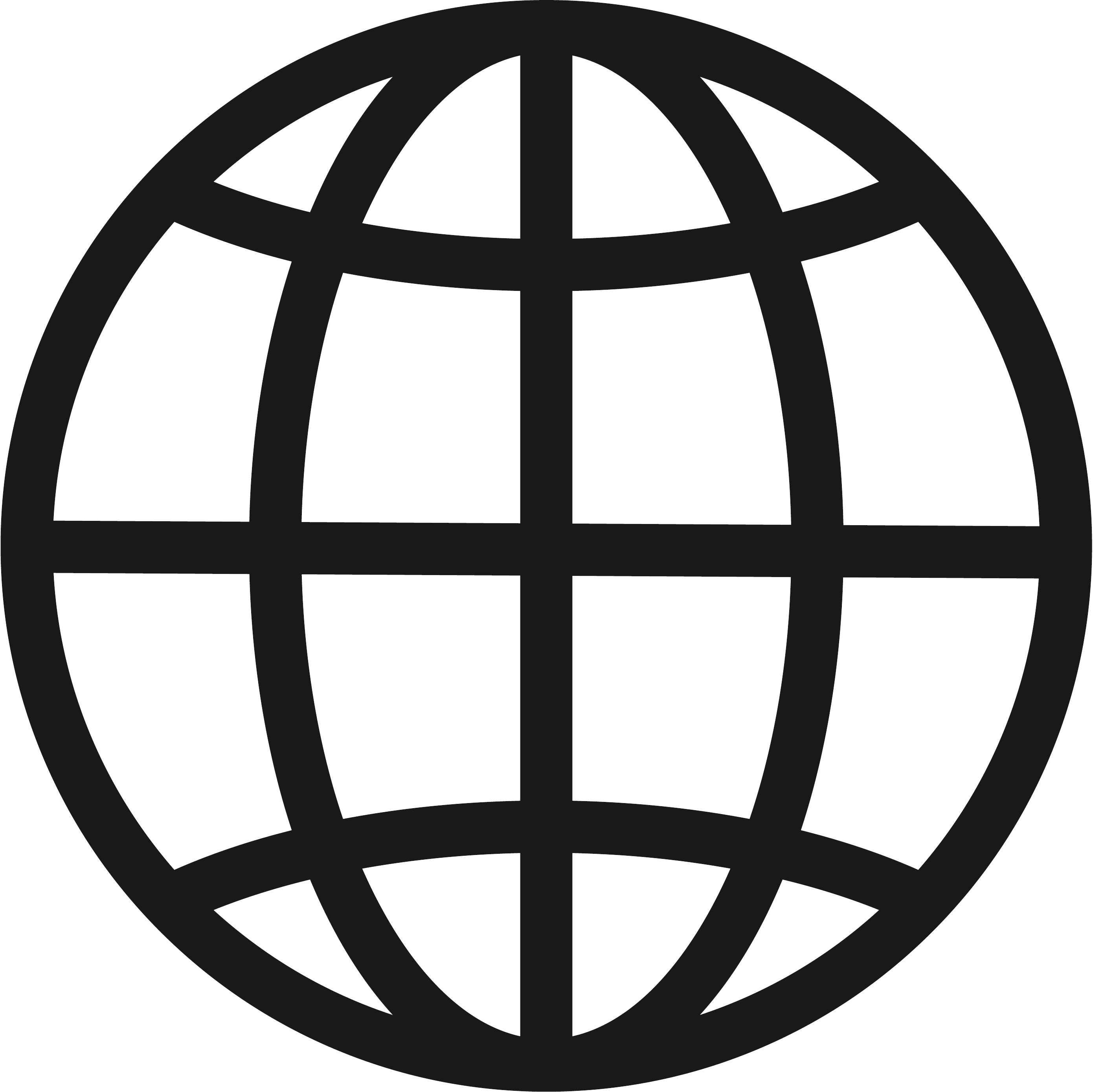}}\xspace}
\newcommand{\leaderboardicon}{\raisebox{-1.5pt}{\includegraphics[height=1.05em]{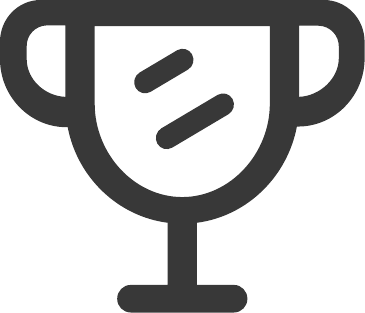}}\xspace}
\newcommand{\github}{\raisebox{-1.5pt}{\includegraphics[height=1.05em]{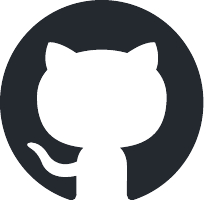}}\xspace}
\newcommand{\huggingface}{\raisebox{-1.5pt}{\includegraphics[height=1.05em]{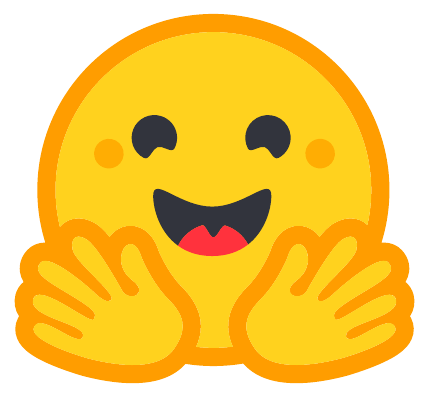}}\xspace}

\newcommand{\best}[1]{\cellcolor{bestcolor}\textbf{#1}}
\newcommand{\second}[1]{\cellcolor{secondcolor}#1}

\DeclareRobustCommand{\capbox}[2]{
    {\setlength{\fboxsep}{0.8pt}
    \raisebox{0pt}[0pt][0pt]{\colorbox{#1}{#2}}}
}
\DeclareRobustCommand{\bestcap}{\capbox{bestcolor}{\textbf{light green}}}
\DeclareRobustCommand{\secondcap}{\capbox{secondcolor}{light yellow}}

\begin{document}

\title{ViSTR-Bench: Can MLLMs Reason from Continuous Visual Cues in Dynamic Scenes?}

\author{Han Li, Si Liu, Zehao Huang, Dongxin Lyu, Longfei Xu, Jiahui Fu, Daxin Tian, Yuliang Xiu, Naiyan Wang \\ [1.2em]
\makebox[0.84\textwidth][c]{%
\makebox[0.21\textwidth][c]{{\worldwideweb \href{\projectpageurl}{{\textcolor{htmlblue}{\text{Project Page}}}}}}%
\makebox[0.21\textwidth][c]{{\leaderboardicon \href{\leaderboardurl}{{\textcolor{htmlblue}{\text{Leaderboard}}}}}}%
\makebox[0.21\textwidth][c]{{\github \href{\githuburl}{{\textcolor{htmlblue}{\text{Evaluation Code}}}}}}%
\makebox[0.21\textwidth][c]{{\huggingface \href{\dataseturl}{{\textcolor{htmlblue}{\text{ViSTR-Bench}}}}}}}
\thanks{Han Li is with the School of Artificial Intelligence, Beihang University, Beijing 100191, China, and also with the Zhongguancun Academy, Beijing 100094, China.
E-mail: lihan0620@buaa.edu.cn.}
\thanks{Si Liu, Longfei Xu, and Jiahui Fu are with the School of Artificial Intelligence, Beihang University, Beijing 100191, China.
E-mails: \{liusi, xulongfei, jiahuifu\}@buaa.edu.cn.}
\thanks{Zehao Huang and Naiyan Wang are in Beijing, China.
E-mails: \{zehaohuang18, winsty\}@gmail.com.}
\thanks{Dongxin Lyu and Yuliang Xiu are with the School of Engineering, Westlake University, Hangzhou 310030, China.
E-mails: \{lyudongxin, xiuyuliang\}@westlake.edu.cn.}
\thanks{Daxin Tian is with the School of Transportation Science and Engineering, Beihang University, Beijing 100191, China, and also with the Zhongguancun Academy, Beijing 100094, China.
E-mail: dtian@buaa.edu.cn.}
}

\markboth{}
{Han Li \MakeLowercase{\textit{et al.}}: ViSTR-Bench}


\maketitle

\begin{abstract}
Multimodal Large Language Models (MLLMs) have achieved remarkable success across diverse expert-level tasks, but they still struggle with fundamental abilities that humans naturally develop through continuous observation of the real world, such as spatial perception and dynamic reasoning.
Recent studies have recognized this gap and introduced dedicated benchmarks to evaluate the spatial-temporal capabilities of MLLMs.
However, existing benchmarks mostly focus on static scenes or require exact quantitative predictions, leaving intuitive reasoning from temporal cues largely underexplored.
In this paper, we introduce the \textbf{Vi}sual \textbf{S}patial-\textbf{T}emporal \textbf{R}easoning \textbf{Bench}mark (\textbf{ViSTR-Bench}), a novel evaluation suite designed to systematically assess whether MLLMs can perform qualitative reasoning from continuous visual cues in dynamic scenes.
Guided by the principles of \textit{temporal emphasis}, \textit{reasoning orientation}, and \textit{qualitative evaluation}, ViSTR-Bench establishes a comprehensive four-dimensional evaluations covering Motion Perception, Spatial Relations, Outcome Prediction, and Physical Dynamics.
The benchmark comprises 15 distinct subtasks and 1,340 high-quality video question-answer pairs spanning diverse tabletop, indoor, and outdoor scenarios.
Extensive evaluations of a broad spectrum of state-of-the-art proprietary, open-source, and specialized spatial MLLMs reveal that, despite their strong general video understanding capabilities, current models still face substantial bottlenecks in complex spatial-temporal reasoning and remain far below human performance.
\end{abstract}


\section{Introduction}

\begin{figure*}[!t]
  \centering
  \subfloat[]{
  \begin{minipage}[t][0.28\textheight][t]{0.57\linewidth}
    \centering
    \includegraphics[height=0.28\textheight]{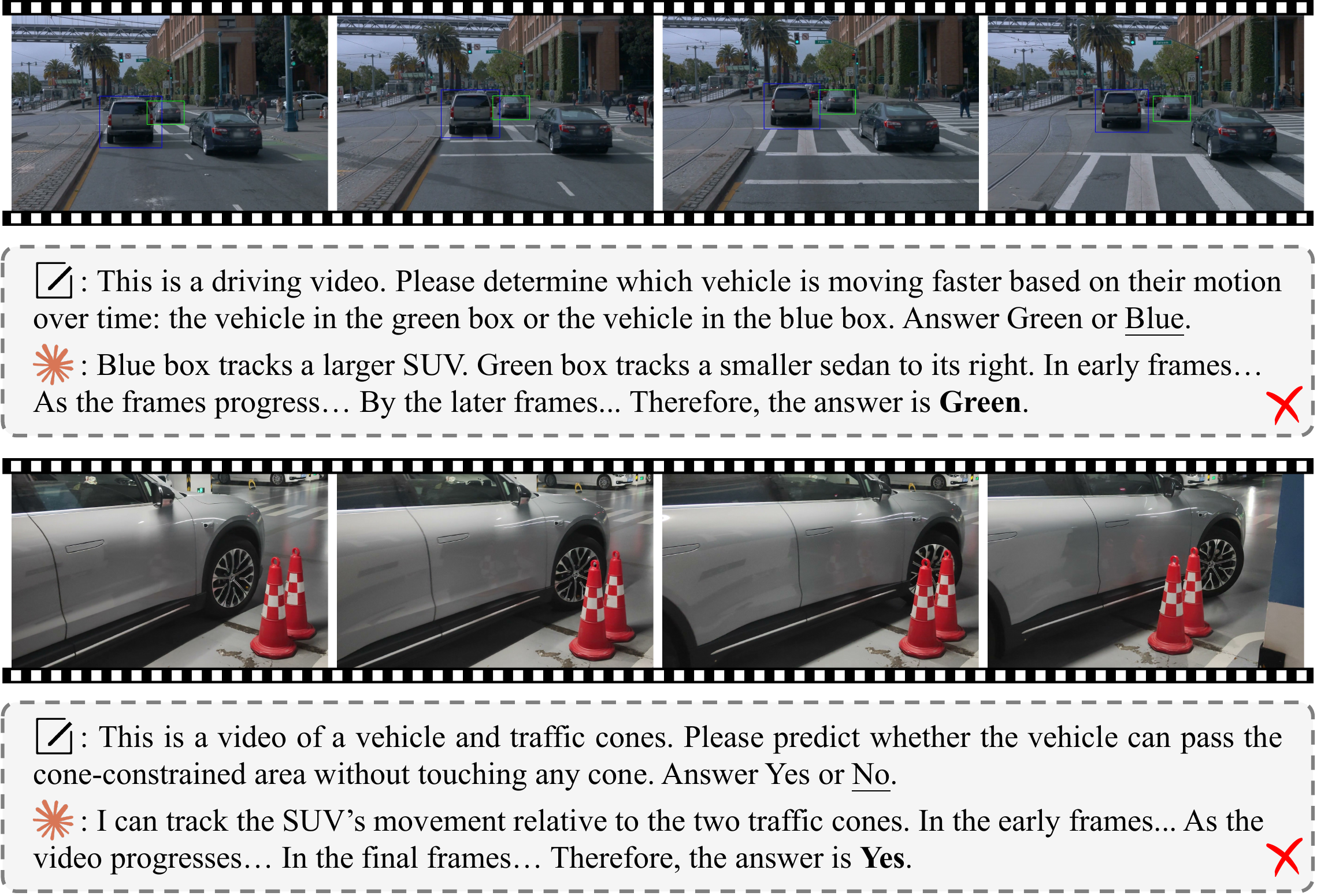}
    \label{fig:1a}
  \end{minipage}}
  \hspace{0.01\linewidth}
  \subfloat[]{
  \begin{minipage}[t][0.28\textheight][t]{0.38\linewidth}
    \centering
    \includegraphics[height=0.28\textheight]{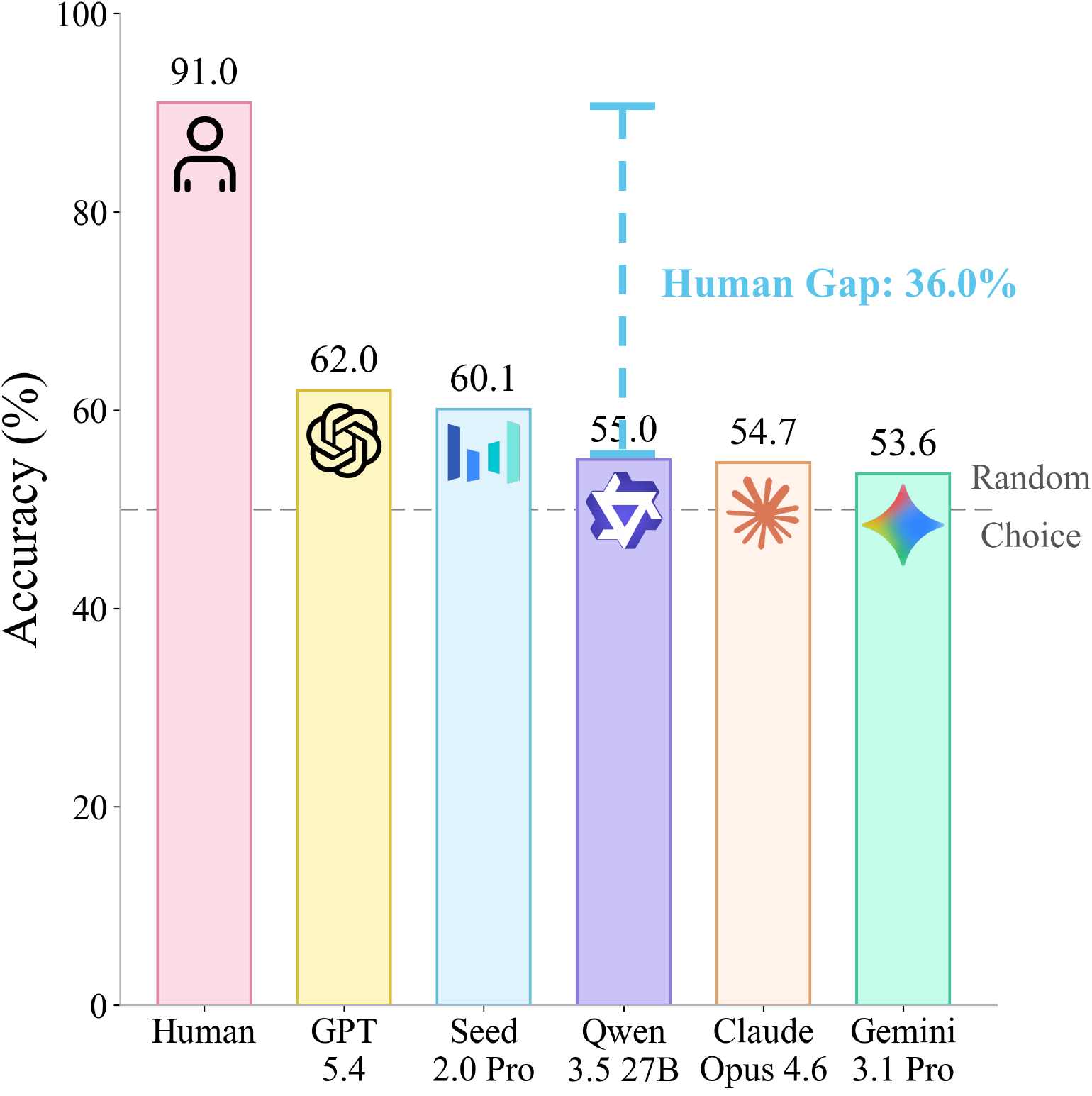}
    \label{fig:1b}
  \end{minipage}}
  \caption{\textbf{Motivating examples and benchmark results.} (a) Two representative examples from ViSTR-Bench, where underlined options indicate the correct answers. Although these questions are visually straightforward for humans, Claude~\cite{Claude} still produces incorrect answers, highlighting the difficulty of reasoning from continuous visual cues in dynamic scenes. (b) Quantitative evaluation results on ViSTR-Bench, showing that current MLLMs remain far from human-level spatial-temporal reasoning performance.}
  \label{fig:1}
\end{figure*}


\IEEEPARstart{I}{n} recent years, the Multimodal Large Language Models (MLLMs) have demonstrated remarkable proficiency across diverse domains, ranging from complex analytical tasks such as mathematical reasoning and programming~\cite{MathVista,MATH-Vision,MATHVERSE,MMCode,Design2Code} to high-quality visual content generation~\cite{Emu3,NExT-GPT,MiniGPT-5}. 
For example, GPT-5 Pro~\cite{GPT_GPQA_2025} achieved 88.4\% accuracy on GPQA~\cite{GPQA}, a PhD-level scientific reasoning benchmark on which human experts attain only 65.0\% accuracy. 
Gemini 2.5 Deep Think~\cite{Gemini_ICPC_2025} also achieved gold-medal-level performance at the International Collegiate Programming Contest (ICPC) World Finals, ranking second among 139 university teams.
However, despite these expert-level capabilities, which typically require long-term specialized training for humans to acquire, MLLMs still struggle with fundamental abilities that humans naturally develop through continuous observation and interaction with the environment, including spatial perception, temporal awareness, and dynamic reasoning, as illustrated in Figure~\ref{fig:1a}.
These spatial-temporal reasoning capabilities are indispensable for real-world applications such as autonomous driving~\cite{LMDrive,DriveLM,DriveGPT4}, robotics~\cite{OpenVLA,RT-2}, and embodied AI~\cite{SayCan,PaLM-E}.

While recent efforts have recognized this gap and introduced benchmarks to evaluate the visual spatial intelligence of MLLMs, several critical limitations remain in current task designs. 
First, existing evaluations~\cite{OmniSpatial,MMSI-Video-Bench,SpaceR,VSI-Bench,Cambrian-S,SPAR-Bench} predominantly focus on static spatial attributes, such as object scale and scene geometry, while largely neglecting the modeling of continuous temporal information. 
Second, although some benchmarks incorporate dynamic scenes~\cite{Dyn-Bench,STI-Bench,OST-Bench,VideoLoom,MLLM-4D,DSI-Bench,DSR-Bench,VLM4D}, their tasks are often restricted to low-level perception, such as object counting or trajectory tracking, falling short of high-level reasoning about dynamic spatial relations and future outcomes.
Third, many tasks~\cite{STI-Bench,VideoLoom,MLLM-4D,DSR-Bench} rely on quantitative evaluation, which may conflate numerical prediction accuracy with genuine spatial-temporal reasoning ability.

To address these limitations, we introduce the \textbf{Vi}sual \textbf{S}patial-\textbf{T}emporal \textbf{R}easoning \textbf{Bench}mark (\textbf{ViSTR-Bench}), designed to systematically evaluate whether MLLMs can reason from continuous visual cues in dynamic scenes. 
Guided by three core principles, namely \textit{temporal emphasis}, \textit{reasoning orientation}, and \textit{qualitative evaluation}, ViSTR-Bench is structured around four task dimensions:
(1) \textbf{Motion Perception} evaluates whether models can identify and compare object motion over time, especially under ego-motion and environmental distractors;
(2) \textbf{Spatial Relations} requires models to reason about evolving geometric relationships and spatial constraints, such as observer-object relations under camera motion and collision-free passage feasibility;
(3) \textbf{Outcome Prediction} assesses the ability to infer current motion states from historical cues and anticipate future outcomes before they are directly observed; 
(4) \textbf{Physical Dynamics} focuses on intrinsic physical properties, dependencies, and stability conditions that are ambiguous in static images but become evident through dynamic interactions.

Specifically, ViSTR-Bench consists of 15 diverse subtasks and 1,340 video question-answer pairs, spanning tabletop, indoor, and outdoor scenes. 
The data are collected from multiple sources, including public datasets (e.g., Ego4D~\cite{Ego4D}, ScanNet~\cite{ScanNet}, ScanNet++~\cite{ScanNet++}, ARKitScenes~\cite{ARKitScenes}, Waymo Open Dataset~\cite{Waymo}, and Motion-X~\cite{MotionX}), web videos (e.g., YouTube), and self-collected recordings. 
We conduct extensive evaluations on a diverse set of advanced MLLMs, including proprietary general-purpose MLLMs (e.g., GPT~\cite{GPT-5}, Gemini~\cite{Gemini}, Claude~\cite{Claude}, Seed~\cite{Seed}, and MiMo~\cite{MiMo-V2.5}), open-source general-purpose MLLMs (e.g., LLaVA~\cite{LLaVA-OneVision-1.5}, Qwen~\cite{Qwen3.5}, InternVL~\cite{InternVL3.5}, and GLM~\cite{GLM-4.6V}), and specialized spatial MLLMs (e.g., VG-LLM~\cite{VG-LLM}, Spatial-MLLM~\cite{Spatial-MLLM}, Spatial-SSRL~\cite{Spatial-SSRL}, and GeoThinker~\cite{GeoThinker}). As shown in Figure~\ref{fig:1b}, experimental results demonstrate that, despite the impressive progress of current MLLMs in general video question answering, they still exhibit substantial bottlenecks in tasks that require subtle motion perception, complex relational reasoning, and temporal cue modeling, with a considerable gap remaining compared to human performance.

Our main contributions are summarized as follows:
\begin{itemize}
\item We introduce the \textbf{Vi}sual \textbf{S}patial-\textbf{T}emporal \textbf{R}easoning \textbf{Bench}mark (\textbf{ViSTR-Bench}) to systematically evaluate whether MLLMs can perform qualitative spatial-temporal reasoning from continuous visual cues in dynamic scenes.

\item We establish a four-dimensional evaluation framework covering Motion Perception, Spatial Relations, Outcome Prediction, and Physical Dynamics, and construct a benchmark comprising 15 diverse subtasks and 1,340 high-quality video question-answer pairs across tabletop, indoor, and outdoor scenarios.

\item We conduct extensive evaluations across proprietary, open-source, and specialized spatial MLLMs, revealing that current models, despite their progress in general video question answering and static spatial reasoning, still fall substantially short of human performance in reasoning about dynamic physical scenes.

\item We provide systematic diagnostic analyses of text-based Chain-of-Thought prompting strategies, visual input formats, and model failure modes, together with pilot studies showing that explicitly providing task-relevant spatial and motion evidence can substantially improve selected reasoning tasks.
\end{itemize}

\section{Related Work}

\begin{figure*}[!t]
  \centering
  \includegraphics[height=0.75\textheight,keepaspectratio]{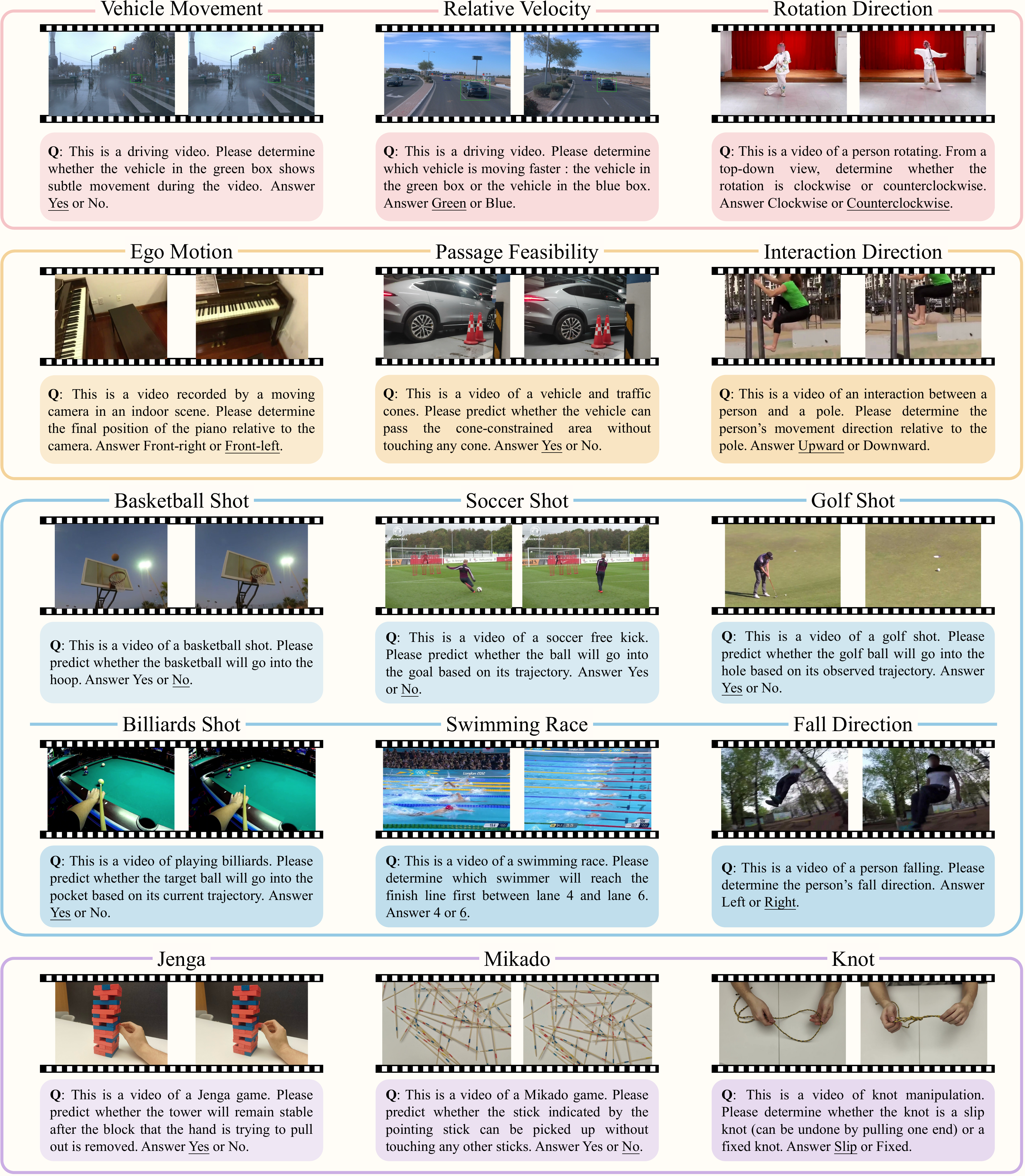}
  \caption{\textbf{Representative task examples from ViSTR-Bench.} Underlined options indicate the correct answers. Each task is formulated as a qualitative binary-choice question that requires models to aggregate temporal evidence and reason about dynamic visual scenes.}
  \label{fig:2}
\end{figure*}

\subsection{Video Benchmarks for MLLMs}
Following the success of MLLMs in static image tasks~\cite{SEED-Bench,MMBench,MM-Vet,MMMU}, numerous benchmarks have emerged to evaluate their video understanding capabilities~\cite{MMBench-Video,Video-MME,MVBench,VITATECS,MSQA,TempCompass,OpenEQA,EgoSchema,Video-Bench,TOMATO,SpatialEval,MM-Ego}. 
However, most of them treat videos merely as temporal sequences of 2D images, leaving reasoning within 3D dynamic scenes largely underexplored. 
To address this, recent studies have evaluated the spatial intelligence of MLLMs~\cite{OmniSpatial,MMSI-Video-Bench,SpaceR,VSI-Bench,Cambrian-S,SPAR-Bench}, though they primarily focus on static properties like object size and relative positioning. 
Subsequent spatio-temporal benchmarks~\cite{Dyn-Bench,STI-Bench,OST-Bench,VideoLoom,MLLM-4D,DSI-Bench,DSR-Bench,VLM4D} introduce dynamic evaluations, such as trajectory tracking and speed estimation. 
Despite these advances, existing benchmarks predominantly assess low-level perception rather than high-level reasoning, and their reliance on quantitative predictions often conflates numerical precision with genuine spatio-temporal reasoning capabilities.

\subsection{MLLMs with Spatial Intelligence}
While MLLMs exhibit remarkable visual understanding capabilities~\cite{LLaVA-OneVision-1.5,Claude,LongVILA,LLaVA-OneVision,VILA,GPT-5,Gemini,InternVL3.5,LongVA,LLaVA-Video}, physically grounding them in the 3D world remains challenging. 
This gap has motivated a growing body of work on spatial MLLMs~\cite{SpatialVLM,VLM-3R,3DRS,GeoThinker,SpatialLadder,SpatialCoT,Spatial-SSRL,SpaceR,SoFar,Spatial-MLLM,VST,GeoSR,Think3D,VG-LLM,LLaVA-3D}.
Spatial-MLLM~\cite{Spatial-MLLM} utilizes a spatial encoder initialized with geometric priors to capture 3D structural information. 
Spatial-SSRL~\cite{Spatial-SSRL} proposes a novel self-supervised reinforcement learning paradigm aimed at enhancing LVLM spatial understanding.
Nevertheless, these approaches are largely restricted to static or indoor environments and lack the capacity to explicitly model dynamic scene evolution. 
To overcome this limitation, recent efforts have extended spatial MLLMs toward 4D reasoning for dynamic real-world settings~\cite{LLaVA-4D,Uni4D-LLM}. 
For instance, LLaVA-4D~\cite{LLaVA-4D} integrates 3D spatial coordinates and temporal cues into a unified spatio-temporal prompt, thereby facilitating comprehensive dynamic scene understanding.

\section{ViSTR-Bench}

In this section, we first define the task taxonomy and reasoning objectives of ViSTR-Bench in Sec.~\ref{sec:task_definition}, and then describe the benchmark construction pipeline in Sec.~\ref{sec:benchmark_construction}.

\subsection{Task Definition}
\label{sec:task_definition}

\begin{figure}[!t]
  \centering
  \includegraphics[width=1.0\linewidth]{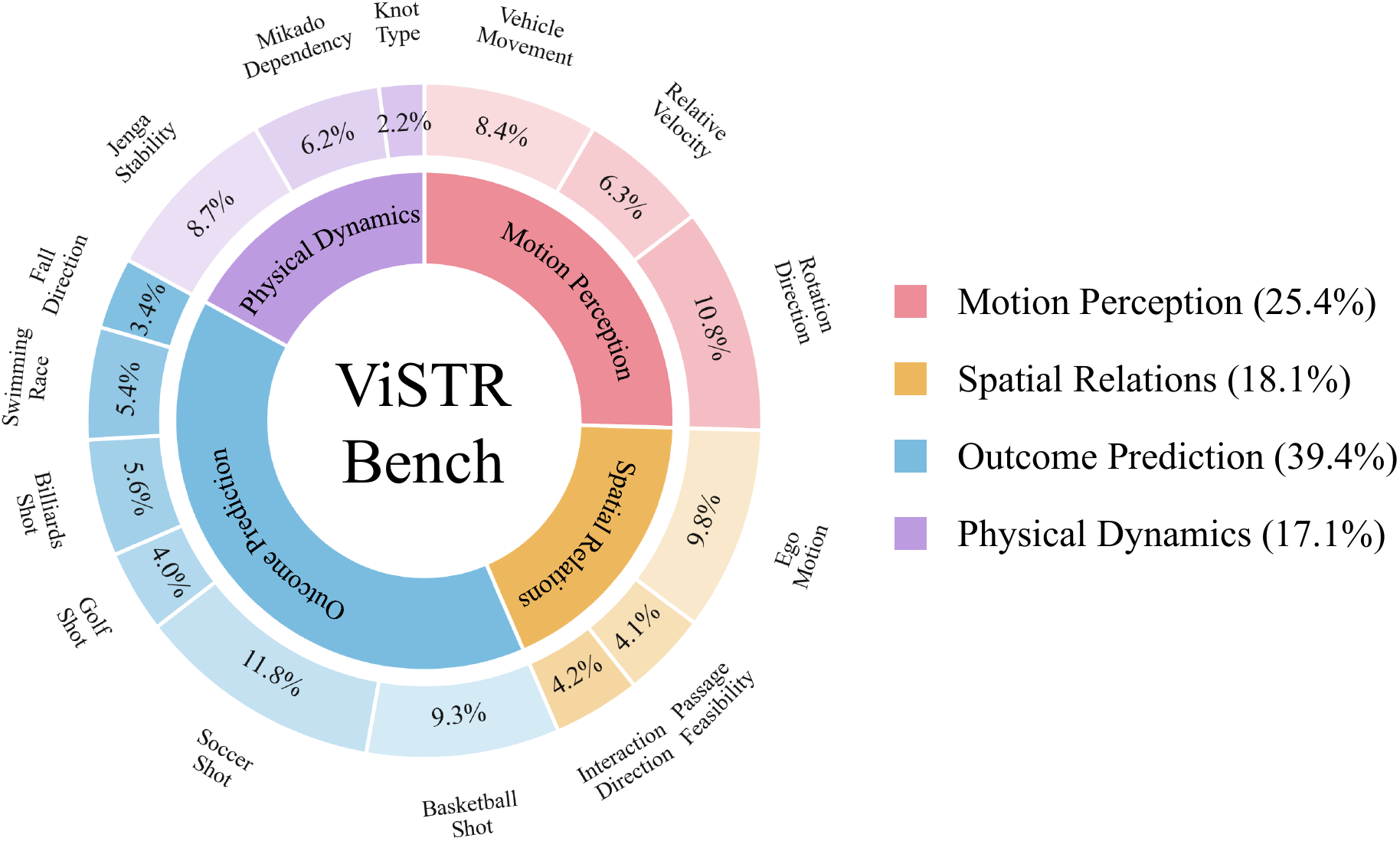}
  \caption{\textbf{Benchmark statistics.} ViSTR-Bench consists of four task dimensions and 15 subtasks.}
  \label{fig:3}
\end{figure}

ViSTR-Bench evaluates whether MLLMs can reason from continuous visual cues in dynamic scenes. Each task requires models to aggregate temporal evidence and make qualitative judgments about motion states, spatial relations, future outcomes, or latent physical properties. 
We organize ViSTR-Bench into four complementary dimensions, namely Motion Perception, Spatial Relations, Outcome Prediction, and Physical Dynamics, which comprise 15 subtasks in total. Representative task examples are shown in Figure~\ref{fig:2}, and the benchmark statistics are summarized in Figure~\ref{fig:3}.

\textbf{Motion Perception.}
This dimension evaluates whether MLLMs can perceive and compare object motion from continuous visual observations.
The core challenge is to distinguish true object motion from apparent visual changes caused by camera movement, viewpoint variation, or environmental distractors. 
\textit{Vehicle Movement} asks models to determine whether a target vehicle is actually moving over time. 
\textit{Relative Velocity} requires models to compare the speeds of two vehicles from an egocentric viewpoint.
\textit{Rotation Direction} examines whether models can identify a person's direction of rotation from varying viewing perspectives.
These tasks test fundamental temporal perception abilities that cannot be reliably solved from a single static frame.

\textbf{Spatial Relations.}
This dimension focuses on reasoning about evolving spatial relationships and geometric constraints in dynamic scenes.
Unlike static spatial understanding, these tasks require models to account for camera motion, object positions, scene layout, and spatial clearance over time. 
\textit{Ego Motion} asks models to infer the spatial relationship between the observing camera and surrounding objects as the camera moves. 
\textit{Passage Feasibility} requires models to determine whether a vehicle can pass through a constrained passage without colliding with obstacles. 
\textit{Interaction Direction} examines whether models can infer a person's movement direction relative to an interacted object from the evolving spatial relationship between them.
These tasks emphasize observer-centric spatial reasoning and constraint-aware scene understanding.

\textbf{Outcome Prediction.}
This dimension evaluates whether MLLMs can infer future outcomes from historical motion cues before the final outcomes are directly observed. 
\textit{Basketball Shot}, \textit{Soccer Shot}, \textit{Golf Shot}, and \textit{Billiards Shot} require models to predict whether a moving object will reach a target based on its observed trajectory. 
\textit{Swimming Race} asks models to determine which swimmer will reach the finish line first from partial observations. 
\textit{Fall Direction} examines whether models can anticipate the direction in which a person will fall from motion cues observed before the fall occurs.
These tasks require models to estimate motion trends, extrapolate trajectories, and reason about outcome-level consequences from temporal evidence.

\textbf{Physical Dynamics.}
This dimension examines whether MLLMs can reason about latent physical properties, dependencies, and stability conditions revealed through dynamic interactions. 
These properties are often ambiguous in static images but can be inferred as the physical process unfolds over time. 
\textit{Jenga Stability} requires models to predict whether a tower remains stable after block removal. 
\textit{Mikado Dependency} asks models to infer whether removing a target stick will disturb other sticks due to contact or support dependencies. 
\textit{Knot Type} requires models to identify the knot structure from its formation or manipulation process. 
These tasks go beyond visible motion and require inferring hidden physical states from temporal dynamics.

\begin{table}[!t]
    \centering
    \caption{
        \textbf{Task-level data source statistics of ViSTR-Bench.}
        For each subtask, we report the total number of QA pairs and the number of samples collected from public datasets, curated web videos, and self-collected recordings.
    }
    \label{tab:detailed_data_sources}
    \small
    \begingroup
    \setlength{\tabcolsep}{2pt}
    \begin{tabularx}{\columnwidth}{l *{4}{>{\centering\arraybackslash}X}}
        \toprule
        \textbf{Task} & \textbf{Total} & \textbf{Public} & \textbf{Web} & \textbf{Collected} \\
        \midrule

        \multicolumn{5}{l}{\textbf{Motion Perception}} \\
        \quad Vehicle Movement & 112 & 112 & 0 & 0 \\
        \quad Relative Velocity & 84 & 84 & 0 & 0 \\
        \quad Rotation Direction & 145 & 41 & 104 & 0 \\
        \midrule

        \multicolumn{5}{l}{\textbf{Spatial Relations}} \\
        \quad Ego Motion & 131 & 131 & 0 & 0 \\
        \quad Passage Feasibility & 55 & 0 & 0 & 55 \\
        \quad Interaction Direction & 56 & 56 & 0 & 0 \\
        \midrule

        \multicolumn{5}{l}{\textbf{Outcome Prediction}} \\
        \quad Basketball Shot & 124 & 124 & 0 & 0 \\
        \quad Soccer Shot & 158 & 0 & 158 & 0 \\
        \quad Golf Shot & 53 & 0 & 53 & 0 \\
        \quad Billiards Shot & 75 & 28 & 47 & 0 \\
        \quad Swimming Race & 72 & 0 & 72 & 0 \\
        \quad Fall Direction & 46 & 8 & 38 & 0 \\
        \midrule

        \multicolumn{5}{l}{\textbf{Physical Dynamics}} \\
        \quad Jenga Stability & 117 & 0 & 0 & 117 \\
        \quad Mikado Dependency & 83 & 0 & 0 & 83 \\
        \quad Knot Type & 29 & 0 & 23 & 6 \\
        \midrule
        
        \textbf{Total} & \textbf{1340} & \textbf{584} & \textbf{495} & \textbf{261} \\
        \bottomrule
    \end{tabularx}
    \endgroup
\end{table}

\subsection{Benchmark Construction}
\label{sec:benchmark_construction}

\begin{figure*}[!t]
  \centering
  \includegraphics[width=0.95\linewidth]{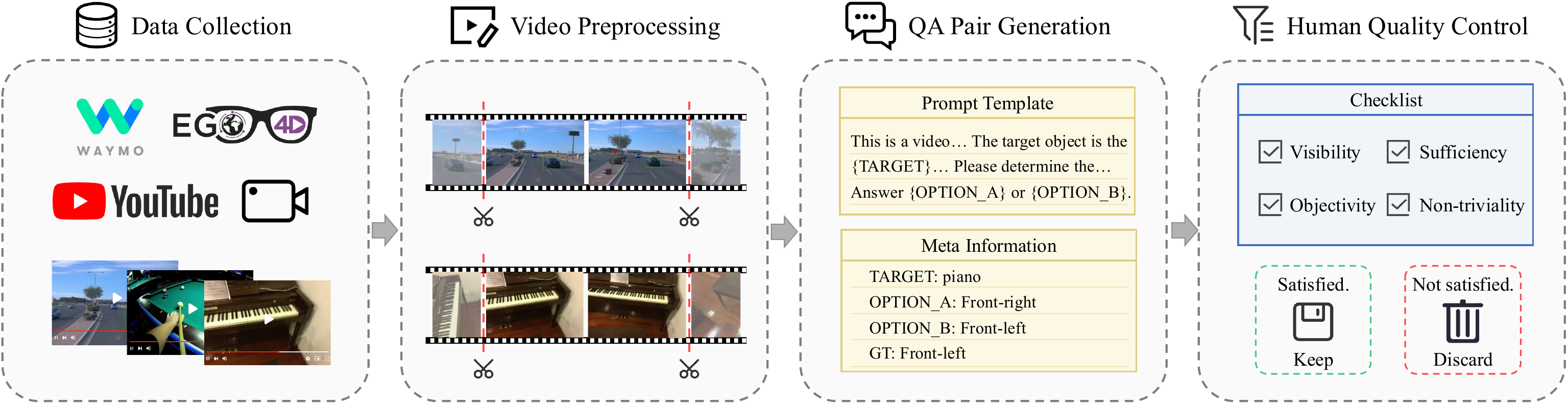}
  \caption{\textbf{Benchmark construction pipeline.} We construct ViSTR-Bench through a systematic multi-stage process, including data collection, video preprocessing, QA pair generation, and human quality control. This pipeline ensures that each sample contains sufficient temporal evidence, supports high-level reasoning, and has an unambiguous qualitative answer.}
  \label{fig:4}
\end{figure*}

We construct ViSTR-Bench through a systematic multi-stage pipeline, including data collection, video preprocessing, question-answer (QA) pair generation, and human quality control, as shown in Figure~\ref{fig:4}. 
This pipeline is designed to ensure that each sample requires temporal evidence, supports high-level reasoning, and has an unambiguous qualitative answer.

\textbf{Data Collection.}
To cover diverse spatial-temporal reasoning scenarios across tabletop, indoor, and outdoor environments, we collect videos from established public datasets, curated web videos, and self-collected recordings.
Specifically, for autonomous driving scenarios, we sample videos from the Waymo Open Dataset~\cite{Waymo}. 
For indoor spatial reasoning, we use egocentric videos from ScanNet~\cite{ScanNet}, ScanNet++~\cite{ScanNet++}, and ARKitScenes~\cite{ARKitScenes}. 
For human-centric reasoning tasks, we select videos from Motion-X~\cite{MotionX}, which provides diverse human motions captured from varying viewpoints.
For sports-related outcome prediction, we extract relevant clips from Ego4D~\cite{Ego4D} and further supplement them with web videos of corresponding sporting events. 
For specialized tasks that are difficult to construct from existing datasets, such as Passage Feasibility and tabletop Physical Dynamics tasks, we conduct task-specific recordings in controlled real-world environments. 
Table~\ref{tab:detailed_data_sources} summarizes the task-level data source statistics of ViSTR-Bench.
More details are provided in Appendix~\ref{sec:detailed_data_sources}. 

\textbf{Video Preprocessing.}
Raw videos inherently contain multiple distinct events, irrelevant temporal context, and explicit outcomes that could trivialize reasoning tasks into simple recognition. 
To systematically eliminate these confounding factors and prevent shortcut learning by models, we process all collected videos through a three-stage preprocessing pipeline:
(1) \textbf{Event Localization.} 
To prevent the entanglement of multiple events and isolate specific reasoning instances, we localize raw videos into discrete, event-level clips and strip away unrelated temporal context.
(2) \textbf{Visual Prompting.} 
For object-centric tasks, textual descriptions alone may introduce referential ambiguity, especially in cluttered scenes with multiple similar objects. 
To explicitly ground target objects, we overlay visual markers (e.g., bounding boxes) onto the corresponding video frames. 
For instance, in the Vehicle Movement and Relative Velocity tasks, we leverage dataset annotations to highlight the specific vehicles in question.
(3) \textbf{Outcome Truncation.} 
For tasks in which the terminal outcome directly reveals the answer, we truncate videos before the outcome becomes visually explicit. 
Instead of using a fixed truncation ratio, we manually determine an instance-specific decision point for each clip, ensuring that the retained segment contains sufficient temporal evidence for human annotators to make a confident qualitative judgment. 
Collectively, these preprocessing steps standardize the visual inputs and mitigate confounding factors from irrelevant context, referential ambiguity, and answer-revealing frames.

\textbf{QA Pair Generation.}
We formulate all instances as binary-choice questions using task-specific templates. Detailed prompt templates are provided in Appendix~\ref{sec:detailed_prompt_templates}.
While most subtasks inherently follow a binary format, those with broader answer spaces, such as the Ego Motion task involving directions including front-left, front-right, back-left, and back-right, are converted to fit this framework. 
Specifically, we pair the ground-truth answer with a plausible distractor randomly sampled from the remaining valid options. To mitigate positional bias during model evaluation, the order of the two candidate options is systematically randomized for each question.

\textbf{Human Quality Control.}
To ensure the reliability of ViSTR-Bench, expert annotators manually filter all candidate QA pairs according to four strict criteria. 
Specifically, we first verify the \textit{visibility} of target objects, ensuring that they are identifiable based on either the textual description or visual prompts. 
We then examine the temporal \textit{sufficiency} of each clip to ensure that it contains adequate visual evidence for reasoning. 
Furthermore, we ensure the \textit{objectivity} of the dataset by requiring all ground-truth answers to be definitive and fully consistent with the observed visual events. 
Finally, we enforce a strict \textit{non-triviality} standard by discarding samples that can be trivially answered from a single static frame or by common-sense language priors. 
Through this rigorous review process, ViSTR-Bench establishes a reliable evaluation foundation for continuous spatial-temporal reasoning.

\textbf{Release and Leaderboard Protocol.}
Upon release, we will make 50\% of ViSTR-Bench (670 QA pairs) publicly available and retain the remaining 50\% (670 QA pairs) as a private held-out set for controlled evaluation.
We will maintain a leaderboard based on the held-out set and periodically update it with results for newly released models.

\section{Evaluation on ViSTR-Bench}

\begin{table*}[!t]
    \centering
    \caption{
        \textbf{Main results on ViSTR-Bench.} 
        We report accuracy (\%) across 15 subtasks grouped into four reasoning dimensions.
        The suffix {\normalfont\itshape -thinking} indicates that the model is evaluated with its built-in thinking mode enabled.
        {\normalfont\itshape Avg.} denotes the overall accuracy averaged over all question-answer pairs, and {\normalfont\itshape Rank} denotes the ranking within each model category.
        The best and second-best results within each model group are highlighted with \bestcap{} and \secondcap{}, respectively.
    }
    \label{tab:main_results}

    \resizebox{\textwidth}{!}{
    \begin{tabular}{l c c ccc ccc cccccc ccc}
        \toprule
        \multirow{2}{*}{\textbf{Method}} 
        & \multirow{2}{*}{\textbf{Rank}} 
        & \multirow{2}{*}{\textbf{Avg.}} 
        & \multicolumn{3}{c}{\textbf{Moti. Perc.}} 
        & \multicolumn{3}{c}{\textbf{Spat. Rela.}} 
        & \multicolumn{6}{c}{\textbf{Outc. Pred.}}  
        & \multicolumn{3}{c}{\textbf{Phys. Dyna.}} \\
        \cmidrule(lr){4-6} \cmidrule(lr){7-9} \cmidrule(lr){10-15} \cmidrule(lr){16-18}
        & & 
        & \makecell{Veh.\\Mov.} & \makecell{Rel.\\Vel.} & \makecell{Rot.\\Dir.}
        & \makecell{Ego\\Mot.} & \makecell{Pas.\\Fea.} & \makecell{Int.\\Dir.}
        & \makecell{Bas.\\Shot} & \makecell{Soc.\\Shot} & \makecell{Golf\\Shot} & \makecell{Bil.\\Shot} & \makecell{Swi.\\Race} & \makecell{Fall\\Dir.}
        & \makecell{Jenga\\Sta.} & \makecell{Mik.\\Dep.} & \makecell{Knot\\Type} \\
        \midrule

        \rowcolor{groupbg}
        \multicolumn{18}{l}{\textbf{Baseline}} \\
        Chance Level (Random) & - & 50.0 & 50.0 & 50.0 & 50.0 & 50.0 & 50.0 & 50.0 & 50.0 & 50.0 & 50.0 & 50.0 & 50.0 & 50.0 & 50.0 & 50.0 & 50.0 \\
        Chance Level (Frequency) & - & 57.9 & 64.3 & 59.5 & 53.1 & 53.4 & 65.5 & 83.9 & 57.3 & 52.5 & 54.7 & 53.3 & 50.0 & 58.7 & 65.0 & 54.2 & 58.6 \\
        \midrule

        \rowcolor{groupbg}
        \multicolumn{18}{l}{\textbf{Proprietary General MLLMs}} \\
        GPT-5.4 & 5 & 56.1 & 66.1 & 67.9 & \second{57.9} & 63.4 & 67.3 & 78.6 & 42.7 & 53.2 & 45.3 & 52.0 & 51.4 & \second{52.2} & 46.2 & 45.8 & 69.0 \\
        GPT-5.4-thinking & \best{1} & \best{62.0} & 57.1 & \second{75.0} & \best{76.6} & \best{75.6} & 58.2 & \best{87.5} & 46.8 & 51.3 & \second{54.7} & \best{58.7} & 50.0 & \best{56.5} & \best{67.5} & 47.0 & \second{72.4} \\
        Gemini-3-Flash-Preview & 14 & 50.7 & 35.7 & 44.0 & 50.3 & 55.7 & 63.6 & 71.4 & 44.4 & 54.4 & 49.1 & 53.3 & \second{59.7} & 41.3 & 44.4 & 50.6 & 62.1 \\
        Gemini-3.1-Flash-Lite-Preview & 11 & 52.6 & 45.5 & 50.0 & 51.7 & 62.6 & 63.6 & 66.1 & 53.2 & 46.8 & 49.1 & 36.0 & \second{59.7} & 43.5 & 60.7 & 44.6 & 65.5 \\
        Gemini-3.1-Pro-Preview & 9 & 53.6 & 45.5 & 52.4 & 50.3 & \second{72.5} & 65.5 & 73.2 & \best{56.5} & 50.0 & 43.4 & 54.7 & 50.0 & 39.1 & 42.7 & 47.0 & \best{75.9} \\
        Claude-Sonnet-4.6 & 13 & 51.0 & \second{67.9} & 40.5 & 46.9 & 64.9 & 52.7 & 55.4 & 54.8 & 55.1 & 50.9 & 48.0 & 50.0 & 34.8 & 34.2 & 38.6 & 62.1 \\
        Claude-Sonnet-4.6-thinking & 10 & 53.4 & 67.0 & 40.5 & 46.9 & 66.4 & 52.7 & 51.8 & 51.6 & \second{62.0} & \best{58.5} & 52.0 & 51.4 & 28.3 & 48.7 & 44.6 & 62.1 \\
        Claude-Opus-4.6 & 6 & 55.4 & 65.2 & 45.2 & 46.9 & 64.1 & \best{70.9} & 57.1 & 49.2 & \best{63.3} & 50.9 & 40.0 & 50.0 & 43.5 & \second{65.8} & 44.6 & \second{72.4} \\
        Claude-Opus-4.6-thinking & 7 & 54.7 & 65.2 & 40.5 & 46.9 & 71.8 & 67.3 & 62.5 & 44.4 & 56.3 & 52.8 & 46.7 & 50.0 & 37.0 & 65.0 & 48.2 & 55.2 \\
        Seed-2.0-Lite & 8 & 54.4 & 64.3 & 60.7 & 41.4 & 51.1 & 63.6 & 66.1 & 52.4 & 55.7 & 49.1 & 49.3 & 48.6 & 43.5 & 65.0 & \best{55.4} & 48.3 \\
        Seed-2.0-Lite-thinking & 3 & 58.8 & \best{68.8} & 67.9 & 52.4 & 66.4 & 65.5 & 69.6 & 49.2 & 61.4 & \best{58.5} & 48.0 & 51.4 & 43.5 & 65.0 & 51.8 & 51.7 \\
        Seed-2.0-Pro & 4 & 56.8 & 57.1 & 67.9 & 52.4 & 58.0 & \best{70.9} & \second{82.1} & 52.4 & 51.3 & 47.2 & \best{58.7} & 55.6 & 43.5 & 63.2 & 48.2 & 48.3 \\
        Seed-2.0-Pro-thinking & \second{2} & \second{60.1} & 63.4 & \best{81.0} & 57.2 & 67.9 & \second{69.1} & \second{82.1} & \second{55.6} & 53.2 & 49.1 & 54.7 & \best{66.7} & 50.0 & 57.3 & 45.8 & 51.7 \\
        MiMo-V2.5 & 12 & 52.4 & 59.8 & 40.5 & 49.7 & 64.9 & 50.9 & 73.2 & 42.7 & 55.7 & 37.7 & \second{56.0} & 52.8 & 45.7 & 51.3 & 42.2 & 62.1 \\
        MiMo-V2.5-thinking & 9 & 53.6 & 51.8 & 38.1 & 51.7 & 58.0 & 63.6 & 80.4 & 50.8 & 53.8 & 35.8 & 50.7 & 51.4 & \second{52.2} & 60.7 & \second{53.0} & 55.2 \\
        \midrule

        \rowcolor{groupbg}
        \multicolumn{18}{l}{\textbf{Open-source General MLLMs}} \\
        LLaVA-OneVision-1.5-4B & 12 & 47.6 & 58.0 & 33.3 & 44.1 & 48.1 & 65.5 & 76.8 & 43.5 & 47.5 & \second{49.1} & 46.7 & 50.0 & 47.8 & 35.0 & 45.8 & 41.4 \\
        LLaVA-OneVision-1.5-8B & 10 & 49.7 & \second{63.4} & 45.2 & \best{53.1} & 49.6 & \second{67.3} & 67.9 & 39.5 & \second{52.5} & 37.7 & \second{56.0} & 50.0 & 41.3 & 35.0 & 45.8 & 41.4 \\
        Qwen3.5-27B-thinking & \best{1} & \best{55.0} & 60.7 & 50.0 & 51.0 & 62.6 & 65.5 & 76.8 & \best{59.7} & \best{55.1} & \second{49.1} & 49.3 & \best{58.3} & 39.1 & 47.9 & 44.6 & 51.7 \\
        Qwen3.5-35B-A3B-thinking & 11 & 49.0 & 46.4 & 45.2 & 46.9 & 58.0 & 63.6 & 67.9 & \second{48.4} & 50.6 & 35.8 & 52.0 & 40.3 & 39.1 & 41.0 & 48.2 & 55.2 \\
        Qwen3.5-122B-A10B-thinking & 3 & 53.4 & 53.6 & 50.0 & 50.3 & 61.8 & \second{67.3} & \best{82.1} & \second{48.4} & 50.0 & 39.6 & \second{56.0} & 43.1 & 39.1 & \second{55.6} & 48.2 & \best{72.4} \\
        Qwen3.5-397B-A17B-thinking & \second{2} & \second{54.4} & 57.1 & \second{57.1} & \second{51.7} & \second{65.6} & 63.6 & \second{80.4} & \best{59.7} & 51.3 & 43.4 & 48.0 & 45.8 & 34.8 & 52.1 & 38.6 & \second{69.0} \\
        InternVL3.5-8B & 7 & 50.9 & 59.8 & \second{57.1} & 51.0 & 62.6 & \second{67.3} & 66.1 & 44.4 & 47.5 & 43.4 & 53.3 & 47.2 & 34.8 & 35.0 & 45.8 & 51.7 \\
        InternVL3.5-14B & 6 & 51.3 & 48.2 & \best{58.3} & \second{51.7} & 54.2 & 65.5 & 66.1 & 41.9 & 46.8 & 45.3 & 52.0 & \best{58.3} & \second{58.7} & 40.2 & \best{57.8} & 41.4 \\
        InternVL3.5-38B & 4 & 52.3 & 43.8 & 38.1 & 44.8 & 59.5 & 65.5 & 75.0 & 44.4 & \second{52.5} & 43.4 & \best{57.3} & 50.0 & 47.8 & \best{65.0} & 50.6 & 65.5 \\
        InternVL3.5-30B-A3B & 8 & 50.1 & 50.9 & 56.0 & 49.0 & 54.2 & \second{67.3} & 78.6 & 43.5 & 48.7 & 32.1 & 37.3 & 51.4 & \best{60.9} & 39.3 & \second{56.6} & 37.9 \\
        InternVL3.5-241B-A28B & 3 & 53.4 & \best{65.2} & 46.4 & 42.8 & \best{66.4} & \best{69.1} & 76.8 & 45.2 & 46.8 & \second{49.1} & 53.3 & 47.2 & 41.3 & \second{55.6} & 49.4 & 62.1 \\
        Intern-S1-thinking & 5 & 51.7 & 48.2 & \second{57.1} & \best{53.1} & 60.3 & 58.2 & 71.4 & \second{48.4} & 47.5 & 35.8 & \second{56.0} & 48.6 & 45.7 & 47.9 & 45.8 & 58.6 \\
        Intern-S1-Pro-thinking & 10 & 49.7 & 53.6 & 40.5 & 48.3 & 63.4 & 56.4 & 73.2 & 41.9 & 47.5 & \best{58.5} & 45.3 & 41.7 & 37.0 & 48.7 & 42.2 & 55.2 \\
        GLM-4.6V-thinking & 9 & 49.9 & 58.0 & 41.7 & 46.9 & 61.8 & 52.7 & 78.6 & 42.7 & 47.5 & 39.6 & 46.7 & \second{52.8} & 47.8 & 35.9 & 49.4 & 65.5 \\
        \midrule

        \rowcolor{groupbg}
        \multicolumn{18}{l}{\textbf{Specialized Spatial MLLMs}} \\
        SpaceR-SFT-7B & \second{2} & \second{51.6} & 47.3 & 58.3 & 46.2 & \best{61.1} & \best{65.5} & 76.8 & 42.7 & \best{55.7} & 54.7 & \second{52.0} & 47.2 & 43.5 & 35.0 & \second{54.2} & 51.7 \\
        VG-LLM-4B & 7 & 48.2 & \second{50.0} & 48.8 & \second{50.3} & 42.0 & \best{65.5} & \best{83.9} & 42.7 & 47.5 & \best{60.4} & \second{52.0} & 45.8 & 41.3 & 35.0 & \second{54.2} & 3.4 \\
        VG-LLM-8B & 9 & 47.6 & 36.6 & 47.6 & \best{53.1} & 35.1 & \second{63.6} & \second{82.1} & \best{46.0} & 48.1 & 52.8 & \best{54.7} & \second{50.0} & 37.0 & 35.0 & 45.8 & 65.5 \\
        Spatial-MLLM-135K-4B & 10 & 47.3 & 42.0 & 50.0 & 46.9 & 57.3 & 34.5 & 75.0 & 41.9 & 43.7 & 50.9 & 46.7 & \best{51.4} & 47.8 & 35.0 & 47.0 & 65.5 \\
        Spatial-MLLM-820K-4B & 3 & 51.3 & \best{64.3} & \second{59.5} & 46.9 & 47.3 & 58.2 & 76.8 & 42.7 & 47.5 & 54.7 & 46.7 & \best{51.4} & 58.7 & 35.0 & 53.0 & \second{69.0} \\
        SpatialLadder-3B & 8 & 47.8 & 42.9 & \best{63.1} & 47.6 & 42.0 & 61.8 & 48.2 & \second{45.2} & 51.9 & 47.2 & 49.3 & \best{51.4} & 43.5 & 35.0 & \second{54.2} & 41.4 \\
        Spatial-SSRL-7B & 5 & 50.3 & 46.4 & 52.4 & 39.3 & \second{58.8} & \best{65.5} & 78.6 & 43.5 & \second{54.4} & 43.4 & 49.3 & 48.6 & 52.2 & 35.0 & \second{54.2} & 65.5 \\
        VST-7B-RL & 6 & 49.4 & \best{64.3} & 51.2 & 42.8 & 44.3 & \best{65.5} & 48.2 & 42.7 & 49.4 & \second{58.5} & 42.7 & \second{50.0} & 56.5 & \second{40.2} & \best{56.6} & 48.3 \\
        GeoThinker-Qwen2.5VL-7B & \best{1} & \best{52.8} & \best{64.3} & 41.7 & 46.2 & 42.7 & \best{65.5} & 80.4 & 41.1 & 47.5 & 56.6 & 46.7 & \second{50.0} & \second{63.0} & \best{65.0} & \second{54.2} & \second{69.0} \\
        GeoThinker-Qwen3VL-8B & 4 & 50.5 & \best{64.3} & \second{59.5} & 46.2 & 39.7 & \best{65.5} & 67.9 & 42.7 & 47.5 & 54.7 & 46.7 & \second{50.0} & \best{67.4} & 35.0 & 49.4 & \best{72.4} \\
        \midrule

        \rowcolor{groupbg}
        \multicolumn{18}{l}{\textbf{Human Evaluation}} \\
        Human & - & 91.0 & 90.6 & 97.0 & 100.0 & 99.6 & 85.5 & 100.0 & 81.5 & 82.9 & 87.7 & 77.3 & 84.7 & 100.0 & 94.4 & 98.8 & 77.6 \\
        \bottomrule
    \end{tabular}
    }
\end{table*}

In this section, we evaluate existing MLLMs on ViSTR-Bench.
All benchmark results are computed using the complete set of 1,340 QA pairs.
We first describe the evaluation setup in Sec.~\ref{sec:evaluation_setup} and present the main results in Sec.~\ref{sec:main_results}.
We then study the effect of text-based CoT prompting in Sec.~\ref{sec:text_cot} and analyze the impact of different visual input formats in Sec.~\ref{sec:visual_input}.
We further present error analysis in Sec.~\ref{sec:error_analysis} and preliminary explorations of model improvement strategies in Sec.~\ref{sec:model_improvement}.
The qualitative visualizations are in Appendix~\ref{sec:visualization}.

\subsection{Evaluation Setup}
\label{sec:evaluation_setup}

\textbf{Evaluated Models.}
We evaluate a broad range of existing MLLMs on ViSTR-Bench, grouped into three categories. 
First, we consider proprietary general-purpose MLLMs, including GPT-5.4~\cite{GPT-5}, Gemini-3/3.1~\cite{Gemini}, Claude-4.6~\cite{Claude}, Seed-2.0~\cite{Seed}, and MiMo-V2.5~\cite{MiMo-V2.5} series. 
For models with controllable thinking modes, we report results for both the standard and thinking variants to examine whether enabling built-in thinking improves their multi-step planning and spatial-temporal reasoning abilities. 
Second, we evaluate open-source general-purpose MLLMs, including LLaVA-OneVision-1.5~\cite{LLaVA-OneVision-1.5}, Qwen3.5~\cite{Qwen3.5}, InternVL3.5~\cite{InternVL3.5}, Intern-S1~\cite{Intern-S1,Intern-S1-Pro}, and GLM-4.6V~\cite{GLM-4.6V}.
For open-source models that support thinking modes, we enable the thinking mode by default. 
Third, we include specialized spatial MLLMs that are explicitly designed to enhance spatial intelligence, including SpaceR~\cite{SpaceR}, VG-LLM~\cite{VG-LLM}, Spatial-MLLM~\cite{Spatial-MLLM}, SpatialLadder~\cite{SpatialLadder}, Spatial-SSRL~\cite{Spatial-SSRL}, VST~\cite{VST}, and GeoThinker~\cite{GeoThinker}. 
Since ViSTR-Bench focuses on visual spatial-temporal reasoning from videos, we only include models that operate purely on visual inputs and text prompts, without requiring additional ground-truth 3D information such as depth maps, camera poses, or point clouds.

\textbf{Input Protocol.}
For models that natively support video input, such as Gemini-3/3.1~\cite{Gemini}, Seed-2.0~\cite{Seed}, MiMo-V2.5~\cite{MiMo-V2.5}, Qwen3.5~\cite{Qwen3.5}, and GLM-4.6V~\cite{GLM-4.6V}, we directly provide the original video clips.
The maximum video resolution is constrained to $1920 \times 1280$, and the maximum Base64-encoded video size is set to 10 MB.
For models that only support image input, such as GPT-5.4~\cite{GPT-5}, Claude-4.6~\cite{Claude}, Intern-S1~\cite{Intern-S1,Intern-S1-Pro}, and InternVL3.5~\cite{InternVL3.5}, we uniformly sample 16 frames from each video.
This protocol enables a unified evaluation across video-based and image-based MLLMs while preserving the temporal cues required by our tasks.

\textbf{Evaluation Metric.}
All tasks in ViSTR-Bench are formulated as qualitative binary-choice questions.
We therefore use accuracy (\%) as the primary evaluation metric.
Specifically, we report per-task accuracy to analyze model performance across different reasoning abilities, and overall accuracy averaged over all QA pairs as the main summary metric.

\textbf{Baselines and Human Evaluation.}
To provide reference performance levels, we include two chance-level baselines.
\emph{Chance Level (Random)} corresponds to randomly selecting one option from the candidate answers for each question. Under our binary-choice setting, this yields an expected accuracy of 50\% across all tasks.
\emph{Chance Level (Frequency)} always selects the most frequent answer within each task. This baseline captures potential dataset biases, such as imbalanced answer distributions.
In addition, we conduct human evaluation to estimate human-level performance on ViSTR-Bench. The human results serve as an approximate upper-bound reference, indicating the performance gap between current MLLMs and human visual spatial-temporal reasoning.

\begin{figure*}[!t]
  \centering
  \includegraphics[width=0.9\linewidth]{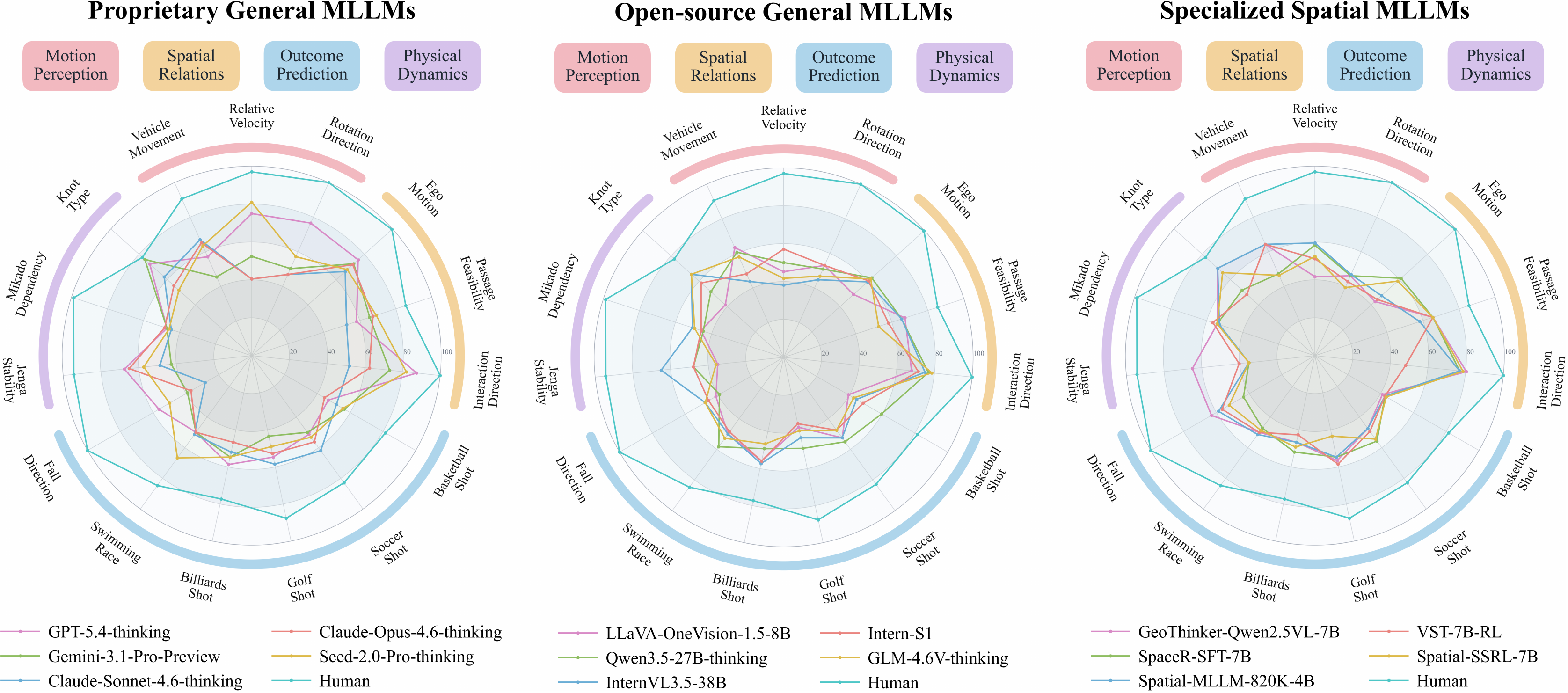}
  \caption{\textbf{Radar comparison across model groups.} We visualize representative models from proprietary general-purpose MLLMs, open-source general-purpose MLLMs, and specialized spatial MLLMs across the 15 subtasks in ViSTR-Bench. Human performance is shown as a reference, highlighting the substantial gap between current MLLMs and human-level spatial-temporal reasoning.}
  \label{fig:5}
\end{figure*}

\subsection{Main Results}
\label{sec:main_results}

Table~\ref{tab:main_results} and Figure~\ref{fig:5} report the comprehensive evaluation results on ViSTR-Bench. We summarize the main observations as follows.

\textbf{Current MLLMs remain far from human-level spatial-temporal reasoning.}
Existing MLLMs show limited performance on ViSTR-Bench.
The best-performing model, GPT-5.4-thinking, achieves an overall accuracy of 62.0\%, exceeding Chance Level (Frequency) by 4.1 percentage points while remaining 29.0 percentage points below the human performance of 91.0\%.
Notably, only three evaluated models outperform the frequency-based baseline, while all open-source general-purpose and specialized spatial MLLMs remain below it.
These results indicate that qualitative spatial-temporal reasoning from continuous visual cues remains a major challenge for current MLLMs, which still struggle to reliably leverage dynamic visual evidence for robust reasoning.

\textbf{Proprietary general-purpose MLLMs perform best overall.}
Among all evaluated model groups, proprietary general-purpose MLLMs achieve the best overall results.
GPT-5.4-thinking and Seed-2.0-Pro-thinking rank first and second, with accuracies of 62.0\% and 60.1\%, respectively, followed by Seed-2.0-Lite-thinking at 58.8\% and Seed-2.0-Pro at 56.8\%.
In contrast, open-source general-purpose MLLMs generally perform worse, with most models achieving accuracies between 48.0\% and 55.0\%.
The best-performing open-source model, Qwen3.5-27B-thinking, reaches 55.0\% but still trails GPT-5.4-thinking by 7.0 percentage points.
These results indicate that proprietary models currently hold a clear advantage in continuous spatial-temporal reasoning, although substantial room for improvement remains across all model groups.

\textbf{Spatial MLLMs show limited generalization to dynamic reasoning.}
Although specialized spatial MLLMs are designed to enhance spatial intelligence, they do not show clear advantages on ViSTR-Bench.
The best model in this group, GeoThinker-Qwen2.5VL-7B, achieves only 52.8\% overall accuracy, which is close to random chance and lower than many general-purpose MLLMs.
Moreover, we observe that several specialized spatial MLLMs exhibit biased single-option prediction patterns on certain tasks.
As a result, their performance often matches Chance Level (Frequency) or its complementary value.
This suggests that spatial capability learned from static or geometry-centric training datasets does not directly transfer to dynamic spatial-temporal reasoning.
ViSTR-Bench requires models to aggregate temporal evidence, track state changes, and reason about future outcomes or physical interactions, which remain weakly captured by current spatial MLLMs.

\textbf{Thinking mode is helpful but not consistently reliable.}
For models that support built-in thinking modes, enabling thinking often improves overall performance.
For example, GPT-5.4 improves from 56.1\% to 62.0\%, and Seed-2.0-Pro improves from 56.8\% to 60.1\%.
However, the improvement is not universal.
Claude-Opus-4.6 slightly drops from 55.4\% to 54.7\% after enabling thinking, and the gains for MiMo-V2.5 are marginal.
This indicates that explicit reasoning mechanisms can benefit spatial-temporal reasoning, but current thinking modes are not consistently beneficial across models.

\textbf{Performance varies substantially across reasoning dimensions.}
Models generally perform better on Motion Perception and Spatial Relations, where several top models achieve accuracies above 65\% on individual subtasks.
In contrast, Outcome Prediction and Physical Dynamics remain more challenging, with many results close to chance level.
These tasks require anticipating future outcomes or reasoning about latent physical dependencies, rather than merely recognizing visible motion or spatial relations.
This capability imbalance suggests that current MLLMs are better at describing observed dynamics than at performing temporally grounded prediction and intuitive physical reasoning.

\begin{table*}[!t]
    \centering
    \caption{
        \textbf{Effect of text-based CoT prompting on ViSTR-Bench.}
        We evaluate Gemini-3.1-Pro-Preview~\cite{Gemini} with {\normalfont\itshape Direct Prompting} and various text-based CoT prompting strategies.
        {\normalfont\itshape Avg.} denotes the overall accuracy (\%) averaged over all question-answer pairs, and $\Delta$ denotes the accuracy change relative to {\normalfont\itshape Direct Prompting}.
        The best and second-best results are highlighted with \bestcap{} and \secondcap{}, respectively.
    }
    \label{tab:text_cot}

    \resizebox{\textwidth}{!}{
    \begin{tabular}{l c c ccc ccc cccccc ccc}
        \toprule
        \multirow{2}{*}{\textbf{Prompting Method}} 
        & \multirow{2}{*}{\textbf{Avg.}} 
        & \multirow{2}{*}{$\boldsymbol{\Delta}$} 
        & \multicolumn{3}{c}{\textbf{Moti. Perc.}} 
        & \multicolumn{3}{c}{\textbf{Spat. Rela.}} 
        & \multicolumn{6}{c}{\textbf{Outc. Pred.}}  
        & \multicolumn{3}{c}{\textbf{Phys. Dyna.}} \\
        \cmidrule(lr){4-6} \cmidrule(lr){7-9} \cmidrule(lr){10-15} \cmidrule(lr){16-18}
        & & 
        & \makecell{Veh.\\Mov.} & \makecell{Rel.\\Vel.} & \makecell{Rot.\\Dir.}
        & \makecell{Ego\\Mot.} & \makecell{Pas.\\Fea.} & \makecell{Int.\\Dir.}
        & \makecell{Bas.\\Shot} & \makecell{Soc.\\Shot} & \makecell{Golf\\Shot} & \makecell{Bil.\\Shot} & \makecell{Swi.\\Race} & \makecell{Fall\\Dir.}
        & \makecell{Jenga\\Sta.} & \makecell{Mik.\\Dep.} & \makecell{Knot\\Type} \\
        \midrule

        \rowcolor{groupbg}
        \multicolumn{18}{l}{\textbf{Gemini-3.1-Pro-Preview}} \\
        Direct Prompting & 53.6 & - & \second{45.5} & 52.4 & 50.3 & 72.5 & \best{65.5} & 73.2 & \second{56.5} & 50.0 & 43.4 & \best{54.7} & \second{50.0} & 39.1 & 42.7 & 47.0 & \best{75.9} \\
        Zero-shot CoT & 54.5 & +0.9 & 41.1 & \second{56.0} & \best{55.2} & \second{74.0} & 58.2 & \second{75.0} & \best{57.3} & \second{52.5} & \best{50.9} & \second{46.7} & 48.6 & \best{54.3} & 42.7 & 49.4 & 65.5 \\
        Self-Consistency & 54.6 & +1.0 & 41.1 & 52.4 & 52.4 & \second{74.0} & \second{63.6} & \best{78.6} & 55.6 & \best{53.2} & 47.2 & \second{46.7} & \best{56.9} & 43.5 & 47.9 & 48.2 & \second{69.0} \\
        Plan-and-Solve & \second{54.9} & \second{+1.3} & \second{45.5} & 50.0 & \second{54.5} & \best{76.3} & 61.8 & \second{75.0} & 54.0 & 51.3 & \second{49.1} & 44.0 & 45.8 & \second{47.8} & \second{53.8} & \second{50.6} & \second{69.0} \\
        Manual CoT & \best{55.2} & \best{+1.6} & \best{46.4} & \best{59.5} & 51.0 & 71.8 & 54.5 & \second{75.0} & 51.6 & \second{52.5} & 47.2 & 42.7 & \best{56.9} & 41.3 & \best{58.1} & \best{53.0} & \best{75.9} \\
        \bottomrule
    \end{tabular}
    }
\end{table*}

\begin{table*}[!t]
    \centering
    \caption{
        \textbf{Effect of visual input format on ViSTR-Bench.}
        We evaluate Gemini-3.1-Pro-Preview~\cite{Gemini} with different visual input formats.
        {\normalfont\itshape Avg.} denotes the overall accuracy (\%) averaged over all question-answer pairs, and $\Delta$ denotes the accuracy change relative to {\normalfont\itshape Text-only}.
        The best and second-best results are highlighted with \bestcap{} and \secondcap{}, respectively.
    }
    \label{tab:visual_input}

    \resizebox{\textwidth}{!}{
    \begin{tabular}{l c c ccc ccc cccccc ccc}
        \toprule
        \multirow{2}{*}{\textbf{Input Format}} 
        & \multirow{2}{*}{\textbf{Avg.}} 
        & \multirow{2}{*}{$\boldsymbol{\Delta}$} 
        & \multicolumn{3}{c}{\textbf{Moti. Perc.}} 
        & \multicolumn{3}{c}{\textbf{Spat. Rela.}} 
        & \multicolumn{6}{c}{\textbf{Outc. Pred.}}  
        & \multicolumn{3}{c}{\textbf{Phys. Dyna.}} \\
        \cmidrule(lr){4-6} \cmidrule(lr){7-9} \cmidrule(lr){10-15} \cmidrule(lr){16-18}
        & & 
        & \makecell{Veh.\\Mov.} & \makecell{Rel.\\Vel.} & \makecell{Rot.\\Dir.}
        & \makecell{Ego\\Mot.} & \makecell{Pas.\\Fea.} & \makecell{Int.\\Dir.}
        & \makecell{Bas.\\Shot} & \makecell{Soc.\\Shot} & \makecell{Golf\\Shot} & \makecell{Bil.\\Shot} & \makecell{Swi.\\Race} & \makecell{Fall\\Dir.}
        & \makecell{Jenga\\Sta.} & \makecell{Mik.\\Dep.} & \makecell{Knot\\Type} \\
        \midrule

        \rowcolor{groupbg}
        \multicolumn{18}{l}{\textbf{Gemini-3.1-Pro-Preview}} \\
        Text-only & 47.0 & - & \second{47.3} & \second{51.2} & 46.9 & 46.6 & 36.4 & 62.5 & 47.6 & 43.0 & \second{52.8} & \second{53.3} & 40.3 & 43.5 & 43.6 & 45.8 & 58.6 \\
        Last Frame & 51.7 & +4.7 & \best{48.2} & 45.2 & \best{57.2} & 54.2 & 45.5 & 64.3 & \second{50.0} & \second{55.7} & 49.1 & 48.0 & \best{58.3} & \best{50.0} & \second{44.4} & 45.8 & 65.5 \\
        Shuffled Frames & 51.9 & +4.9 & \best{48.2} & \second{51.2} & 45.5 & 60.3 & 56.4 & 64.3 & 49.2 & \best{56.3} & \best{62.3} & 42.7 & 48.6 & \second{45.7} & \second{44.4} & \best{50.6} & \second{72.4} \\
        Ordered Frames & \second{52.0} & \second{+5.0} & 46.4 & 47.6 & \second{55.2} & \second{67.2} & \second{58.2} & \second{67.9} & 49.2 & 55.1 & 49.1 & \second{53.3} & 47.2 & 30.4 & \best{48.7} & 36.1 & 62.1 \\
        Original Video & \best{53.6} & \best{+6.6} & 45.5 & \best{52.4} & 50.3 & \best{72.5} & \best{65.5} & \best{73.2} & \best{56.5} & 50.0 & 43.4 & \best{54.7} & \second{50.0} & 39.1 & 42.7 & \second{47.0} & \best{75.9} \\
        \bottomrule
    \end{tabular}
    }
\end{table*}

\begin{figure*}[!t]
  \centering
  \includegraphics[width=0.9\linewidth]{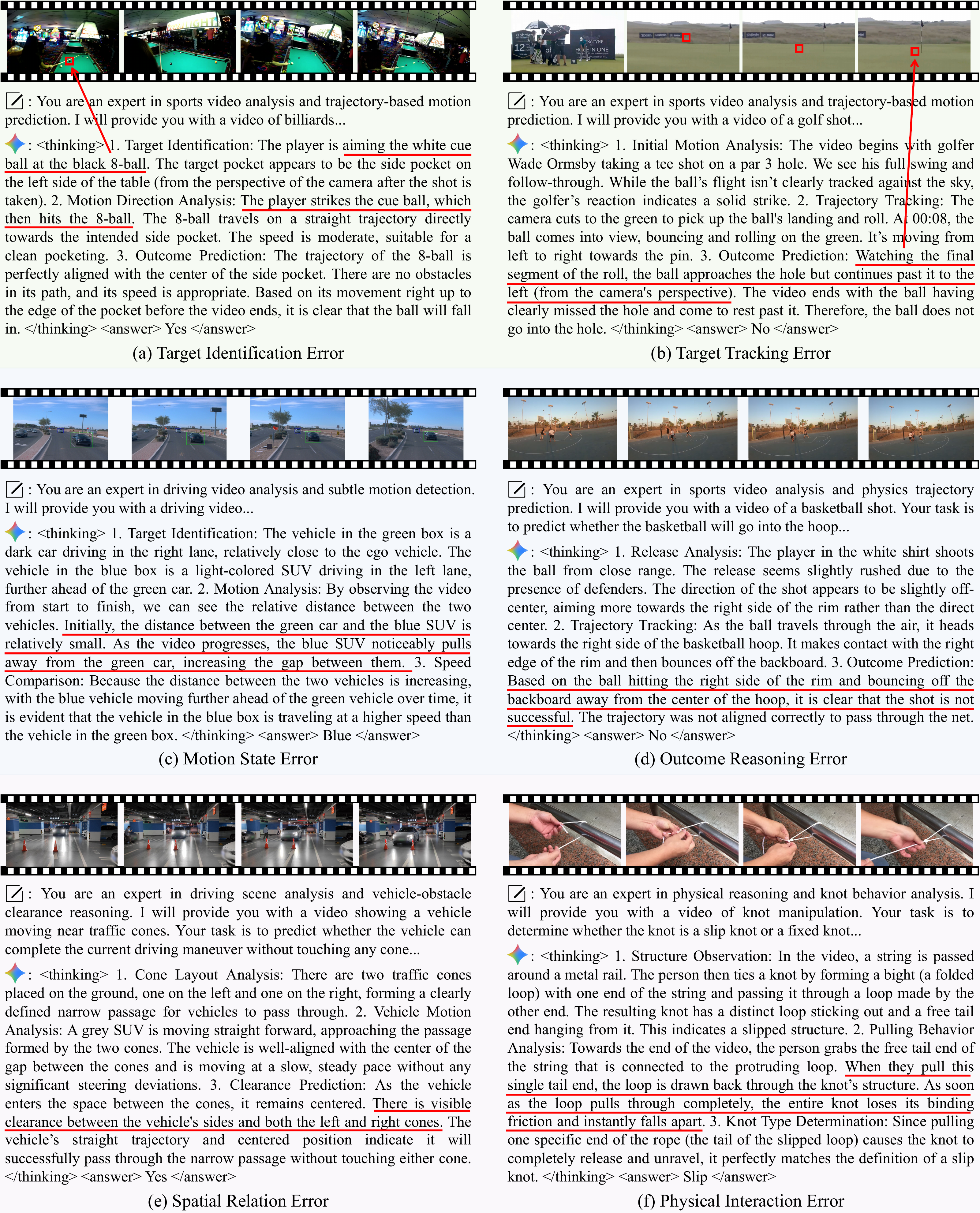}
  \caption{\textbf{Representative examples for each primary error type.}}
  \label{fig:18}
\end{figure*}

\subsection{Effect of Text-based CoT Prompting}
\label{sec:text_cot}

We further investigate whether explicit text-based Chain-of-Thought (CoT) prompting can enhance spatial-temporal reasoning on ViSTR-Bench.
Using Gemini-3.1-Pro-Preview as a representative video-based MLLM, we compare four text-based CoT prompting strategies against Direct Prompting, namely Zero-shot CoT~\cite{zero-shot-cot}, Self-Consistency~\cite{self-consistency-coT}, Plan-and-Solve~\cite{plan-and-solve}, and Manual CoT inspired by OmniSpatial~\cite{OmniSpatial}.
Specifically, Zero-shot CoT appends a generic step-by-step reasoning instruction without task-specific design.
Self-Consistency samples multiple reasoning paths by running the Zero-shot CoT prompt over five independent trials and aggregates the final predictions via majority voting.
Plan-and-Solve instructs the model to first formulate a reasoning plan and then execute it to solve the problem.
Manual CoT decomposes each task into human-crafted, task-specific analysis steps tailored to its underlying spatial-temporal reasoning requirement. 
Detailed prompt templates for Manual CoT are provided in Appendix~\ref{sec:detailed_prompt_templates}.

As reported in Table~\ref{tab:text_cot}, text-based CoT prompting brings only limited improvements on ViSTR-Bench.
Zero-shot CoT slightly improves the overall accuracy from 53.6\% to 54.5\% compared with Direct Prompting, suggesting that simply eliciting generic step-by-step reasoning can provide some benefit but remains insufficient for complex video-based spatial-temporal reasoning.
Self-Consistency and Plan-and-Solve further improve the overall accuracy over Direct Prompting by 1.0\% and 1.3\%, respectively.
This suggests that aggregating multiple reasoning paths or imposing a more structured reasoning process can help the model to some extent, but the gains remain modest.
Manual CoT achieves the best overall performance, improving Direct Prompting by 1.6\% and reaching 55.2\%.
By decomposing each task into task-specific reasoning steps, Manual CoT provides stronger guidance for grounding textual reasoning in relevant visual evidence.
Nevertheless, the improvement is still moderate and highly task-dependent.
Moreover, even the best CoT result remains close to the chance-level baselines, indicating that text-based reasoning prompts alone cannot fully compensate for the limitations of current MLLMs in accurate visual perception and temporal grounding.

\subsection{Effect of Visual Input Format}
\label{sec:visual_input}

We further study how different visual input formats affect model performance on ViSTR-Bench.
Using Gemini-3.1-Pro-Preview as the evaluated model, we compare five input settings: text-only input, the last frame only, shuffled frames with disrupted temporal order, ordered frames uniformly sampled from the video, and the original video input.
As reported in Table~\ref{tab:visual_input}, all visual input formats consistently outperform the text-only setting, showing that visual evidence provides useful cues beyond language priors.
Multi-frame inputs bring slightly larger gains, with shuffled frames and ordered frames improving over the last-frame setting by 0.2\% and 0.3\%, respectively, suggesting that the model can benefit from observing multiple visual states.
However, ordered frames only slightly outperform shuffled frames in overall accuracy, indicating that the model does not reliably exploit the continuous temporal order among frames.
This suggests that current MLLMs may capture coarse temporal-related cues from multiple frames, but still struggle to reason over continuous visual dynamics.

To further verify this observation, we conduct a small diagnostic study on Basketball Shot, a representative outcome prediction task.
Specifically, we select 20 samples and re-edit each video to retain the complete shooting process, including the final shot outcome.
We then provide humans and Gemini-3.1-Pro-Preview with the first 50\%, 75\%, and 100\% of each re-edited video, where the 100\% setting contains the final shot outcome and thus no longer requires prediction.
Human accuracy increases monotonically from 80\% to 90\% and then to 100\%, showing that the additional temporal evidence is informative.
In contrast, Gemini-3.1-Pro-Preview obtains 50\%, 65\%, and 60\%, respectively, suggesting that the model still fails to reliably use continuous outcome-revealing cues even when more of the video is available.

Original video input achieves the best performance, improving over the text-only setting by 6.6\%.
These results indicate that video input provides the most complete visual evidence, while effective reasoning from continuous visual cues remains challenging for current MLLMs.

\subsection{Error Analysis}
\label{sec:error_analysis}

To better understand why current MLLMs exhibit limited performance on ViSTR-Bench, we manually inspect the 600 incorrect predictions made by Gemini-3.1-Pro-Preview under the Manual CoT setting.
For each case, we assign a single primary error type according to the task definition, the ground-truth answer, and the reasoning steps generated by the model following the Manual CoT prompt.
We organize the failures along a general reasoning process: grounding the target, following its temporal evolution, estimating motion or spatial states, and finally drawing outcome-level or physical conclusions.
This leads to six task-agnostic error types, with representative examples shown in Figure~\ref{fig:18}.

\noindent\textbf{Target Identification Error} occurs when the model fails at the initial grounding stage and identifies the wrong target specified by the question.
Typical cases include confusing the target ball in Billiards, mixing up lane numbers in Swimming Race, or localizing the wrong Jenga block or Mikado stick.

\noindent\textbf{Target Tracking Error} occurs after the target has been identified, but the model cannot consistently follow it across frames.
For example, the model may lose track of a moving ball, confuse its position with nearby objects, or fail to associate the same target across consecutive observations.

\noindent\textbf{Motion State Error} refers to failures in estimating basic motion attributes of the target.
In these cases, the model observes the relevant object but misjudges whether it is moving, which object is faster, or which direction a person is rotating.

\noindent\textbf{Spatial Relation Error} captures mistakes in reasoning about geometric relations among the target, the observer, and reference objects.
Representative cases include reversing observer-centric directions in Ego Motion, misjudging whether a vehicle has sufficient clearance to pass through a constrained region, or confusing the movement direction of a person relative to an interacted object.

\noindent\textbf{Outcome Reasoning Error} occurs when the model captures some current motion or intermediate state but fails to extrapolate it to the correct final outcome.
Typical examples include incorrectly predicting whether a shot will succeed, which swimmer will reach the finish line first, or the final falling direction of a person.

\noindent\textbf{Physical Interaction Error} denotes failures in inferring latent physical properties, dependencies, or stability conditions from dynamic evidence.
For example, the model may misjudge whether a Mikado stick is blocked by another stick, whether a Jenga block is load-bearing, or whether a specific knot can be released by pulling one end.

\begin{table}[!t]
    \centering
    \caption{
        \textbf{Distribution of primary error types.}
        We analyze the 600 incorrect predictions made by Gemini-3.1-Pro-Preview~\cite{Gemini} under the Manual CoT setting.
        Each incorrect prediction is assigned one primary error type.
    }
    \label{tab:error_distribution}
    \small
    \begin{tabular}{lrr}
        \toprule
        \textbf{Primary Error Type} & \textbf{\#Errors} & \textbf{Ratio} \\
        \midrule
        Motion State Error & 165 & 27.5\% \\
        Outcome Reasoning Error & 147 & 24.5\% \\
        Target Tracking Error & 110 & 18.3\% \\
        Physical Interaction Error & 95 & 15.8\% \\
        Spatial Relation Error & 76 & 12.7\%  \\
        Target Identification Error & 7 & 1.2\%  \\
        \midrule
        \textbf{Total} & \textbf{600} & \textbf{100.0\%} \\
        \bottomrule
    \end{tabular}
\end{table}

Table~\ref{tab:error_distribution} reports the resulting distribution.
\emph{Motion State Error} is the most frequent category, suggesting that current MLLMs still struggle to determine whether a prompted object is moving and to compare or classify motion-related attributes such as relative speed and rotation direction.
\emph{Outcome Reasoning Error} is the second-largest category, indicating that models often fail to extrapolate observed motion or intermediate states to the correct final outcome.
\emph{Target Tracking Error} ranks third and reflects failures in maintaining temporally consistent evidence, especially in ball-centric outcome prediction tasks.
\emph{Physical Interaction Error} is mainly observed in Jenga Stability, Mikado Dependency, and Knot Type, showing limited ability to infer latent support relations, object dependencies, and stability conditions.
\emph{Spatial Relation Error} mainly arises in Ego Motion, Passage Feasibility, and Interaction Direction, where the model reverses observer-centric directions, misjudges spatial clearance, or confuses relative movement directions.
By contrast, \emph{Target Identification Error} accounts for only 7 cases, mostly occurring in Billiards.
Overall, these results suggest that the main bottlenecks of current MLLMs on ViSTR-Bench are not simple failures to recognize target objects, but failures to track dynamic evidence, estimate motion and spatial states, extrapolate outcomes, and reason about latent physical interactions.

\subsection{Preliminary Exploration on Model Improvement}
\label{sec:model_improvement}

\begin{figure*}[!t]
  \centering
  \subfloat[]{
    \includegraphics[height=0.23\linewidth]{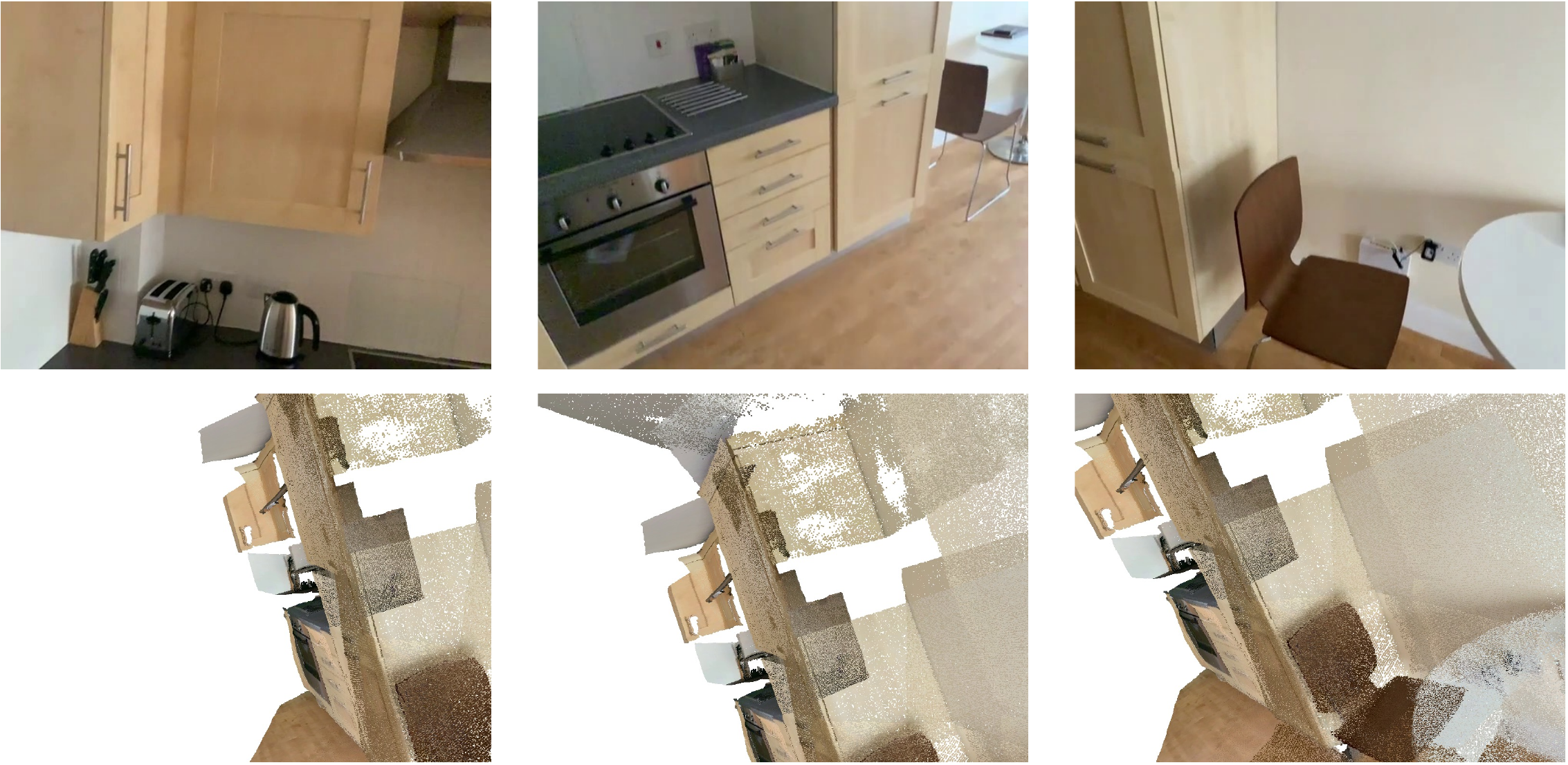}
    \label{fig:37_a}
  }
  \hfill
  \subfloat[]{
    \includegraphics[height=0.23\linewidth]{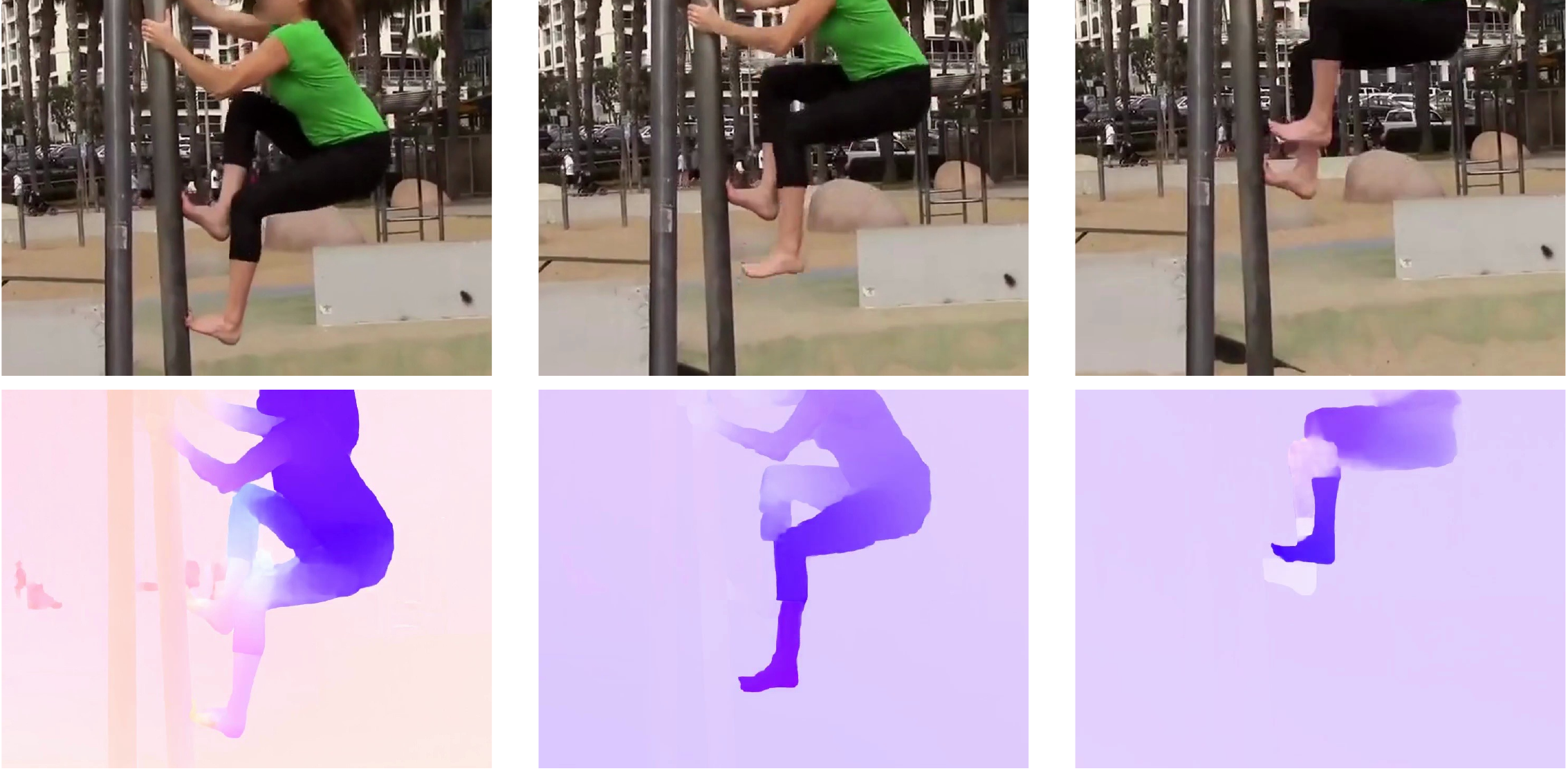}
    \label{fig:37_b}
  }
  \caption{\textbf{Preliminary exploration on model improvement.}
  We examine two complementary strategies for making task-relevant visual evidence more explicit.
  (a) For input-centric augmentation on Ego Motion, we reconstruct the scene using VGGT-$\Omega$~\cite{VGGT-Omega} and render auxiliary novel views to provide explicit 3D spatial evidence.
  (b) For tool-augmented reasoning on Interaction Direction, we use WAFT~\cite{WAFT} to extract optical-flow cues and follow Perception Program~\cite{PerceptionProgram} to convert them into compact, language-native summaries.
  The first row shows sampled frames from the original videos, while the second row shows rendered novel views in (a) and estimated optical-flow maps in (b).
  These pilot studies improve GPT-5.4~\cite{GPT-5} accuracy from 63.4\% to 80.2\% on Ego Motion and from 78.6\% to 83.9\% on Interaction Direction.}
  \label{fig:37}
\end{figure*}

The error analysis shows that current MLLMs still struggle to reason from continuous visual cues in dynamic scenes.
Based on these observations, we discuss several complementary directions for improving model performance.
First, a data-centric direction is to construct larger-scale training data specifically designed for dynamic spatial-temporal reasoning.
Although current MLLMs are trained on massive image-text and video-text corpora, existing datasets~\cite{SPAR-Bench,LLaVA-Video} mainly focus on static visual recognition or general video understanding, and provide limited supervision for fine-grained dynamic reasoning.
Second, an input-centric direction is to enrich the visual input with more explicit spatial evidence beyond the original camera trajectory.
For example, auxiliary novel views can be rendered from the reconstructed scene and added to the original video, providing complementary geometric and layout cues that are not directly observed from the captured viewpoints.
Third, a tool-augmented direction is to build agentic systems that decompose complex video reasoning problems into intermediate subproblems and use external tools to extract task-relevant visual evidence.
For instance, detectors and trackers can localize target objects, optical flow can estimate motion patterns, and 3D reconstruction can recover scene geometry.
This paradigm is promising because it offloads low-level visual perception to specialized tools, allowing MLLMs to focus more on high-level spatial-temporal reasoning.

Figure~\ref{fig:37} illustrates our preliminary exploration of two of these directions on tasks where the corresponding visual evidence is particularly relevant, with GPT-5.4 as the baseline model.
For the input-centric direction, we study the Ego Motion task by enriching the original visual input with explicit 3D spatial evidence.
Specifically, we reconstruct the scene from the original video using VGGT-$\Omega$~\cite{VGGT-Omega} and render 10 auxiliary novel views to provide complementary layout and viewpoint cues beyond the sampled frames.
This augmentation improves accuracy from 63.4\% to 80.2\%, indicating that geometry-enhanced inputs can make task-relevant spatial cues more accessible to current MLLMs.
For the tool-augmented direction, we study the Interaction Direction task by extracting explicit motion evidence with external tools.
We use WAFT~\cite{WAFT} to generate optical-flow maps and follow Perception Program~\cite{PerceptionProgram} to convert the tool outputs into compact, structured, language-native summaries that MLLMs can directly parse and reason over.
This pipeline improves accuracy from 78.6\% to 83.9\%, suggesting that explicit motion evidence can help models better capture interaction patterns and infer relative movement directions.
Rather than providing a complete solution, these pilot studies demonstrate that making task-relevant spatial and motion evidence more explicit is a promising direction for improving the spatial-temporal reasoning capabilities of current MLLMs.

\section{Conclusion and Discussion}

\subsection{Conclusion}
In this paper, we introduced ViSTR-Bench, a visual spatial-temporal reasoning benchmark designed to evaluate whether MLLMs can reason from continuous visual cues in dynamic scenes. 
Unlike prior benchmarks that mainly focus on static spatial attributes, low-level temporal perception, or quantitative prediction, ViSTR-Bench emphasizes temporal evidence, reasoning-oriented task design, and qualitative evaluation. It covers four complementary dimensions, with 15 subtasks and 1,340 high-quality video question-answer pairs across tabletop, indoor, and outdoor scenarios. Through comprehensive evaluations of proprietary, open-source, and specialized spatial MLLMs, we find that current models remain substantially below human performance. We hope ViSTR-Bench can serve as a diagnostic testbed for future research on physically grounded multimodal intelligence, embodied AI, and dynamic world modeling.

\subsection{Limitations and Future Work}
\label{sec:limitations_future_work}

While ViSTR-Bench provides a diagnostic testbed for visual spatial-temporal reasoning, it also has several limitations that motivate future research.

\noindent\textbf{Task and data coverage.}
ViSTR-Bench covers 15 subtasks across tabletop, indoor, and outdoor scenarios, but it is not intended to exhaust all forms of spatial-temporal reasoning in the physical world.
The current benchmark focuses on relatively short clips with clearly specified targets and binary decisions.
In contrast, many real-world embodied tasks require longer-horizon planning, multi-agent interaction, and adaptation to continuously changing goals.
Future work can extend ViSTR-Bench to broader scene categories, longer temporal contexts, and more interactive scenarios, enabling evaluation under settings closer to robotics, autonomous driving, and embodied AI applications.

\noindent\textbf{Evaluation format.}
All tasks in ViSTR-Bench are formulated as qualitative binary-choice questions.
This design reduces annotation ambiguity, mitigates the impact of minor numerical errors, and enables controlled comparisons across different MLLMs.
However, binary accuracy only measures the final decision and cannot fully capture graded spatial-temporal understanding or the correctness of intermediate reasoning steps.
Future benchmarks may combine qualitative decisions with richer answer formats, such as open-ended explanations or structured annotations of object states, spatial relations, and physical dependencies.

\noindent\textbf{Model improvement.}
Our experiments and error analysis show that current MLLMs mainly struggle with target tracking, outcome extrapolation, spatial relation estimation, and latent physical interaction reasoning.
The preliminary exploration in Sec.~\ref{sec:model_improvement} suggests that richer spatial evidence can provide useful cues, but it remains only an initial step toward improving dynamic scene reasoning.
Promising directions include constructing larger training corpora for dynamic spatial-temporal reasoning, integrating external perception tools such as object trackers, optical flow estimators, and 3D reconstruction modules, and developing world-modeling mechanisms that explicitly maintain object states and physical relations over time.
We hope ViSTR-Bench can support these directions by serving not only as an evaluation benchmark, but also as a diagnostic resource for identifying underdeveloped components of dynamic visual reasoning.


\bibliographystyle{IEEEtran}
\bibliography{main}

\vfill


\clearpage
\appendices

\section{Detailed Data Sources}
\label{sec:detailed_data_sources}

In this section, we provide a detailed description of each data source and its corresponding usage in ViSTR-Bench.

\textbf{Waymo Open Dataset}~\cite{Waymo} is a large-scale autonomous driving dataset comprising 1,150 driving segments. Each segment lasts approximately 20 seconds and provides synchronized multi-view videos with rich annotations for traffic agents, including vehicles, pedestrians, and cyclists. The dataset is widely used for autonomous driving perception tasks such as 3D object detection, object tracking, and motion forecasting.
In ViSTR-Bench, we use videos from the Waymo Open Dataset~\cite{Waymo} for driving-related motion perception tasks, including Vehicle Movement and Relative Velocity. We also leverage the available vehicle annotations to localize target vehicles and overlay visual prompts on the corresponding video frames.

\textbf{ScanNet}~\cite{ScanNet} is a richly annotated RGB-D dataset of real-world indoor scenes, comprising more than 1,500 indoor scans and over 2.5 million RGB-D views. The dataset covers diverse indoor environments, such as bedrooms, offices, and living rooms. ScanNet is widely used for 3D reconstruction, semantic segmentation, and instance segmentation.
In ViSTR-Bench, we use ScanNet~\cite{ScanNet} as one of the data sources for the Ego Motion task. Given an indoor egocentric video and a target object, the model is required to infer the final spatial relationship between the observing camera and the target object.

\textbf{ScanNet++}~\cite{ScanNet++} is an extension of ScanNet~\cite{ScanNet}, offering more detailed geometry, denser visual observations, and more complete scene coverage.
It contains 460 indoor scenes, covering more than 1,000 object categories and over 21,000 object instances.
In ViSTR-Bench, we use ScanNet++~\cite{ScanNet++} as an additional data source for the Ego Motion task.

\textbf{ARKitScenes}~\cite{ARKitScenes} is a large-scale indoor RGB-D dataset captured using mobile devices equipped with LiDAR sensors. It contains 5,047 captures from 1,661 unique indoor scenes. ARKitScenes~\cite{ARKitScenes} is commonly used for 3D reconstruction, depth estimation, and indoor object detection.
In ViSTR-Bench, we use ARKitScenes~\cite{ARKitScenes} as another data source for the Ego Motion task.

\textbf{Ego4D}~\cite{Ego4D} is a large-scale egocentric video dataset that records unscripted daily human activities from a first-person perspective. It contains more than 3,700 hours of video collected by 926 participants across 74 locations in nine countries. The dataset covers diverse activities in homes, workplaces, outdoor environments, and recreational scenes. Ego4D~\cite{Ego4D} is widely used for egocentric video understanding, activity recognition, and human-object interaction analysis.
In ViSTR-Bench, we use Ego4D~\cite{Ego4D} as a public data source for outcome prediction tasks, including Basketball Shot and a subset of Billiards Shot examples.

\textbf{Motion-X}~\cite{MotionX} is a large-scale 3D expressive whole-body human motion dataset comprising 81.1K motion sequences and 15.6 million precise 3D whole-body pose annotations represented using SMPL-X~\cite{SMPL-X}. 
The dataset is constructed from large-scale online videos and eight existing motion datasets and provides expressive whole-body annotations covering body movements, hand gestures, and facial expressions. 
Motion-X~\cite{MotionX} supports tasks such as text-driven whole-body motion generation and 3D whole-body human mesh recovery. 
In ViSTR-Bench, we use Motion-X~\cite{MotionX} as a data source for human-centric spatial-temporal reasoning tasks, including Rotation Direction, Interaction Direction, and Fall Direction.

\section{Detailed Prompt Templates}
\label{sec:detailed_prompt_templates}

\subsection{Direct Prompting}

Table~\ref{tab:direct_prompt_templates} presents the Direct Prompting templates used in ViSTR-Bench.
Each prompt specifies the visual context, the reasoning target, and the required answer format for the corresponding subtask.
Placeholders such as \texttt{\{TARGET\}}, \texttt{\{PERSPECTIVE\}}, \texttt{\{OPTION\_A\}}, and \texttt{\{OPTION\_B\}} are replaced with instance-specific metadata during QA pair generation.
Most subtasks adopt fixed binary answer spaces, including Yes/No for judgment tasks, Green/Blue for comparing prompted vehicles, Slip/Fixed for knot classification, and Clockwise/Counterclockwise for rotation direction.
For the remaining subtasks, binary choices are constructed from task-specific label spaces.
Ego Motion uses relative-position labels such as Front-left, Front-right, Back-left, and Back-right, Swimming Race uses lane numbers, while Interaction Direction and Fall Direction use directionality labels.

\subsection{Manual CoT}

\begin{figure*}[p]
  \centering
  \includegraphics[width=0.95\linewidth]{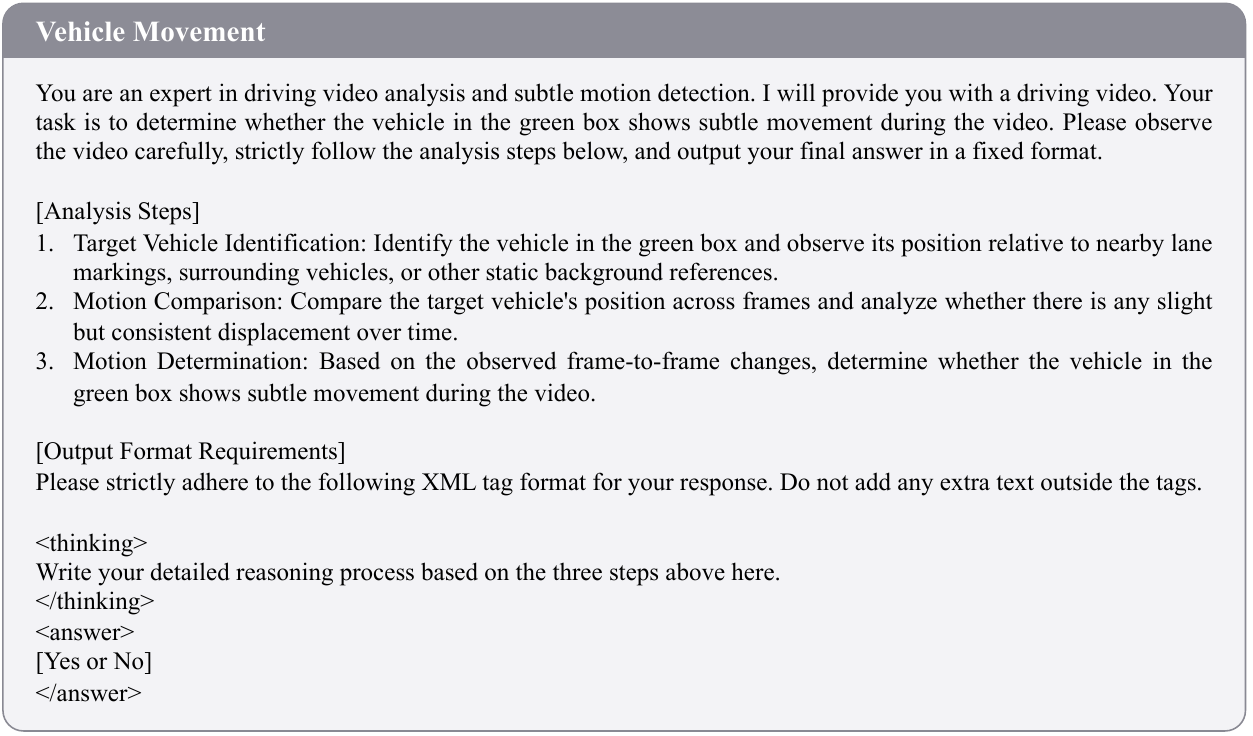}
  \caption{\textbf{Manual CoT prompt template for Vehicle Movement.}}
  \label{fig:6}
\end{figure*}

\begin{figure*}[p]
  \centering
  \includegraphics[width=0.95\linewidth]{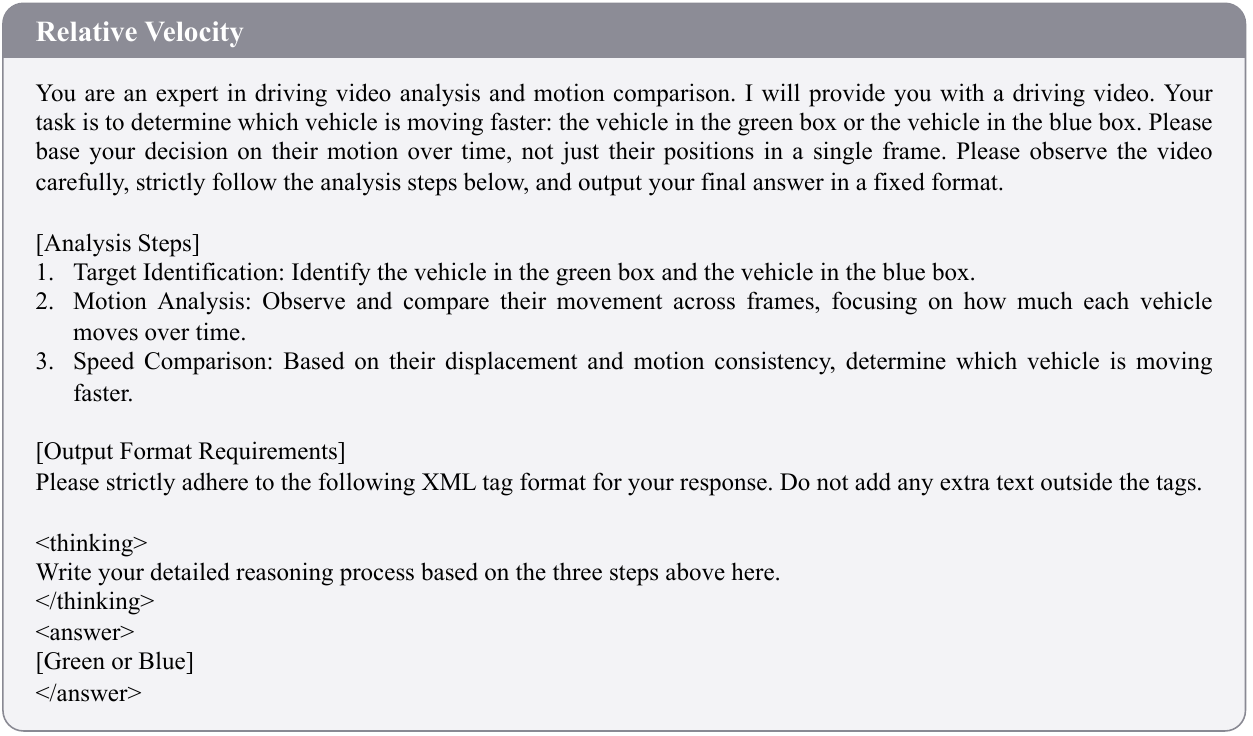}
  \caption{\textbf{Manual CoT prompt template for Relative Velocity.}}
  \label{fig:7}
\end{figure*}

\begin{figure*}[p]
  \centering
  \includegraphics[width=0.95\linewidth]{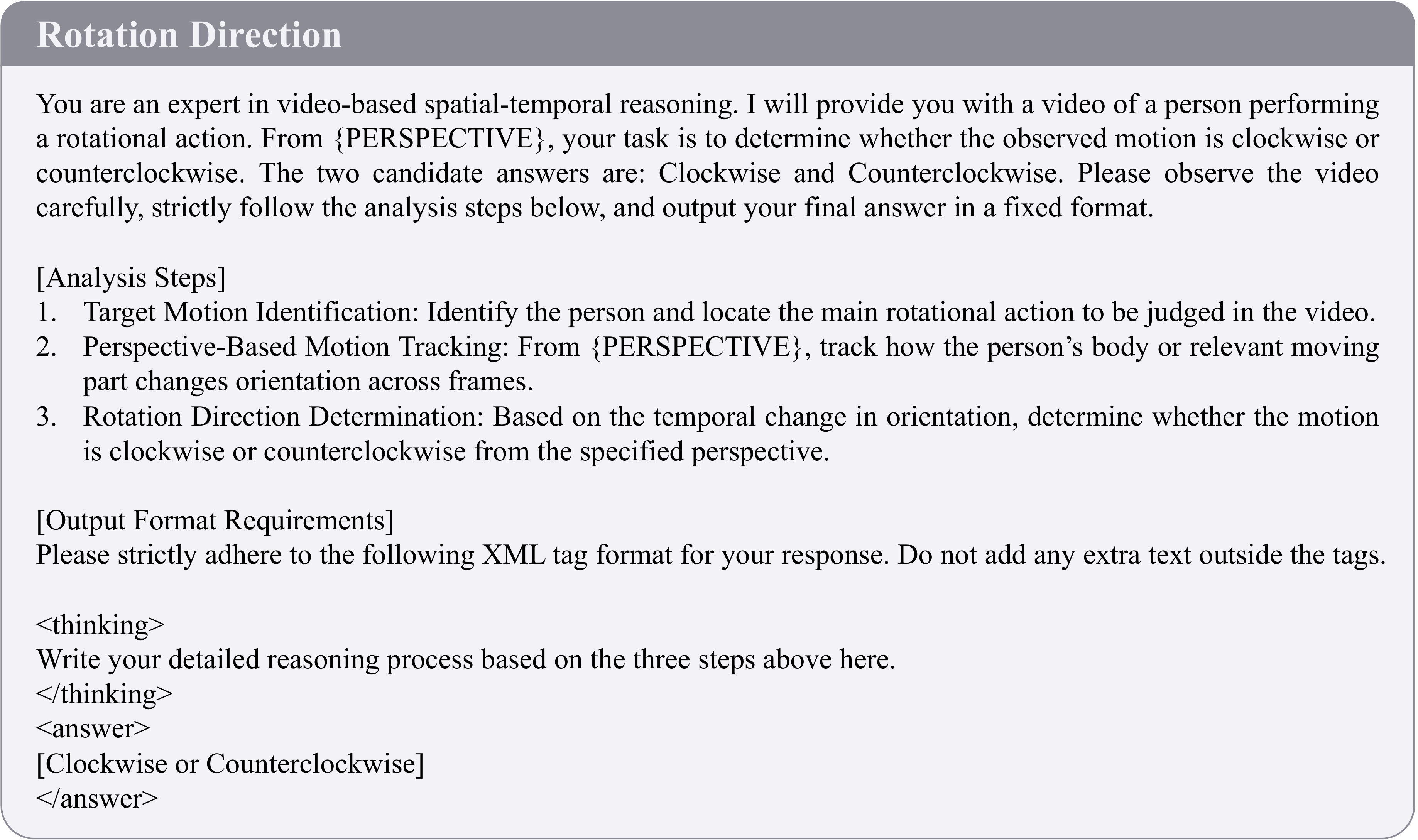}
  \caption{\textbf{Manual CoT prompt template for Rotation Direction.}}
  \label{fig:31}
\end{figure*}

\begin{figure*}[p]
  \centering
  \includegraphics[width=0.95\linewidth]{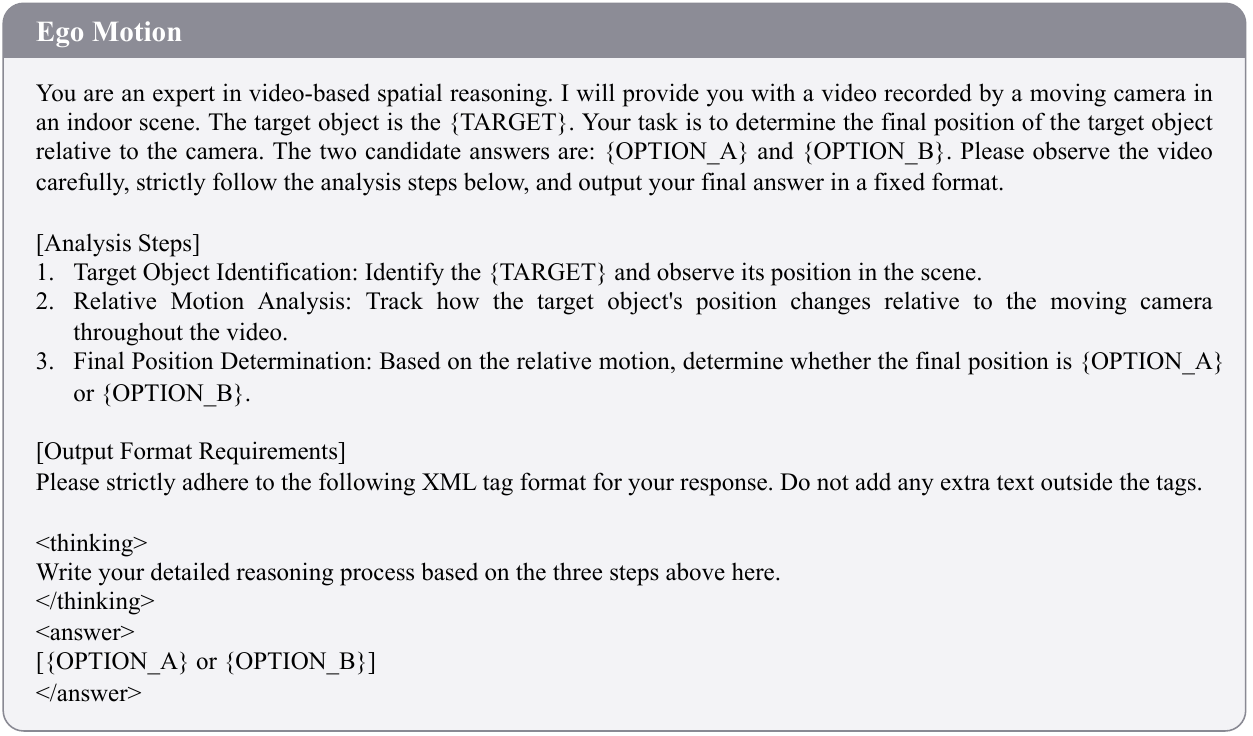}
  \caption{\textbf{Manual CoT prompt template for Ego Motion.}}
  \label{fig:8}
\end{figure*}

\begin{figure*}[p]
  \centering
  \includegraphics[width=0.95\linewidth]{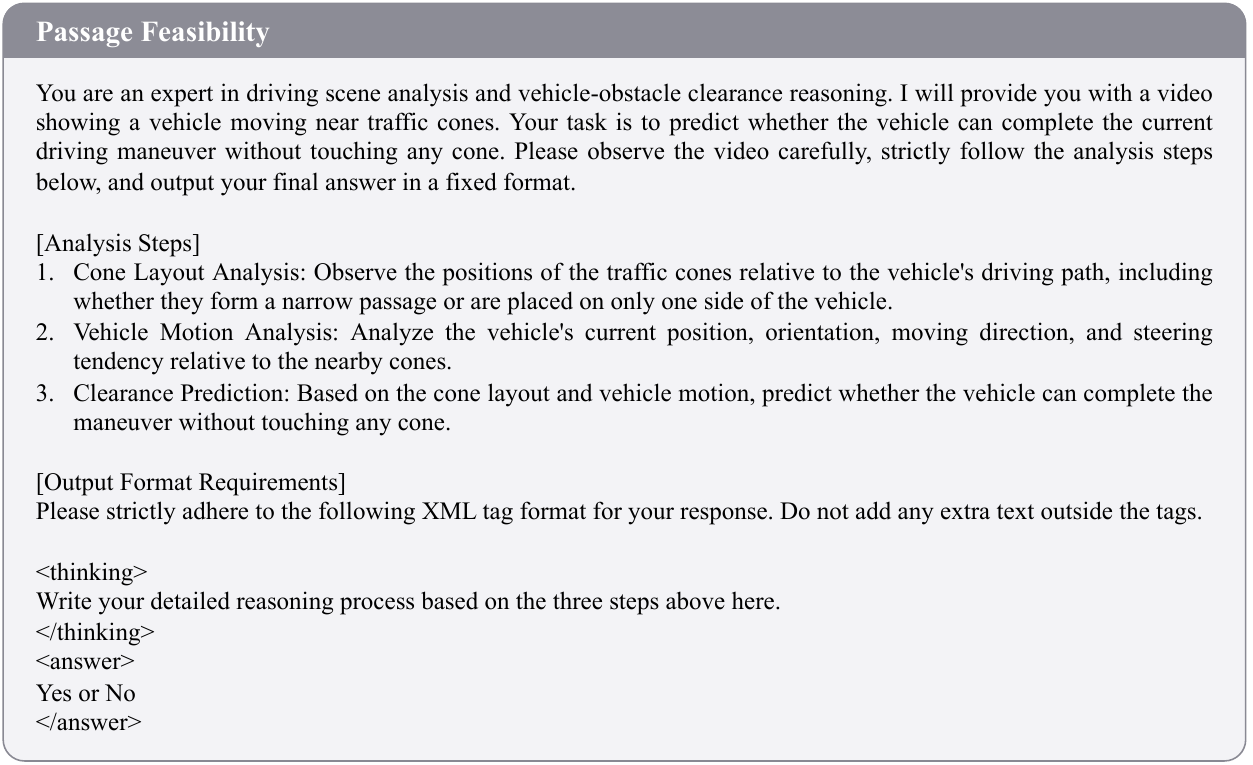}
  \caption{\textbf{Manual CoT prompt template for Passage Feasibility.}}
  \label{fig:9}
\end{figure*}

\begin{figure*}[p]
  \centering
  \includegraphics[width=0.95\linewidth]{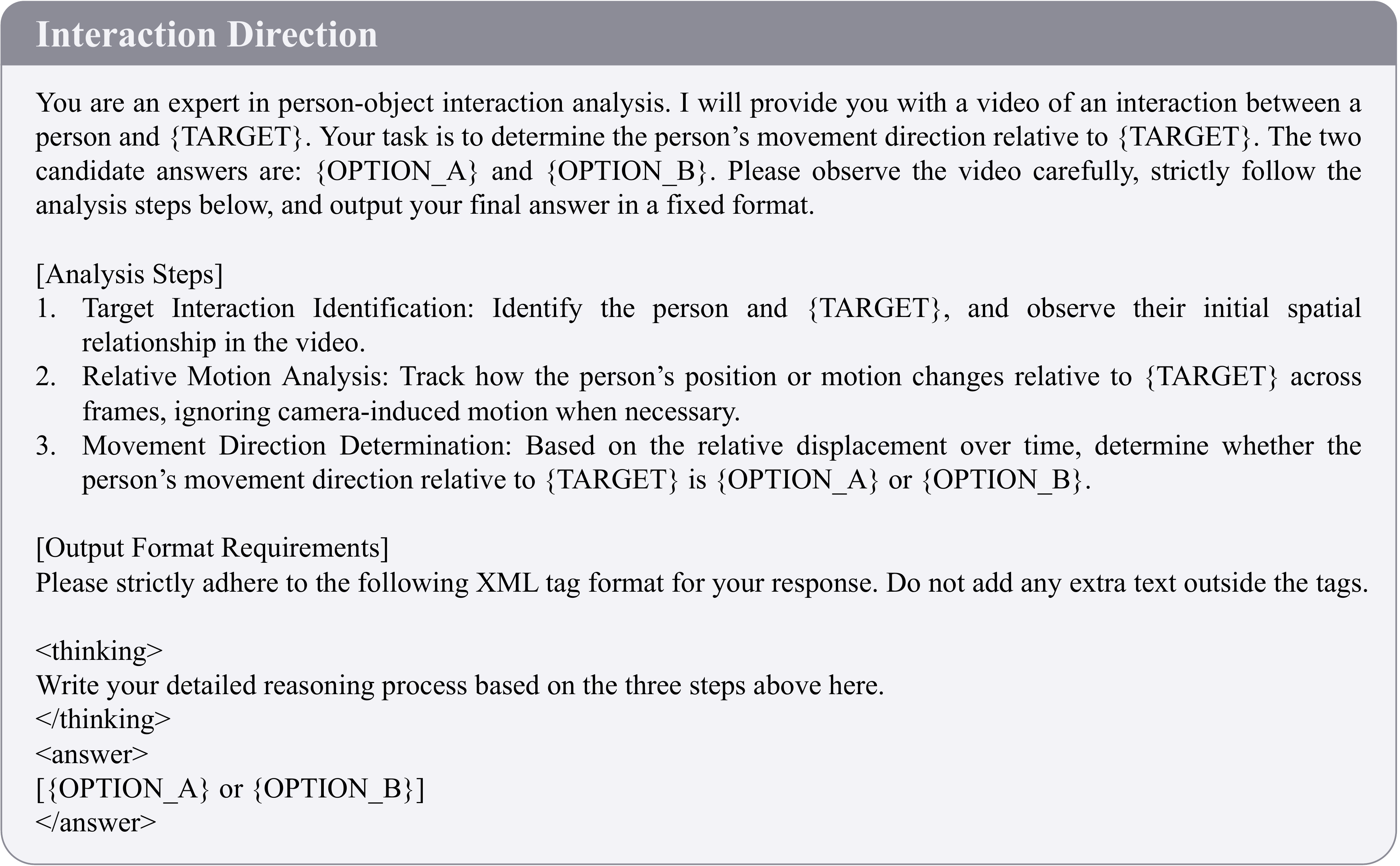}
  \caption{\textbf{Manual CoT prompt template for Interaction Direction.}}
  \label{fig:32}
\end{figure*}

\begin{figure*}[p]
  \centering
  \includegraphics[width=0.95\linewidth]{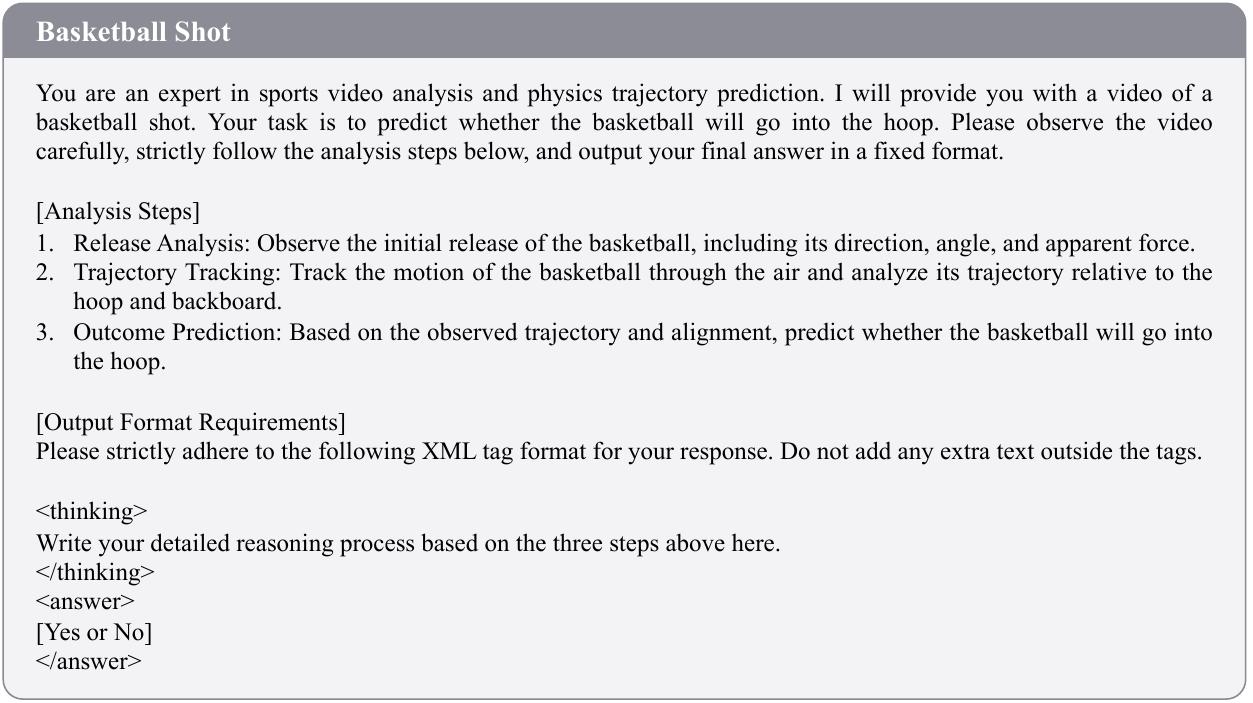}
  \caption{\textbf{Manual CoT prompt template for Basketball Shot.}}
  \label{fig:10}
\end{figure*}

\begin{figure*}[p]
  \centering
  \includegraphics[width=0.95\linewidth]{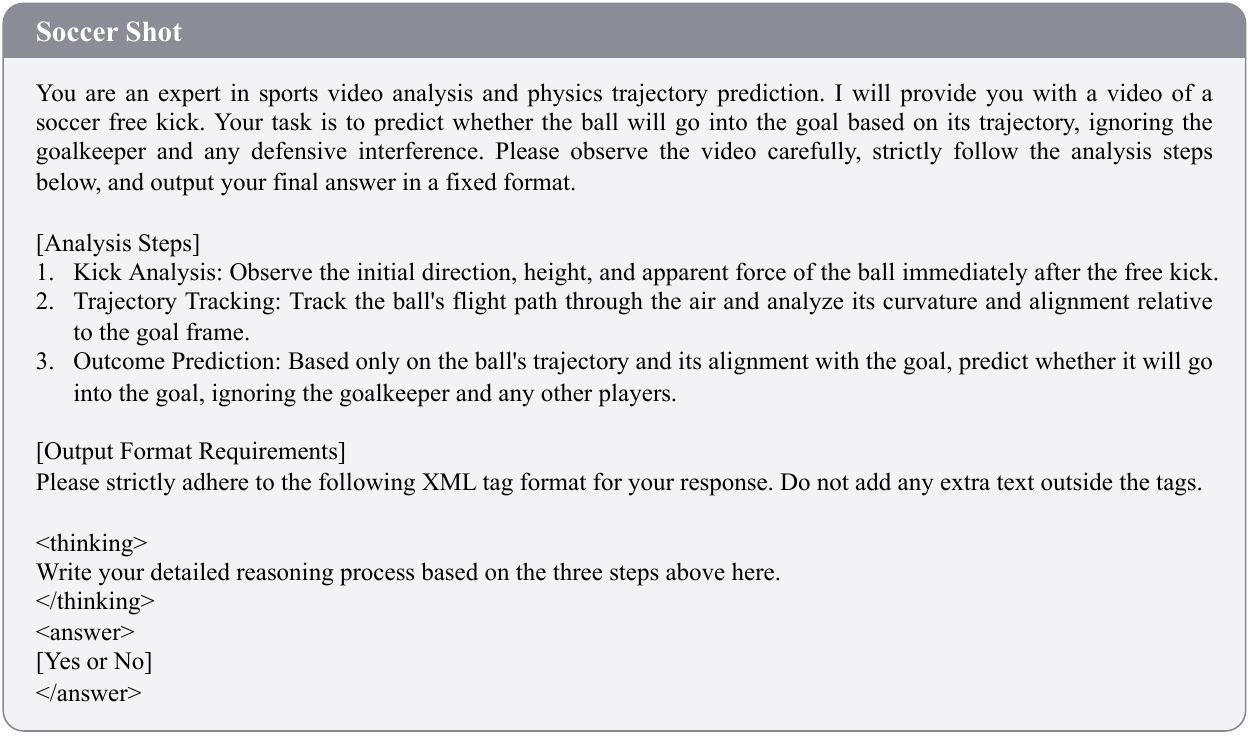}
  \caption{\textbf{Manual CoT prompt template for Soccer Shot.}}
  \label{fig:11}
\end{figure*}

\begin{figure*}[p]
  \centering
  \includegraphics[width=0.95\linewidth]{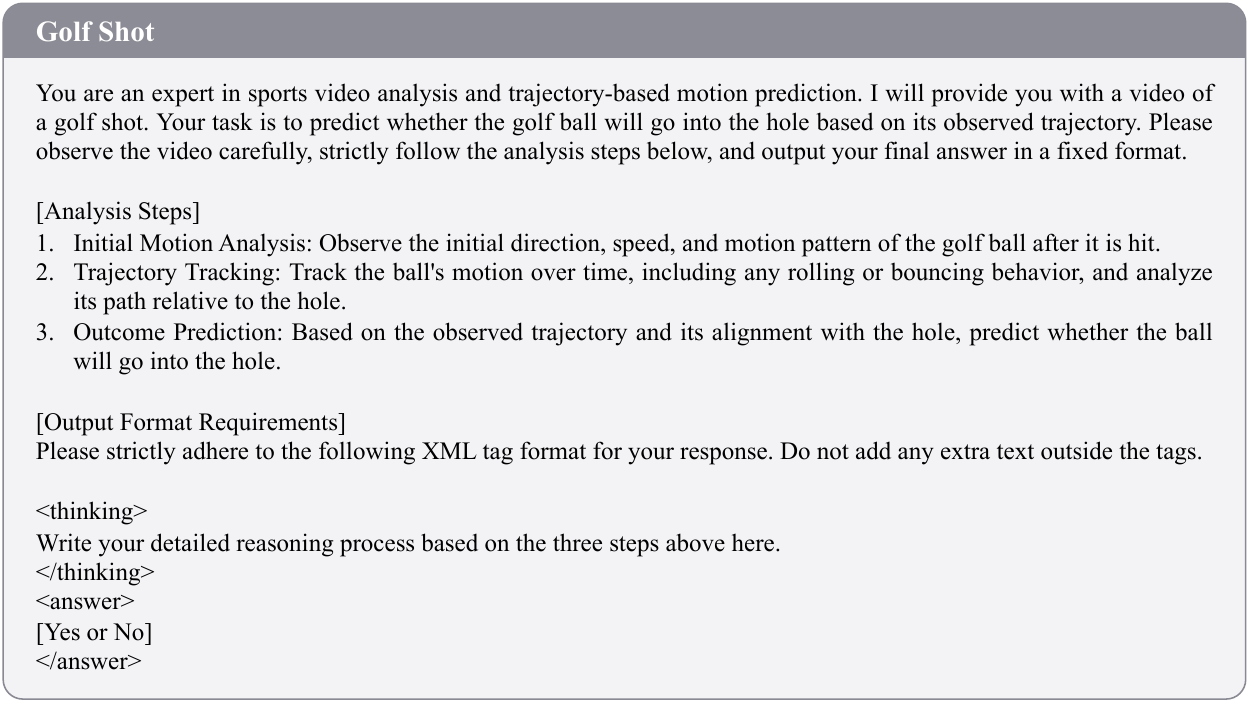}
  \caption{\textbf{Manual CoT prompt template for Golf Shot.}}
  \label{fig:12}
\end{figure*}

\begin{figure*}[p]
  \centering
  \includegraphics[width=0.95\linewidth]{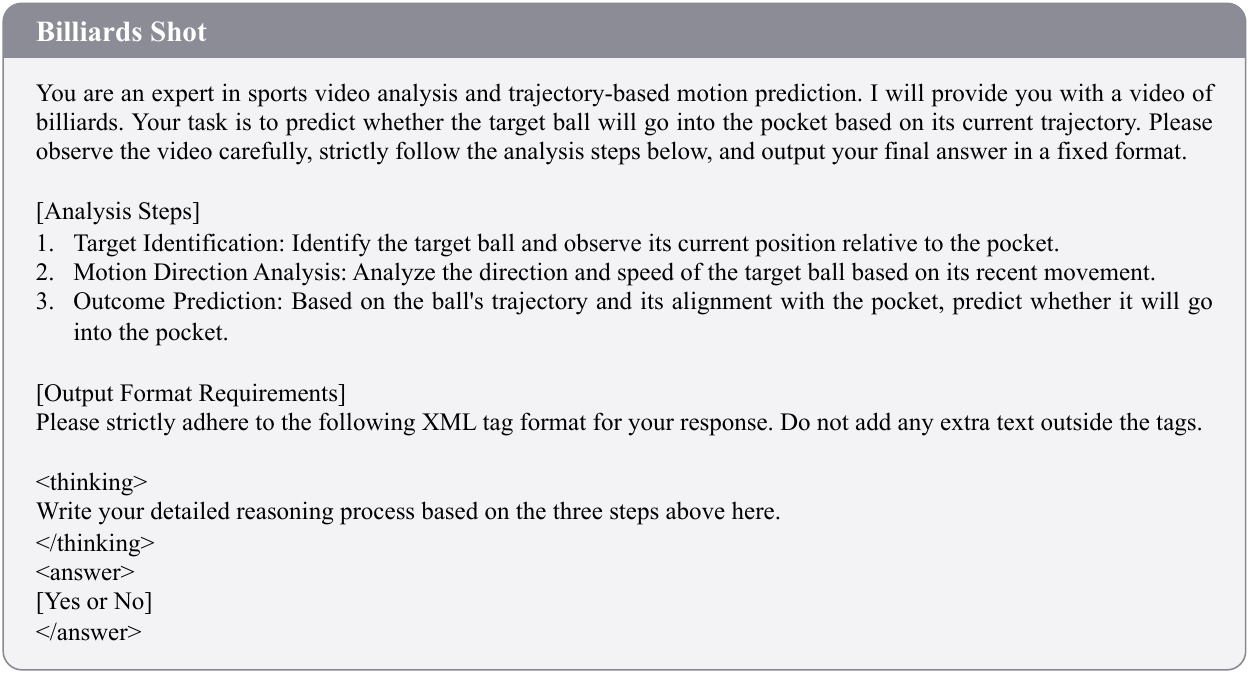}
  \caption{\textbf{Manual CoT prompt template for Billiards Shot.}}
  \label{fig:13}
\end{figure*}

\begin{figure*}[p]
  \centering
  \includegraphics[width=0.95\linewidth]{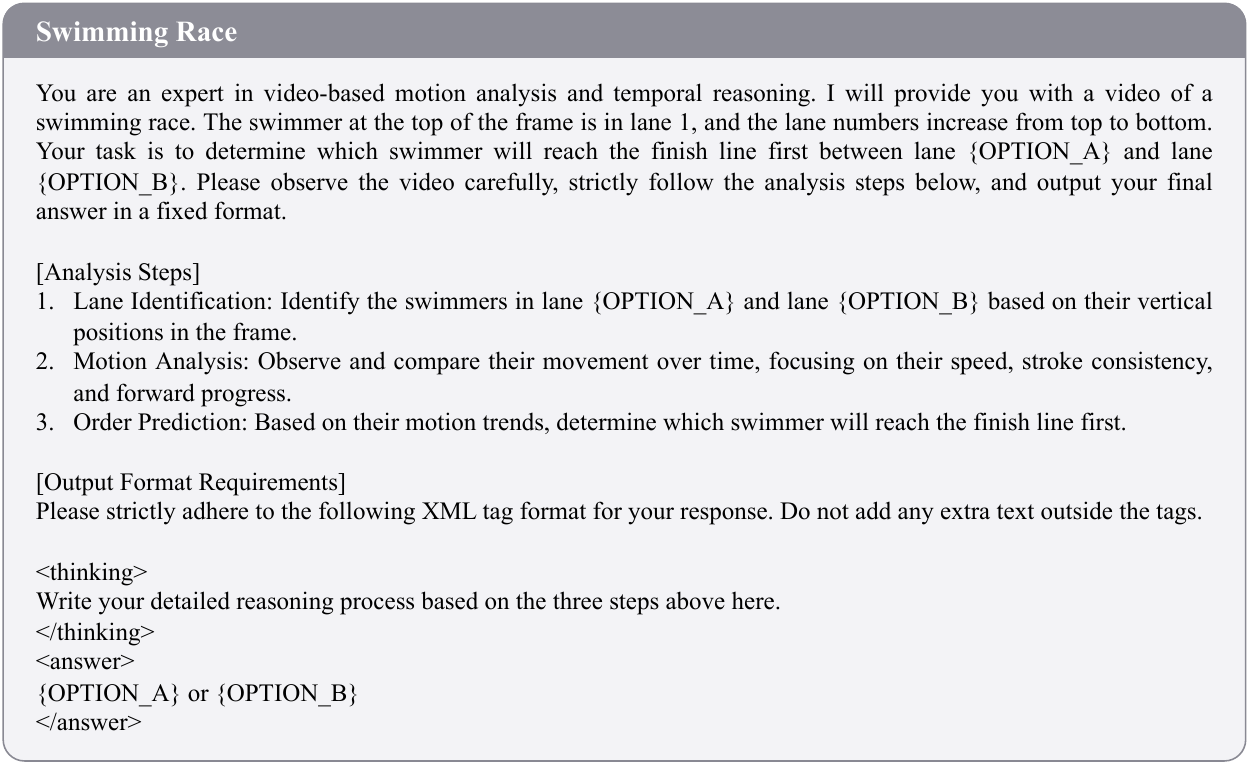}
  \caption{\textbf{Manual CoT prompt template for Swimming Race.}}
  \label{fig:14}
\end{figure*}

\begin{figure*}[p]
  \centering
  \includegraphics[width=0.95\linewidth]{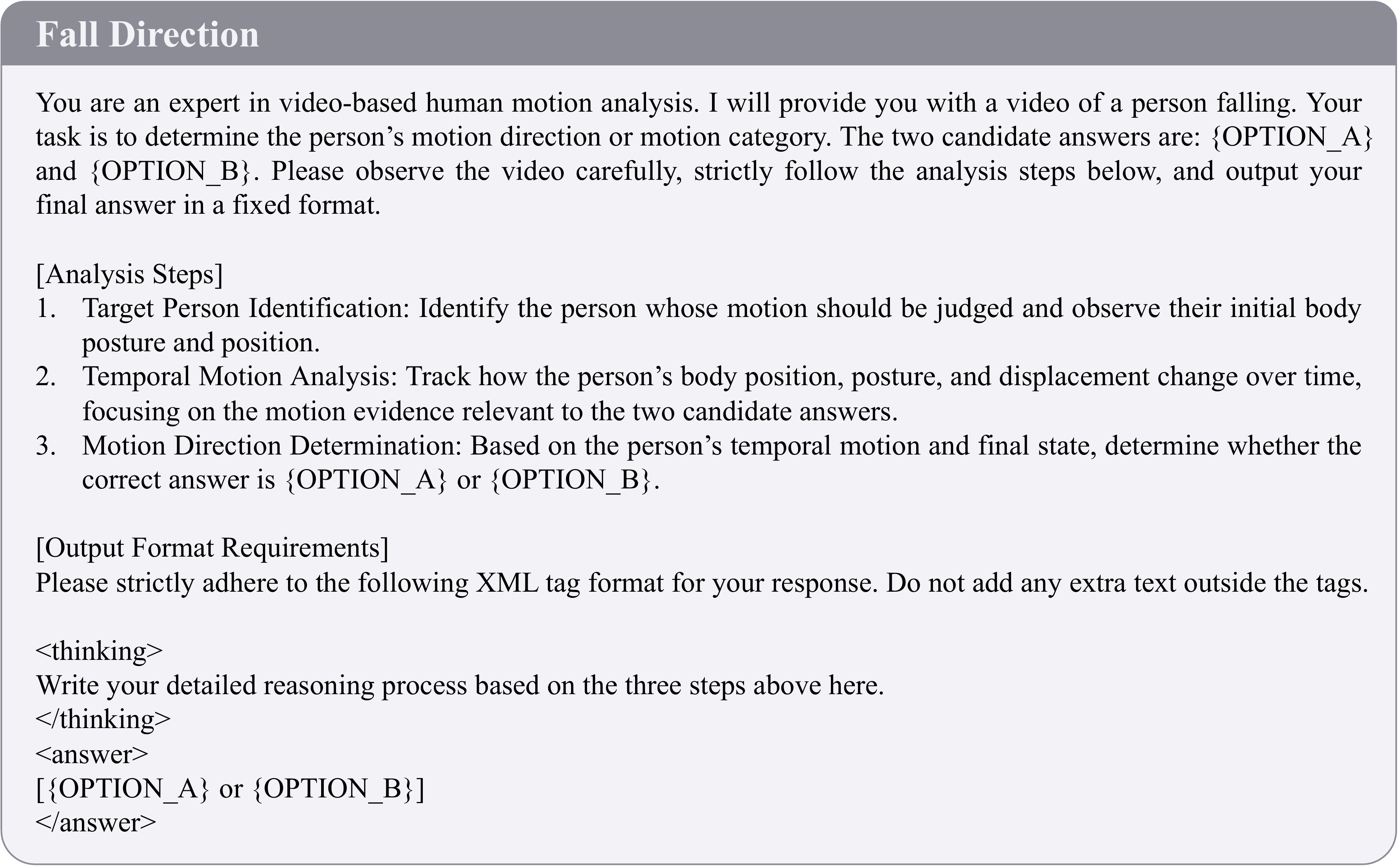}
  \caption{\textbf{Manual CoT prompt template for Fall Direction.}}
  \label{fig:33}
\end{figure*}

\begin{figure*}[p]
  \centering
  \includegraphics[width=0.95\linewidth]{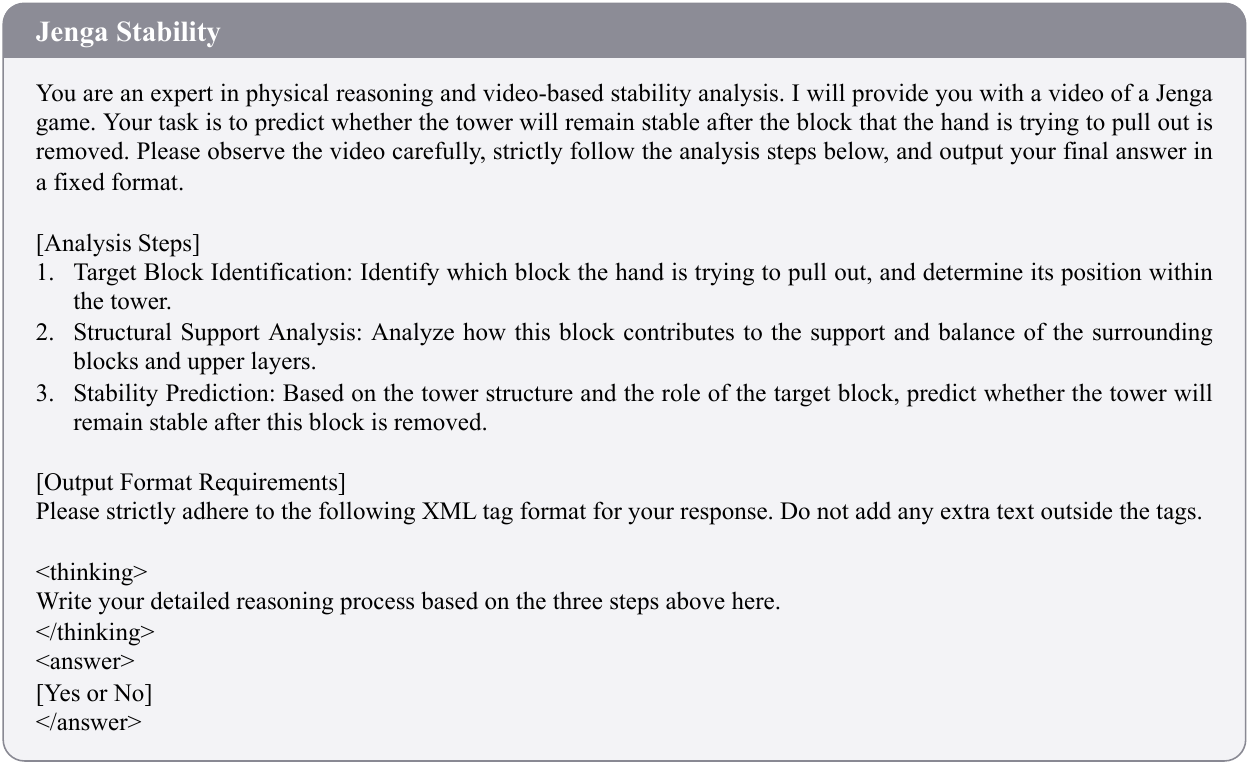}
  \caption{\textbf{Manual CoT prompt template for Jenga Stability.}}
  \label{fig:15}
\end{figure*}

\begin{figure*}[p]
  \centering
  \includegraphics[width=0.95\linewidth]{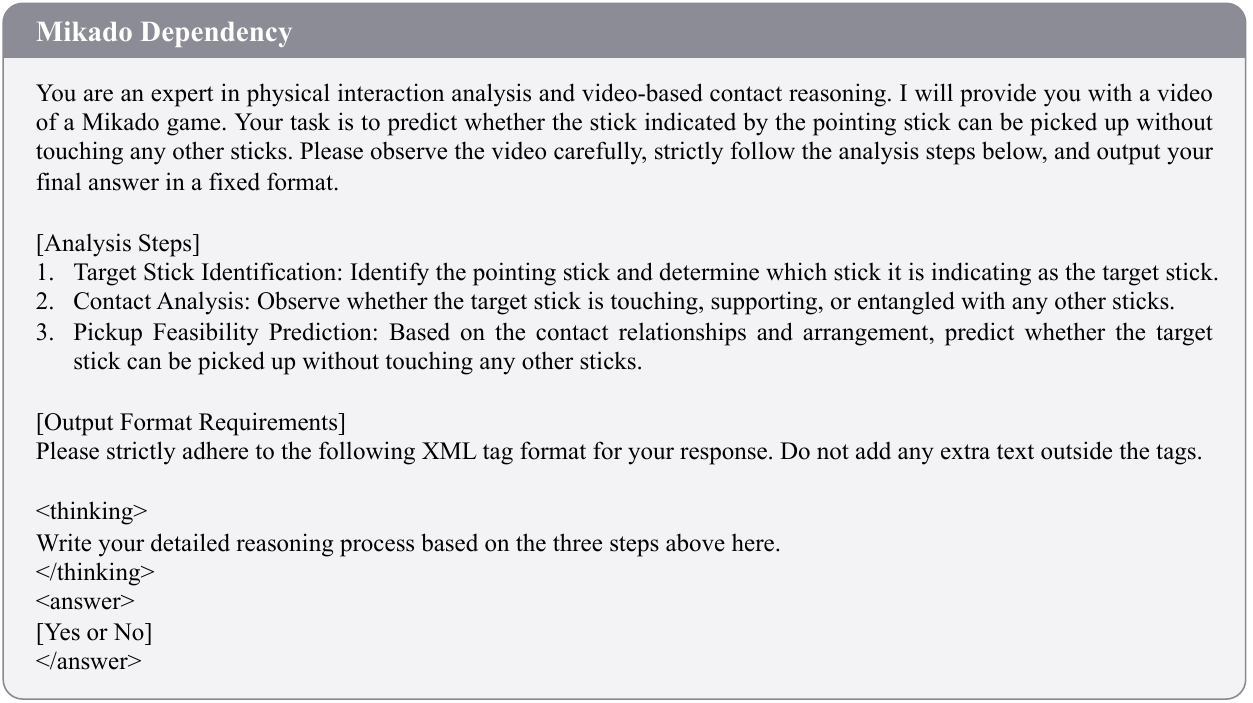}
  \caption{\textbf{Manual CoT prompt template for Mikado Dependency.}}
  \label{fig:16}
\end{figure*}

\begin{figure*}[p]
  \centering
  \includegraphics[width=0.95\linewidth]{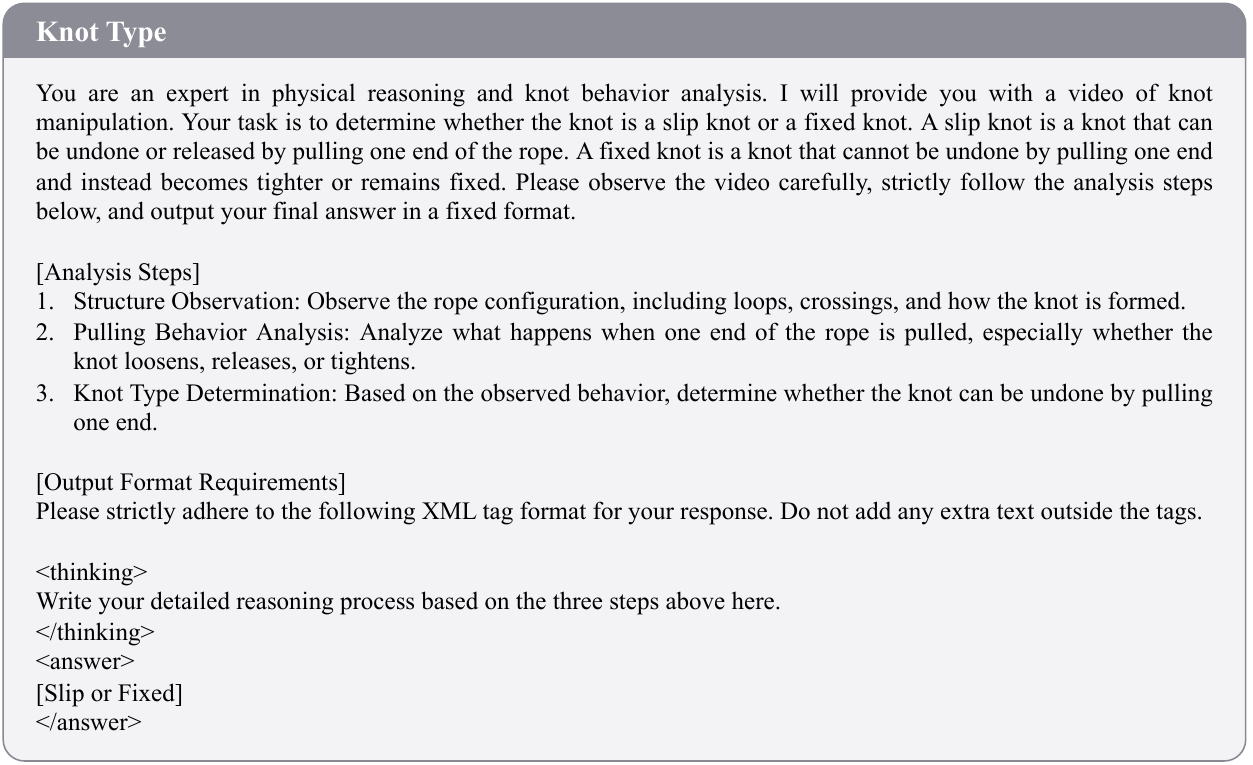}
  \caption{\textbf{Manual CoT prompt template for Knot Type.}}
  \label{fig:17}
\end{figure*}

Inspired by OmniSpatial~\cite{OmniSpatial}, Manual CoT explicitly decomposes each task into visually grounded intermediate reasoning steps.
This design differs from existing methods such as Zero-shot CoT~\cite{zero-shot-cot}, Self-Consistency~\cite{self-consistency-coT}, and Plan-and-Solve~\cite{plan-and-solve}, which rely on generic reasoning instructions or repeated sampling with the same prompt.
Specifically, we manually design a task-specific reasoning procedure for each subtask in ViSTR-Bench, guiding the model to first identify the relevant visual target, then track task-specific spatial-temporal evidence across the video, and finally produce the prediction using the prescribed answer format.
The complete Manual CoT prompt templates are shown in Figures~\ref{fig:6}-\ref{fig:17}.

\section{Qualitative Visualizations}
\label{sec:visualization}

\begin{figure*}[p]
  \centering
  \includegraphics[width=0.95\linewidth]{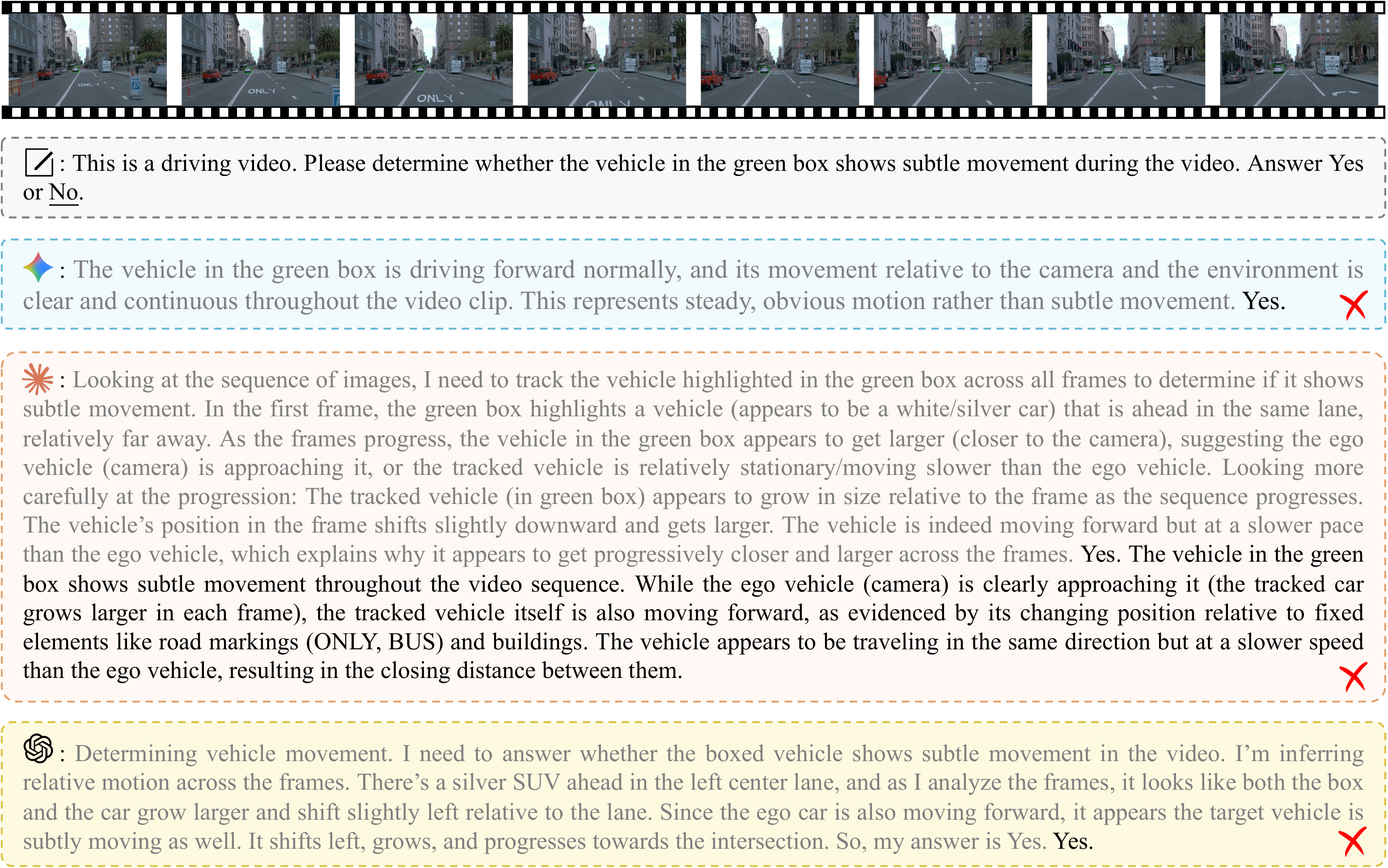}
  \caption{\textbf{Qualitative visualization for Vehicle Movement.}
  We compare the outputs of GPT-5.4-thinking~\cite{GPT-5}, Gemini-3.1-Pro-Preview~\cite{Gemini}, and Claude-Opus-4.6-thinking~\cite{Claude}.
  Gray text denotes built-in thinking content.}
  \label{fig:19}
\end{figure*}

\begin{figure*}[p]
  \centering
  \includegraphics[width=0.95\linewidth]{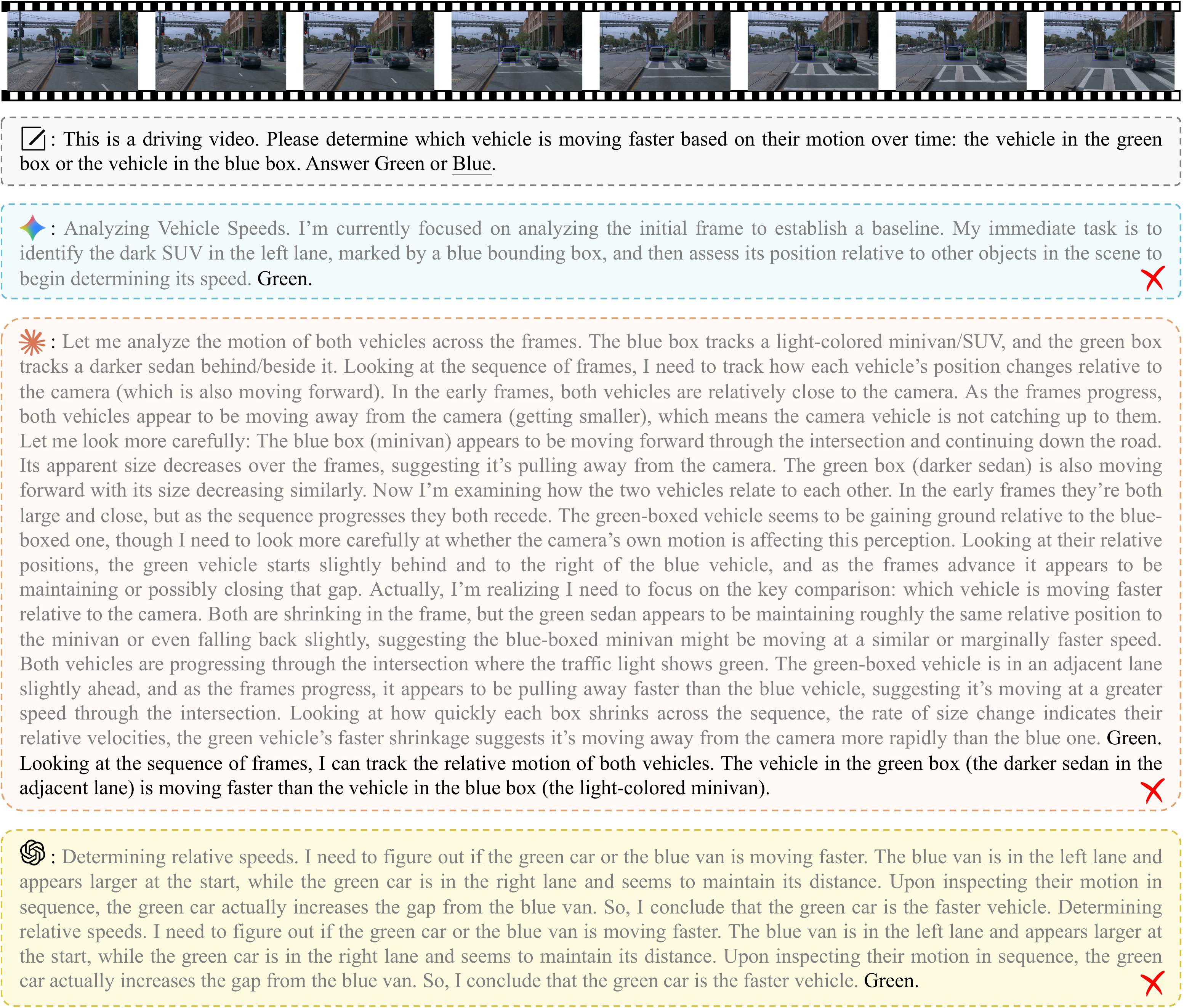}
  \caption{\textbf{Qualitative visualization for Relative Velocity.}
  We compare the outputs of GPT-5.4-thinking~\cite{GPT-5}, Gemini-3.1-Pro-Preview~\cite{Gemini}, and Claude-Opus-4.6-thinking~\cite{Claude}.
  Gray text denotes built-in thinking content.}
  \label{fig:20}
\end{figure*}

\begin{figure*}[p]
  \centering
  \includegraphics[width=0.95\linewidth]{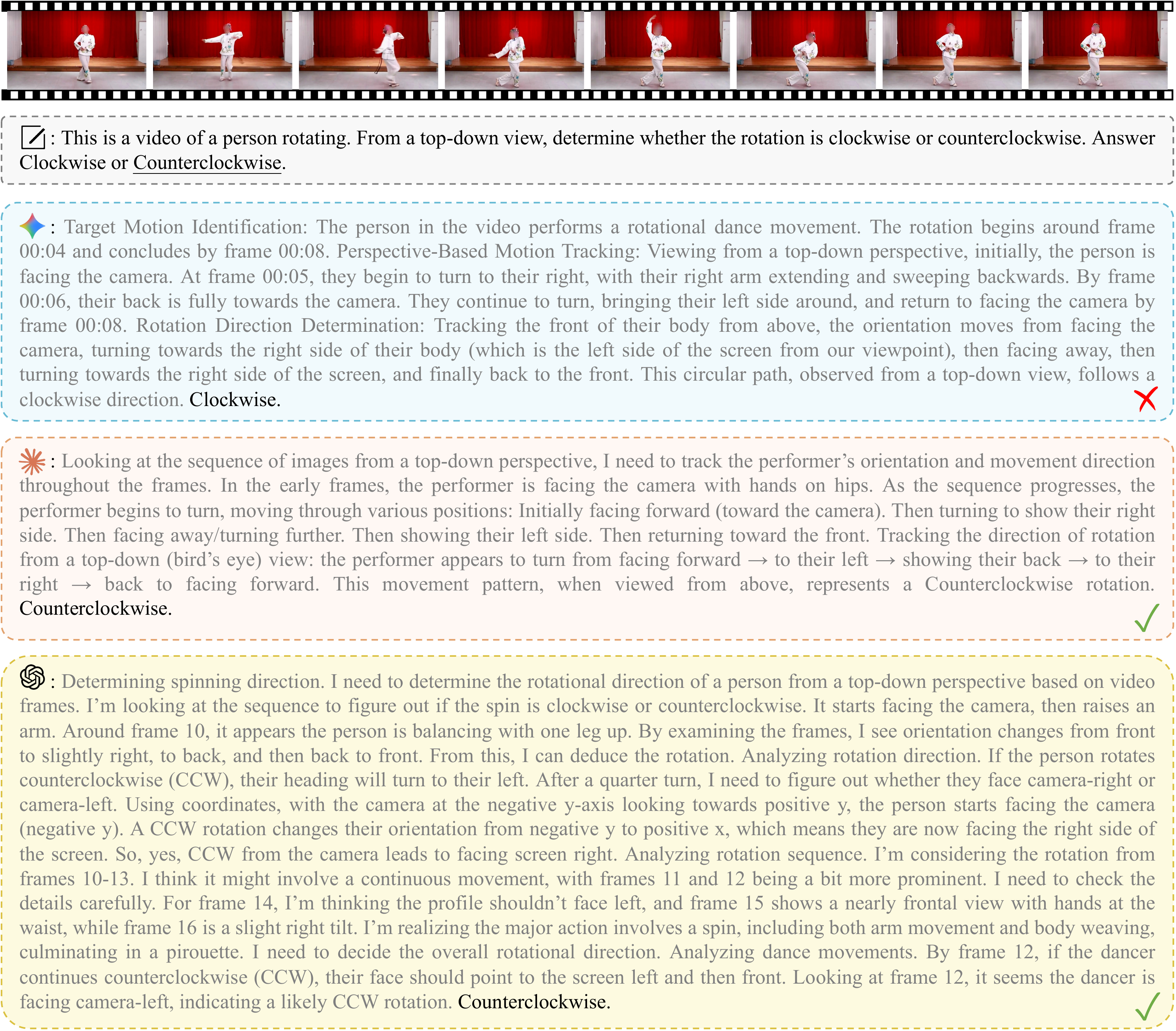}
  \caption{\textbf{Qualitative visualization for Rotation Direction.}
  We compare the outputs of GPT-5.4-thinking~\cite{GPT-5}, Gemini-3.1-Pro-Preview~\cite{Gemini}, and Claude-Opus-4.6-thinking~\cite{Claude}.
  Gray text denotes built-in thinking content.}
  \label{fig:34}
\end{figure*}

\begin{figure*}[p]
  \centering
  \includegraphics[width=0.95\linewidth]{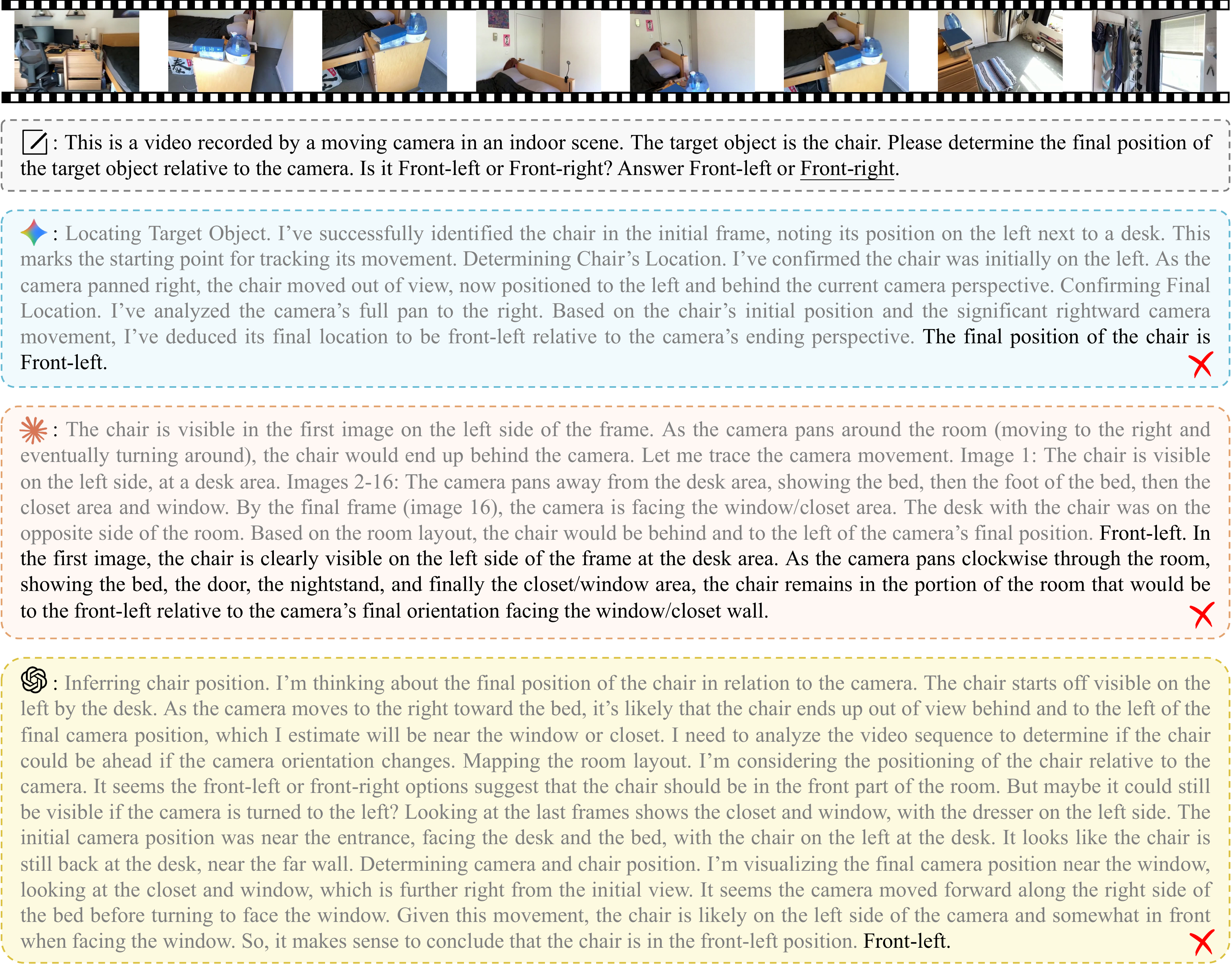}
  \caption{\textbf{Qualitative visualization for Ego Motion.}
  We compare the outputs of GPT-5.4-thinking~\cite{GPT-5}, Gemini-3.1-Pro-Preview~\cite{Gemini}, and Claude-Opus-4.6-thinking~\cite{Claude}.
  Gray text denotes built-in thinking content.}
  \label{fig:21}
\end{figure*}

\begin{figure*}[p]
  \centering
  \includegraphics[width=0.95\linewidth]{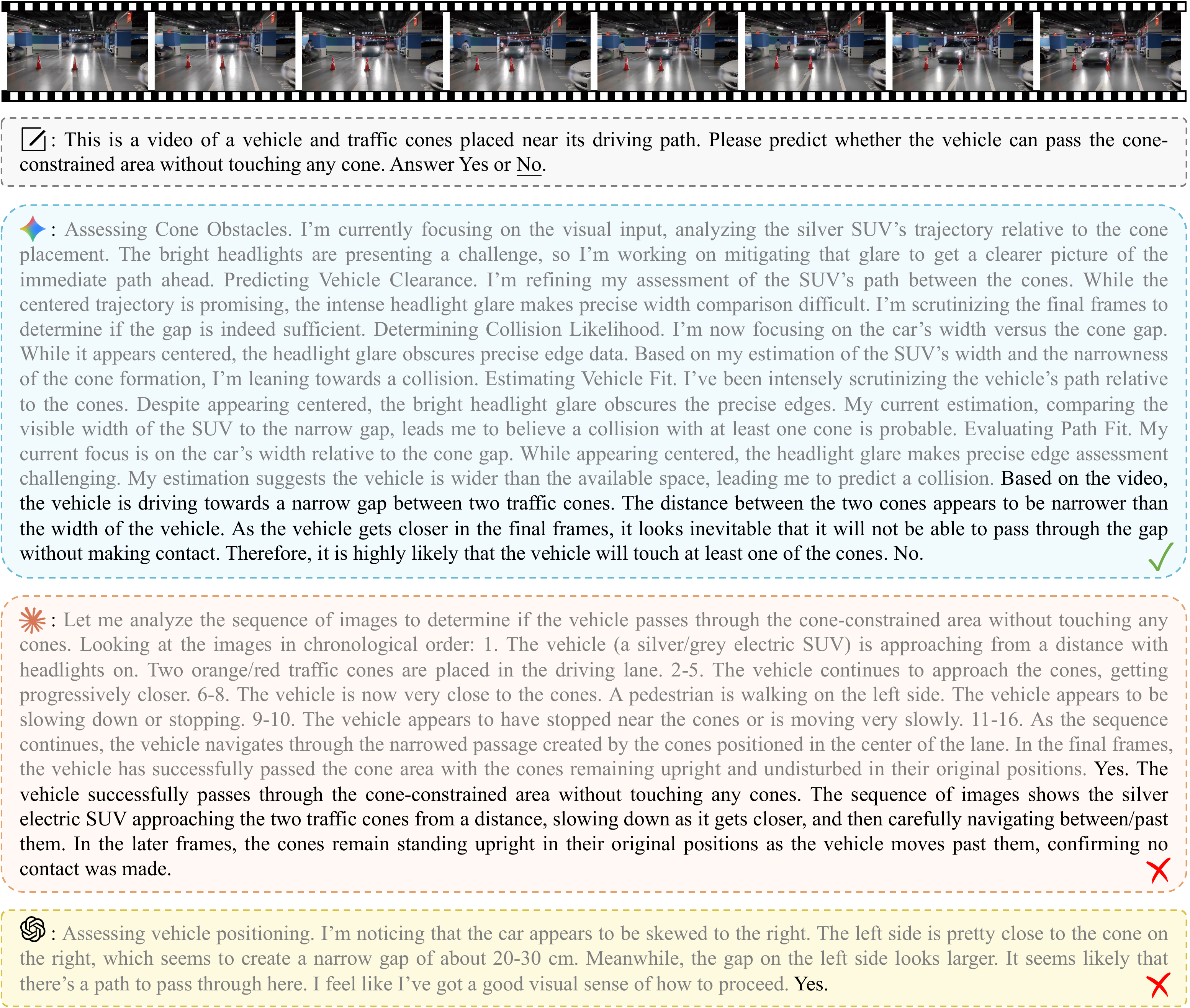}
  \caption{\textbf{Qualitative visualization for Passage Feasibility.}
  We compare the outputs of GPT-5.4-thinking~\cite{GPT-5}, Gemini-3.1-Pro-Preview~\cite{Gemini}, and Claude-Opus-4.6-thinking~\cite{Claude}.
  Gray text denotes built-in thinking content.}
  \label{fig:22}
\end{figure*}

\begin{figure*}[p]
  \centering
  \includegraphics[width=0.95\linewidth]{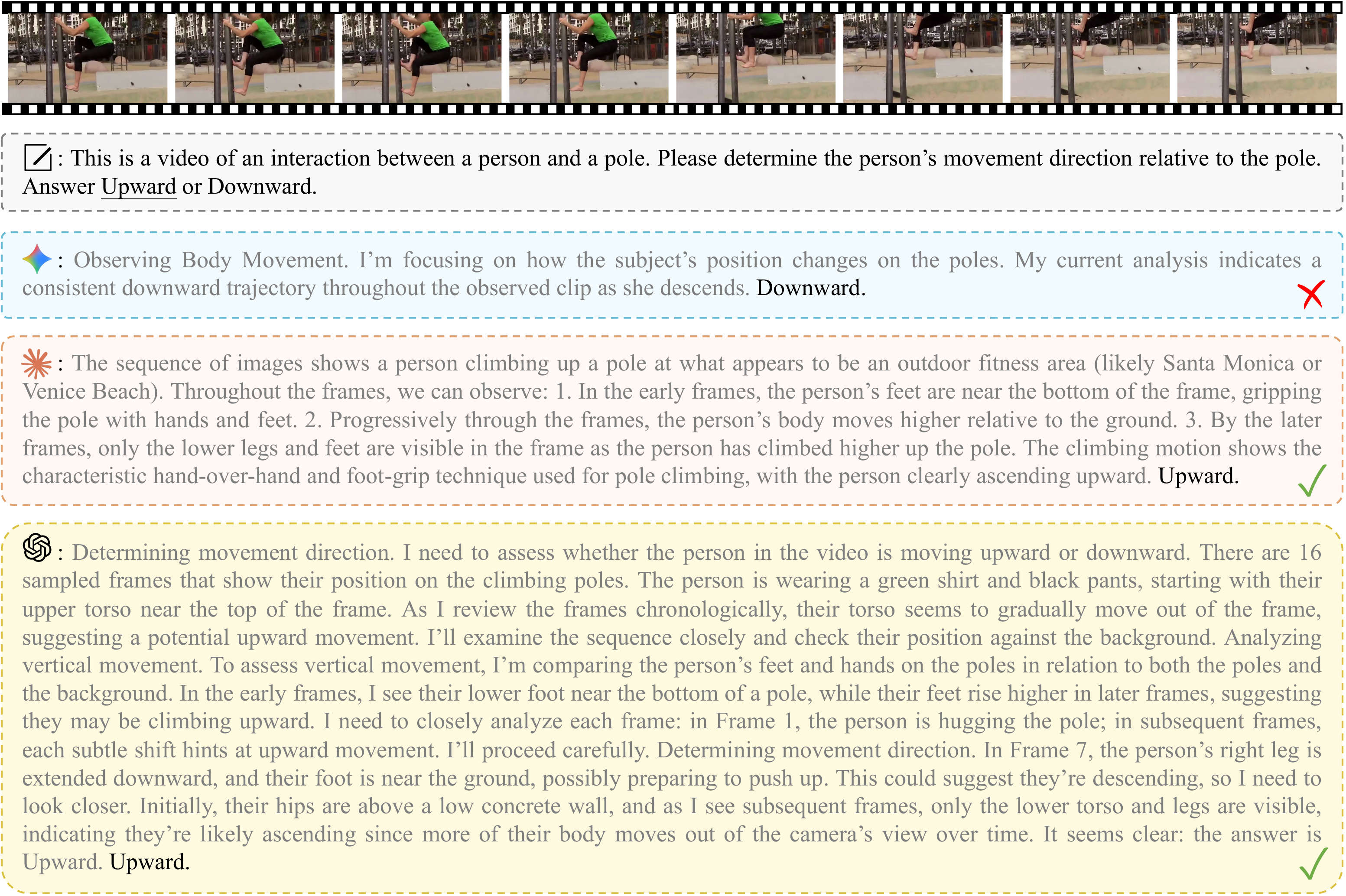}
  \caption{\textbf{Qualitative visualization for Interaction Direction.}
  We compare the outputs of GPT-5.4-thinking~\cite{GPT-5}, Gemini-3.1-Pro-Preview~\cite{Gemini}, and Claude-Opus-4.6-thinking~\cite{Claude}.
  Gray text denotes built-in thinking content.}
  \label{fig:35}
\end{figure*}

\begin{figure*}[p]
  \centering
  \includegraphics[width=0.95\linewidth]{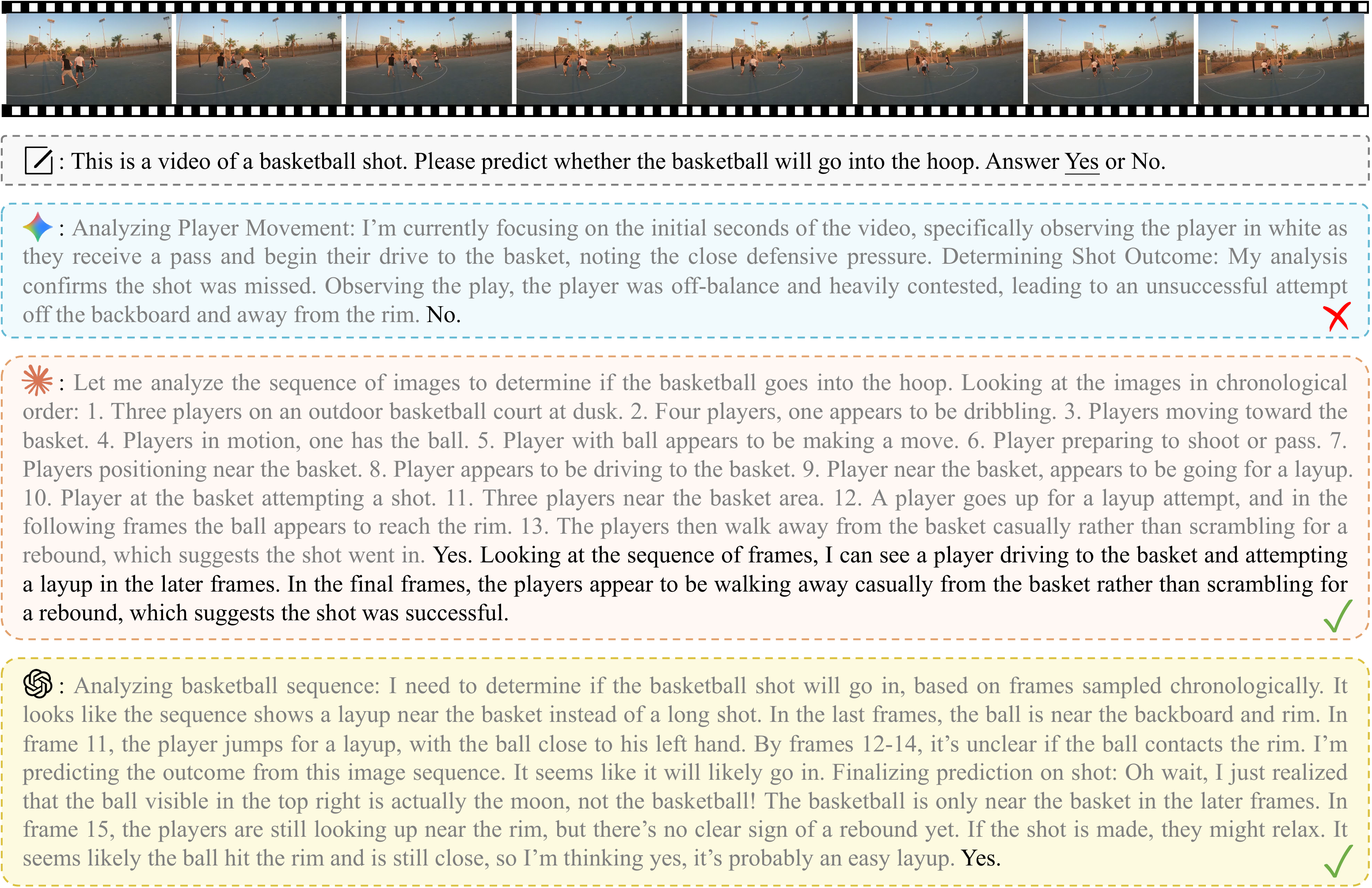}
  \caption{\textbf{Qualitative visualization for Basketball Shot.}
  We compare the outputs of GPT-5.4-thinking~\cite{GPT-5}, Gemini-3.1-Pro-Preview~\cite{Gemini}, and Claude-Opus-4.6-thinking~\cite{Claude}.
  Gray text denotes built-in thinking content.}
  \label{fig:23}
\end{figure*}

\begin{figure*}[p]
  \centering
  \includegraphics[width=0.95\linewidth]{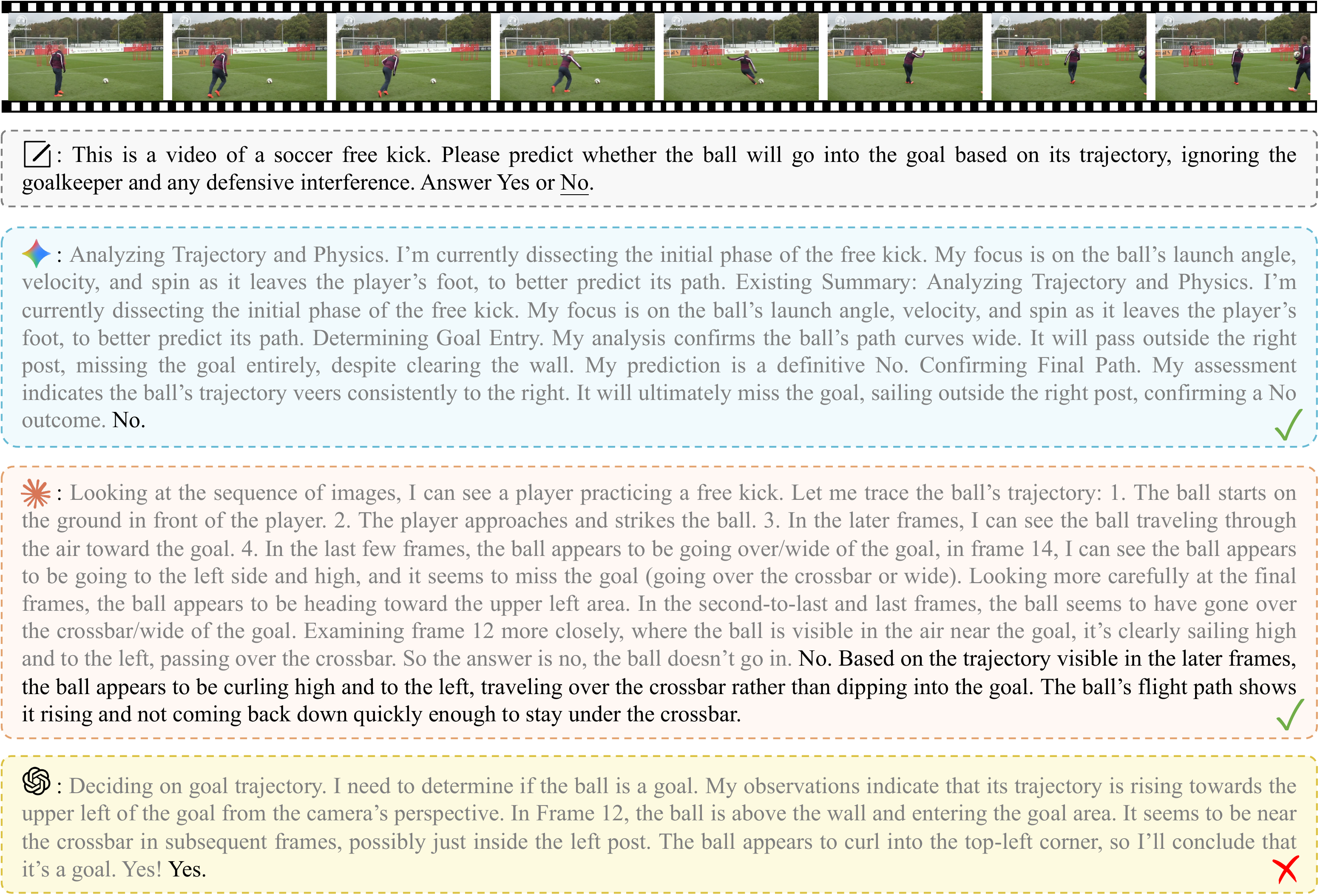}
  \caption{\textbf{Qualitative visualization for Soccer Shot.}
  We compare the outputs of GPT-5.4-thinking~\cite{GPT-5}, Gemini-3.1-Pro-Preview~\cite{Gemini}, and Claude-Opus-4.6-thinking~\cite{Claude}.
  Gray text denotes built-in thinking content.}
  \label{fig:24}
\end{figure*}

\begin{figure*}[p]
  \centering
  \includegraphics[width=0.95\linewidth]{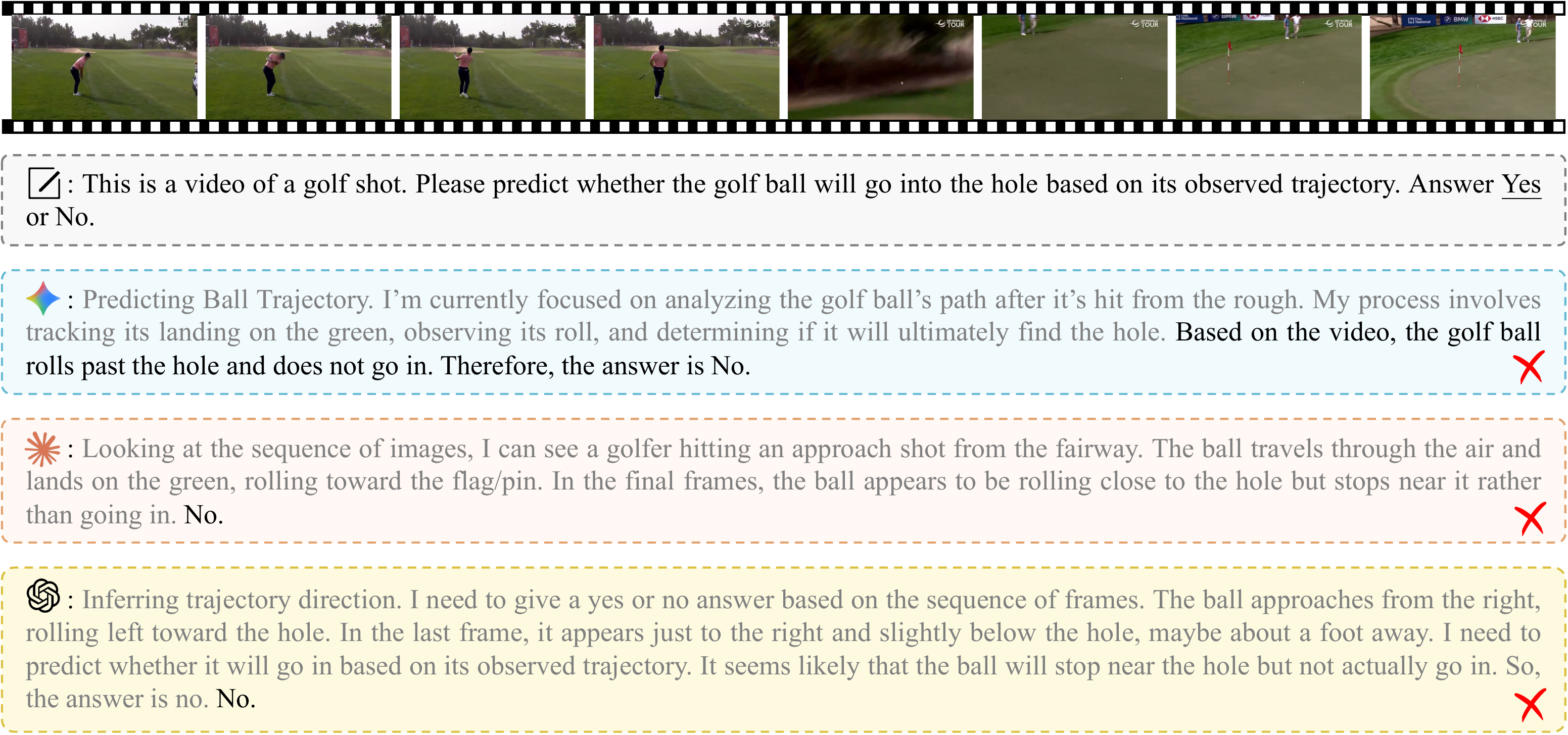}
  \caption{\textbf{Qualitative visualization for Golf Shot.}
  We compare the outputs of GPT-5.4-thinking~\cite{GPT-5}, Gemini-3.1-Pro-Preview~\cite{Gemini}, and Claude-Opus-4.6-thinking~\cite{Claude}.
  Gray text denotes built-in thinking content.}
  \label{fig:25}
\end{figure*}

\begin{figure*}[p]
  \centering
  \includegraphics[width=0.95\linewidth]{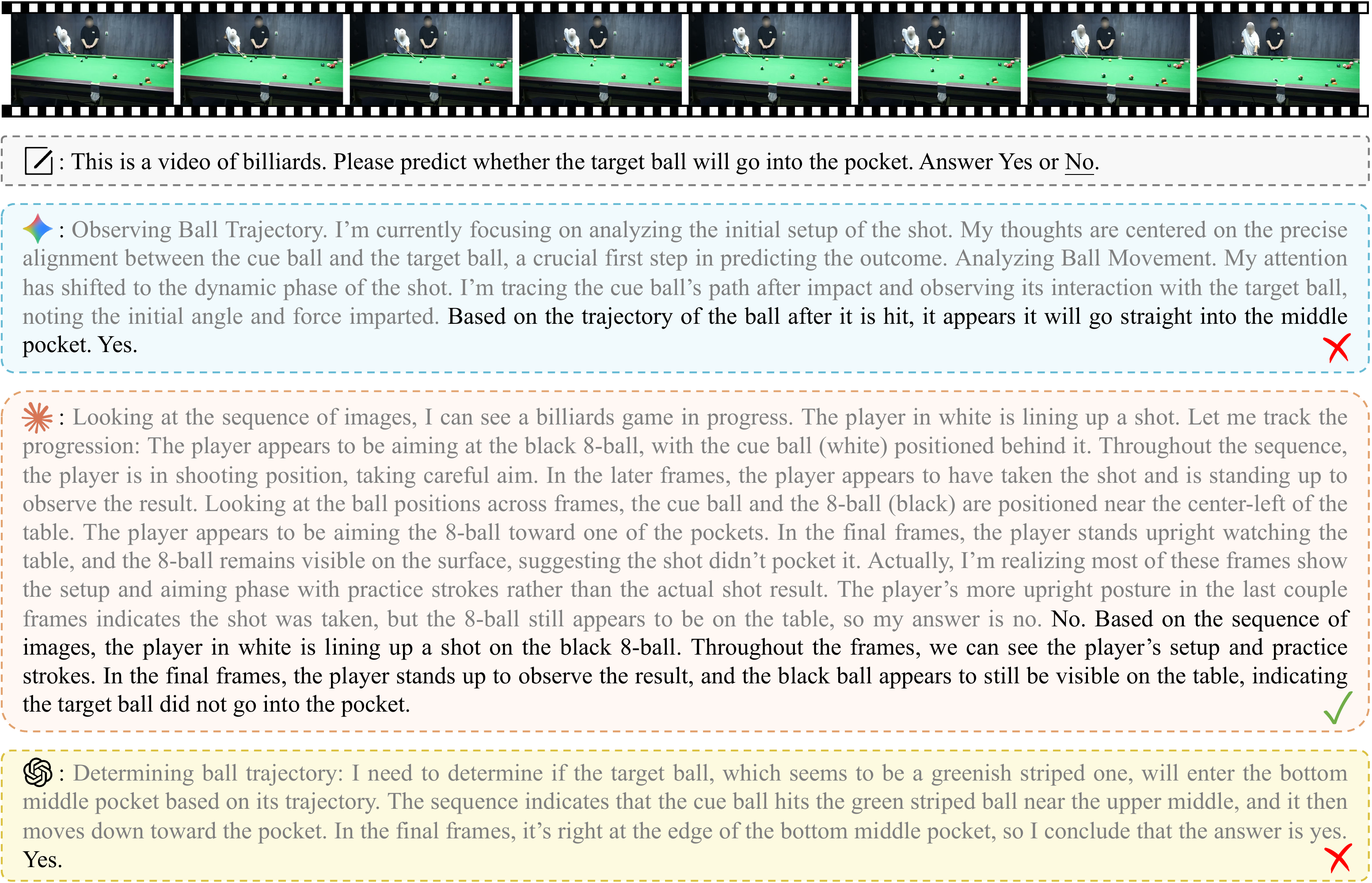}
  \caption{\textbf{Qualitative visualization for Billiards Shot.}
  We compare the outputs of GPT-5.4-thinking~\cite{GPT-5}, Gemini-3.1-Pro-Preview~\cite{Gemini}, and Claude-Opus-4.6-thinking~\cite{Claude}.
  Gray text denotes built-in thinking content.}
  \label{fig:26}
\end{figure*}

\begin{figure*}[p]
  \centering
  \includegraphics[width=0.95\linewidth]{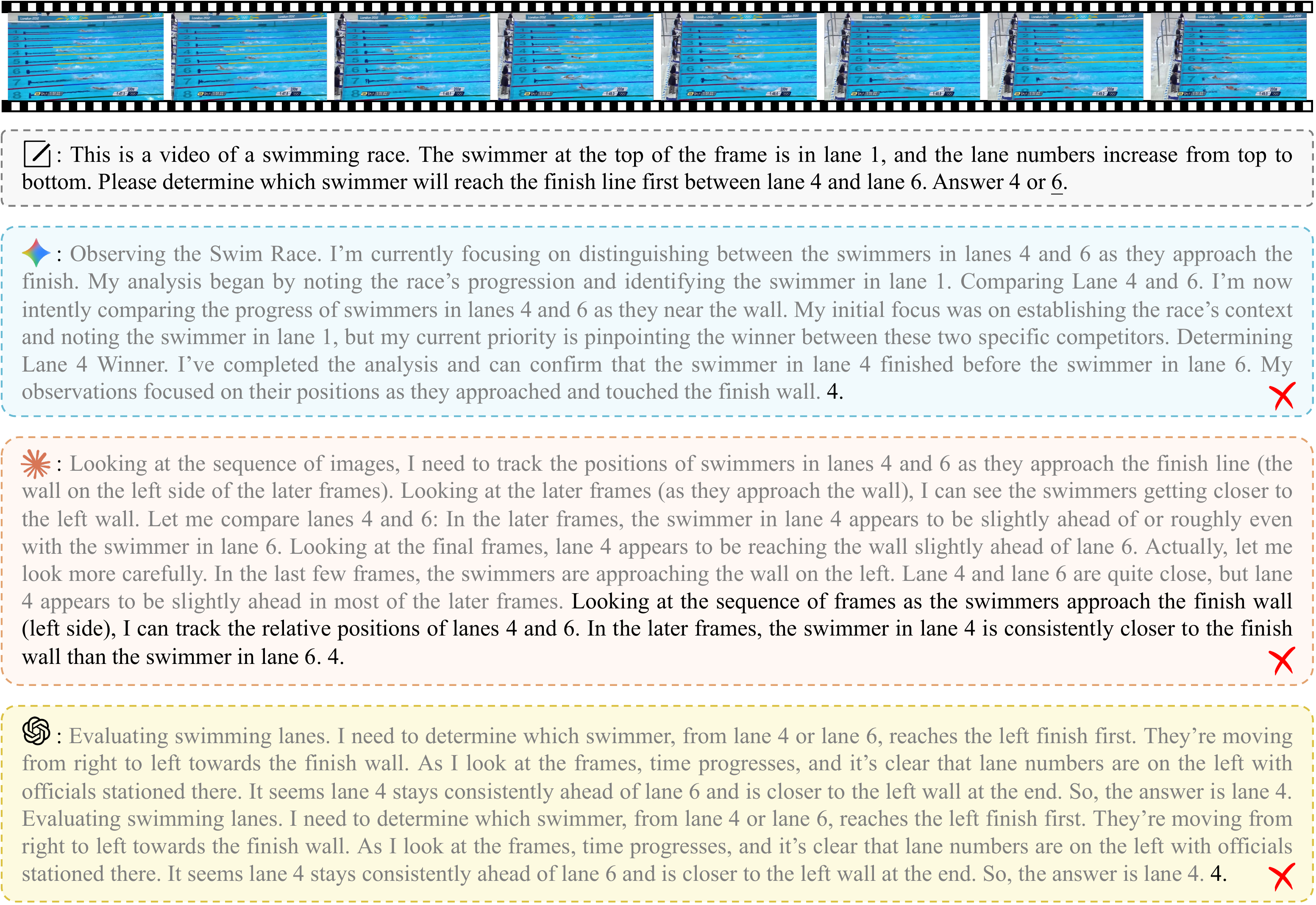}
  \caption{\textbf{Qualitative visualization for Swimming Race.}
  We compare the outputs of GPT-5.4-thinking~\cite{GPT-5}, Gemini-3.1-Pro-Preview~\cite{Gemini}, and Claude-Opus-4.6-thinking~\cite{Claude}.
  Gray text denotes built-in thinking content.}
  \label{fig:27}
\end{figure*}

\begin{figure*}[p]
  \centering
  \includegraphics[width=0.95\linewidth]{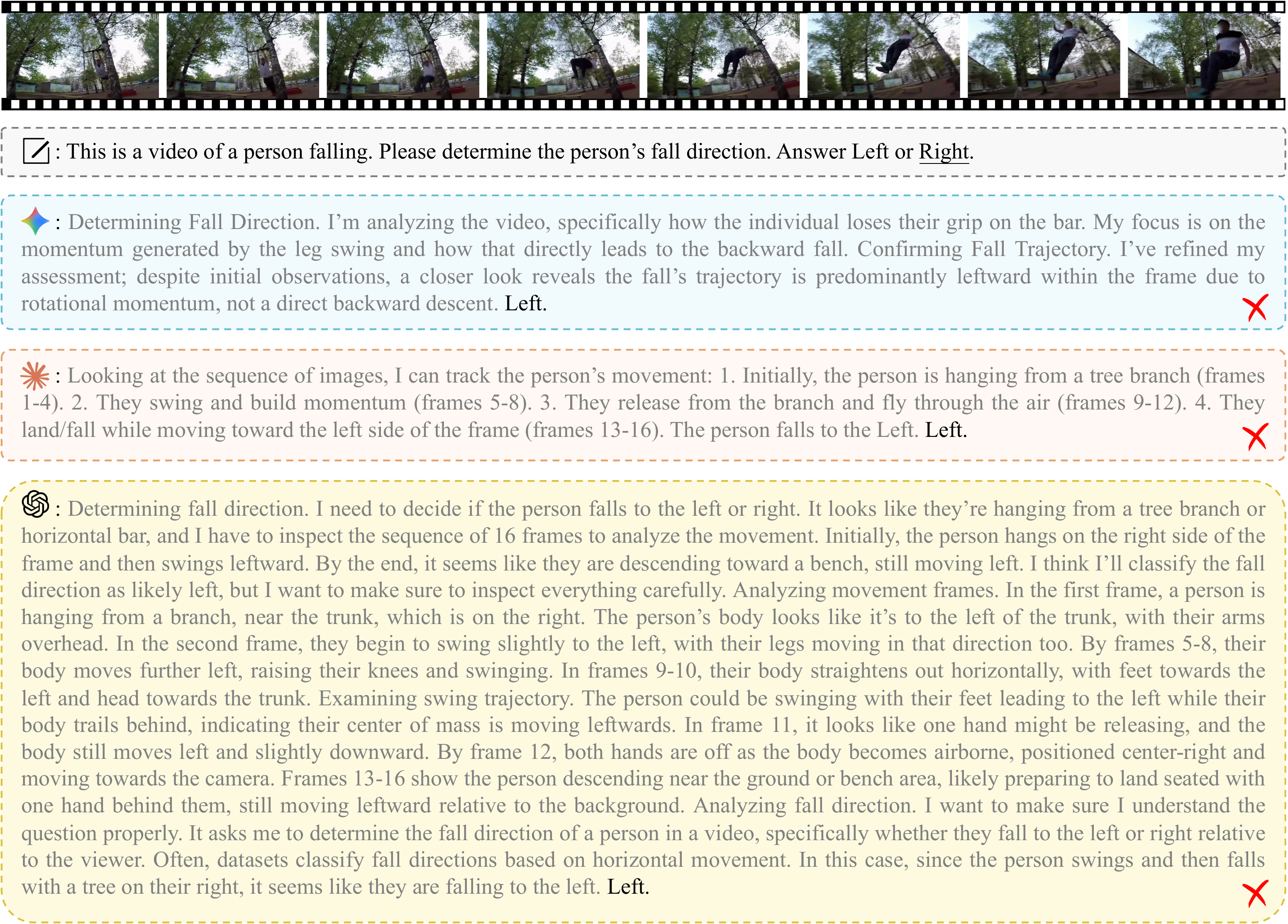}
  \caption{\textbf{Qualitative visualization for Fall Direction.}
  We compare the outputs of GPT-5.4-thinking~\cite{GPT-5}, Gemini-3.1-Pro-Preview~\cite{Gemini}, and Claude-Opus-4.6-thinking~\cite{Claude}.
  Gray text denotes built-in thinking content.}
  \label{fig:36}
\end{figure*}

\begin{figure*}[p]
  \centering
  \includegraphics[width=0.95\linewidth]{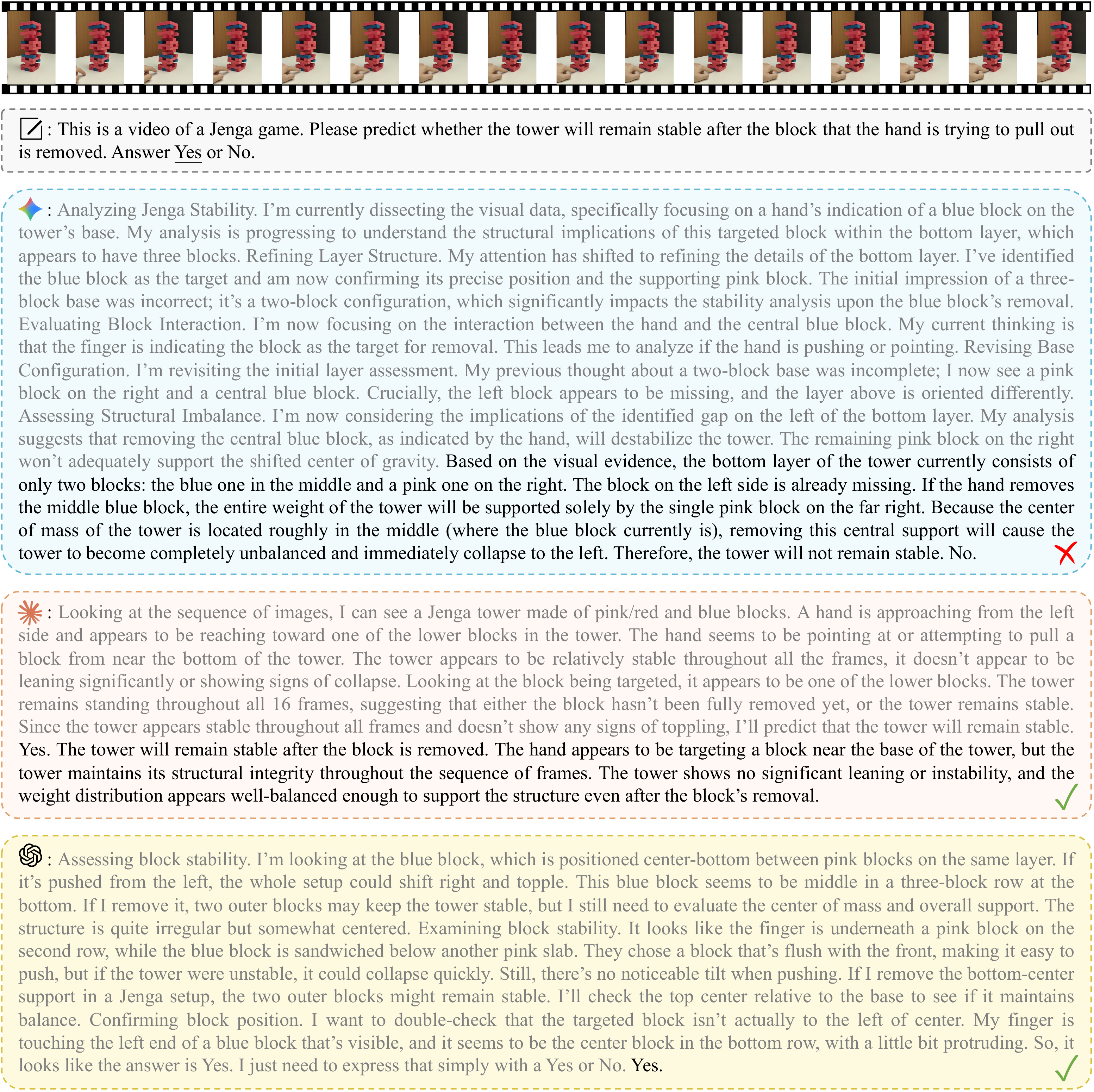}
  \caption{\textbf{Qualitative visualization for Jenga Stability.}
  We compare the outputs of GPT-5.4-thinking~\cite{GPT-5}, Gemini-3.1-Pro-Preview~\cite{Gemini}, and Claude-Opus-4.6-thinking~\cite{Claude}.
  Gray text denotes built-in thinking content.}
  \label{fig:28}
\end{figure*}

\begin{figure*}[p]
  \centering
  \includegraphics[width=0.95\linewidth]{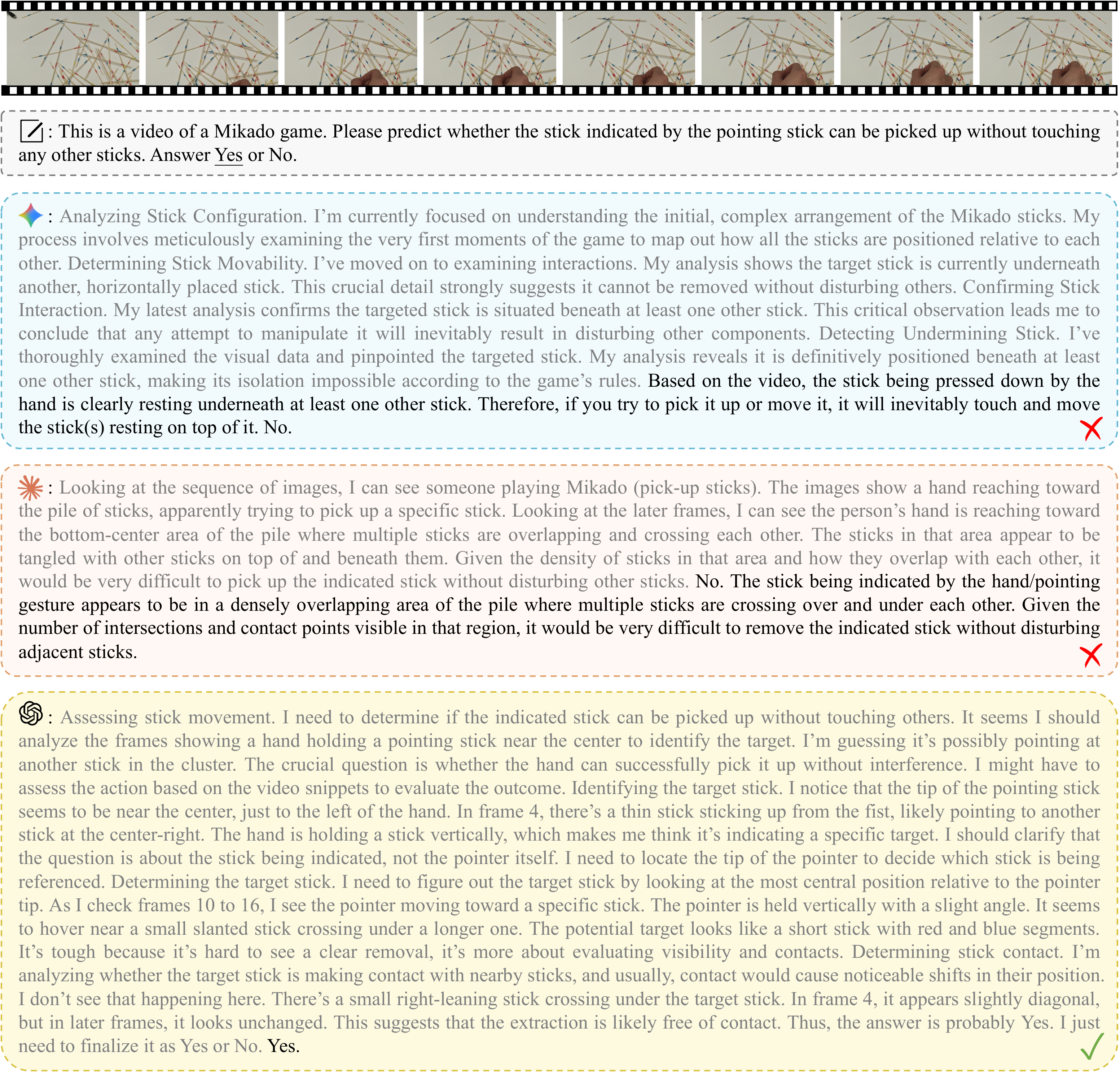}
  \caption{\textbf{Qualitative visualization for Mikado Dependency.}
  We compare the outputs of GPT-5.4-thinking~\cite{GPT-5}, Gemini-3.1-Pro-Preview~\cite{Gemini}, and Claude-Opus-4.6-thinking~\cite{Claude}.
  Gray text denotes built-in thinking content.}
  \label{fig:29}
\end{figure*}

\begin{figure*}[p]
  \centering
  \includegraphics[width=0.95\linewidth]{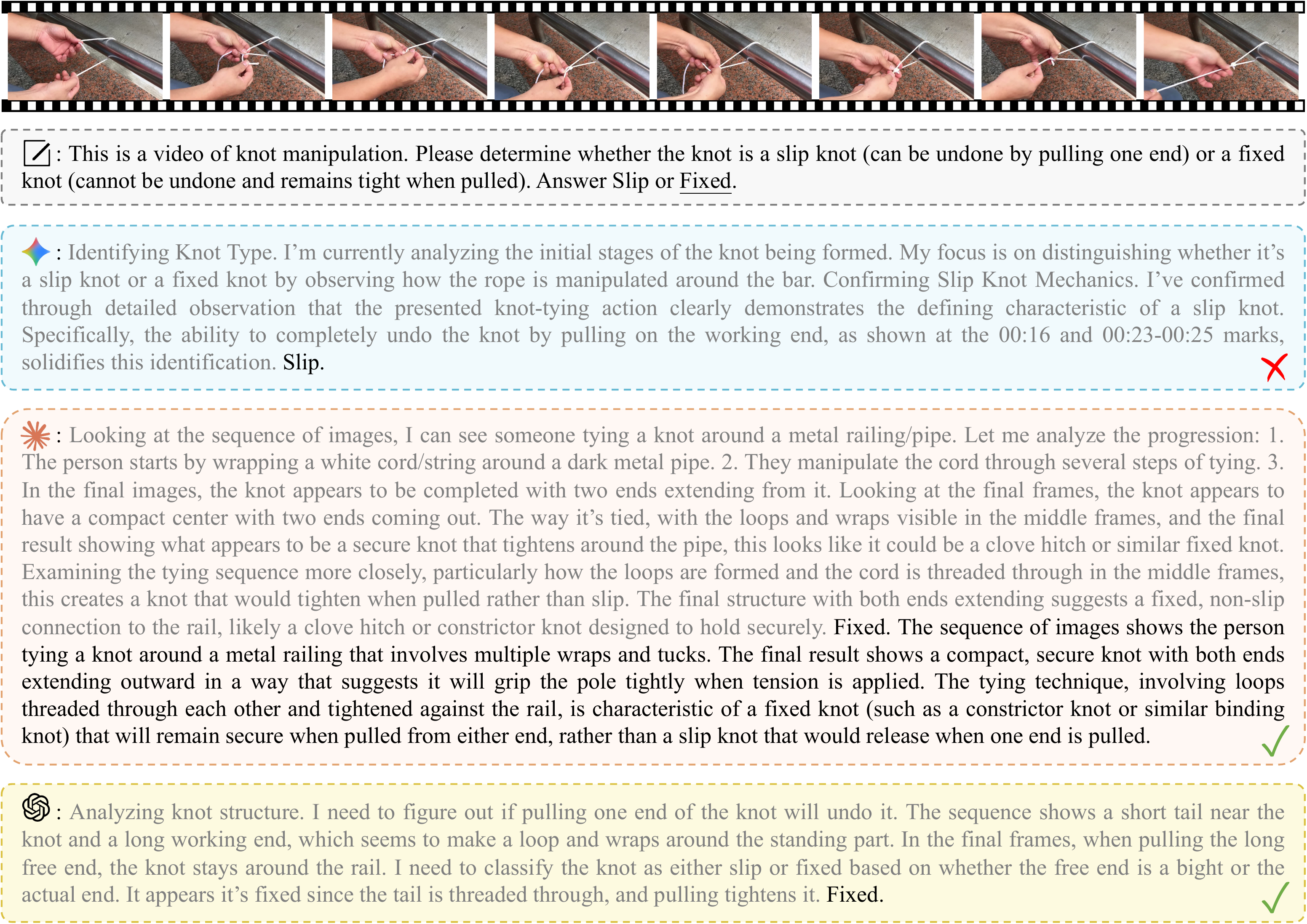}
  \caption{\textbf{Qualitative visualization for Knot Type.}
  We compare the outputs of GPT-5.4-thinking~\cite{GPT-5}, Gemini-3.1-Pro-Preview~\cite{Gemini}, and Claude-Opus-4.6-thinking~\cite{Claude}.
  Gray text denotes built-in thinking content.}
  \label{fig:30}
\end{figure*}

We provide qualitative visualizations for all 15 tasks in ViSTR-Bench.
For each task, we select one representative example and visualize the model outputs from GPT-5.4-thinking~\cite{GPT-5}, Gemini-3.1-Pro-Preview~\cite{Gemini}, and Claude-Opus-4.6-thinking~\cite{Claude}.
The results are shown in Figures~\ref{fig:19}-\ref{fig:30}.

\begin{table*}[!t]
    \centering
    \caption{
        \textbf{Direct Prompting templates used in ViSTR-Bench.}
        Placeholders such as \texttt{\{TARGET\}}, \texttt{\{PERSPECTIVE\}}, \texttt{\{OPTION\_A\}}, and \texttt{\{OPTION\_B\}} are replaced with instance-specific metadata during QA pair generation.
    }
    \label{tab:direct_prompt_templates}
    \small
    \begin{tabular}{p{0.20\textwidth} p{0.70\textwidth}}
        \toprule
        \textbf{Task} & \textbf{Prompt Template} \\
        \midrule

        \multicolumn{2}{l}{\textbf{Motion Perception}} \\
        \quad Vehicle Movement
        & This is a driving video. Please determine whether the vehicle in the green box shows subtle movement during the video. Answer Yes or No. \\

        \quad Relative Velocity
        & This is a driving video. Please determine which vehicle is moving faster based on their motion over time: the vehicle in the green box or the vehicle in the blue box. Answer Green or Blue. \\

        \quad Rotation Direction
        & This is a video of a person. From \texttt{\{PERSPECTIVE\}}, please determine whether the motion is clockwise or counterclockwise. Answer Clockwise or Counterclockwise. \\
        \midrule

        \multicolumn{2}{l}{\textbf{Spatial Relations}} \\
        \quad Ego Motion
        & This is a video recorded by a moving camera in an indoor scene. The target object is the \texttt{\{TARGET\}}. Please determine the final position of the target object relative to the camera. Is it \texttt{\{OPTION\_A\}} or \texttt{\{OPTION\_B\}}? Answer \texttt{\{OPTION\_A\}} or \texttt{\{OPTION\_B\}}. \\

        \quad Passage Feasibility
        & This is a video of a vehicle and traffic cones placed near its driving path. Please predict whether the vehicle can pass the cone-constrained area without touching any cone. Answer Yes or No. \\

        \quad Interaction Direction
        & This is a video of an interaction between a person and \texttt{\{TARGET\}}. Please determine the person's movement direction relative to \texttt{\{TARGET\}}. Answer \texttt{\{OPTION\_A\}} or \texttt{\{OPTION\_B\}}. \\
        \midrule

        \multicolumn{2}{l}{\textbf{Outcome Prediction}} \\
        \quad Basketball Shot
        & This is a video of a basketball shot. Please predict whether the basketball will go into the hoop. Answer Yes or No. \\

        \quad Soccer Shot
        & This is a video of a soccer free kick. Please predict whether the ball will go into the goal based on its trajectory, ignoring the goalkeeper and any defensive interference. Answer Yes or No. \\

        \quad Golf Shot
        & This is a video of a golf shot. Please predict whether the golf ball will go into the hole based on its observed trajectory. Answer Yes or No. \\

        \quad Billiards Shot
        & This is a video of billiards. Please predict whether the target ball will go into the pocket based on its current trajectory. Answer Yes or No. \\

        \quad Swimming Race
        & This is a video of a swimming race. The swimmer at the top of the frame is in lane 1, and the lane numbers increase from top to bottom. Please determine which swimmer will reach the finish line first between lane \texttt{\{OPTION\_A\}} and lane \texttt{\{OPTION\_B\}}. Answer \texttt{\{OPTION\_A\}} or \texttt{\{OPTION\_B\}}. \\

        \quad Fall Direction
        & This is a video of a person falling. Please determine the person's fall direction. Answer \texttt{\{OPTION\_A\}} or \texttt{\{OPTION\_B\}}. \\
        \midrule

        \multicolumn{2}{l}{\textbf{Physical Dynamics}} \\
        \quad Jenga Stability
        & This is a video of a Jenga game. Please predict whether the tower will remain stable after the block that the hand is trying to pull out is removed. Answer Yes or No. \\

        \quad Mikado Dependency
        & This is a video of a Mikado game. Please predict whether the stick indicated by the pointing stick can be picked up without touching any other sticks. Answer Yes or No. \\

        \quad Knot Type
        & This is a video of knot manipulation. Please determine whether the knot is a slip knot (can be undone by pulling one end) or a fixed knot (cannot be undone and remains tight when pulled). Answer Slip or Fixed. \\
        \bottomrule
    \end{tabular}
\end{table*}

\clearpage

\end{document}